\documentclass[10pt]{article} % For LaTeX2e
\usepackage[accepted]{tmlr}
\usepackage{amsmath,amsfonts,bm}

\def\eqref#1{equation~\ref{#1}}
\def\1{\bm{1}}

\def\ra{{\textnormal{a}}}

\def\rx{{\textnormal{x}}}

\def\rva{{\mathbf{a}}}

\def\erva{{\textnormal{a}}}

\def\ervx{{\textnormal{x}}}

\def\rmA{{\mathbf{A}}}

\def\vmu{{\bm{\mu}}}
\def\vtheta{{\bm{\theta}}}
\def\va{{\bm{a}}}

\def\ve{{\bm{e}}}

\def\vx{{\bm{x}}}

\def\eva{{a}}

\def\mA{{\bm{A}}}

\def\mH{{\bm{H}}}
\def\mI{{\bm{I}}}
\def\mJ{{\bm{J}}}

\def\mX{{\bm{X}}}

\def\mSigma{{\bm{\Sigma}}}

\DeclareMathAlphabet{\mathsfit}{\encodingdefault}{\sfdefault}{m}{sl}
\SetMathAlphabet{\mathsfit}{bold}{\encodingdefault}{\sfdefault}{bx}{n}
\newcommand{\tens}[1]{\bm{\mathsfit{#1}}}
\def\tA{{\tens{A}}}

\def\tX{{\tens{X}}}

\def\gG{{\mathcal{G}}}

\def\sA{{\mathbb{A}}}
\def\sB{{\mathbb{B}}}

\def\sS{{\mathbb{S}}}

\def\emA{{A}}

\newcommand{\etens}[1]{\mathsfit{#1}}

\def\etA{{\etens{A}}}

\newcommand{\E}{\mathbb{E}}

\newcommand{\R}{\mathbb{R}}

\newcommand{\KL}{D_{\mathrm{KL}}}
\newcommand{\Var}{\mathrm{Var}}

\newcommand{\Cov}{\mathrm{Cov}}
\newcommand{\normltwo}{L^2}
\newcommand{\normlp}{L^p}

\newcommand{\parents}{Pa} % See usage in notation.tex. Chosen to match Daphne's book.

\usepackage{hyperref}
\usepackage{url}
\usepackage{booktabs}       % professional-quality tables
\usepackage{amsfonts}       % blackboard math symbols
\usepackage{nicefrac}       % compact symbols for 1/2, etc.
\usepackage{microtype}      % microtypography
\usepackage{xcolor}         % colors
\usepackage{graphicx}
\usepackage{subfigure}

\usepackage{amsmath}
\usepackage{amssymb}
\usepackage{mathtools}
\usepackage{amsthm}
\usepackage{wrapfig}
\usepackage{multirow}
\usepackage{subcaption}
\usepackage{comment}
\usepackage{nicefrac}

\theoremstyle{plain}
\newtheorem{theorem}{Theorem}[section]

\theoremstyle{definition}
\newtheorem{definition}[theorem]{Definition}

\newtheorem{hypothesis}[theorem]{Hypothesis}
\theoremstyle{remark}

\usepackage[textsize=tiny]{todonotes}
\definecolor{mydarkgreen}{RGB}{0, 139, 69}

\title{On Memorization in Diffusion Models}

\author{\name Xiangming Gu\footnotemark[1] \email xiangming@u.nus.edu \\
      \addr National University of Singapore
      \AND
      \name Chao Du\footnotemark[2] \email duchao@sea.com \\
      \addr Sea AI Lab
      \AND
      \name Tianyu Pang\footnotemark[2] \email tianyupang@sea.com \\
      \addr Sea AI Lab
      \AND
      \name Chongxuan Li \email chongxuanli@ruc.edu.cn \\
      \addr Gaoling School of Artificial Intelligence\\
      Renmin University of China
      \AND
      \name Min Lin \email linmin@sea.com \\
      \addr Sea AI Lab
      \AND
      \name Ye Wang\footnotemark[2] \email wangye@comp.nus.edu.sg\\
      \addr National University of Singapore
      }

\def\month{02}  % Insert correct month for camera-ready version
\def\year{2025} % Insert correct year for camera-ready version
\def\openreview{\url{https://openreview.net/forum?id=D3DBqvSDbj}} % Insert correct link to OpenReview for camera-ready version

\begin{document}

\maketitle

\renewcommand{\thefootnote}{\fnsymbol{footnote}}
\footnotetext[1]{Work done during an internship at Sea AI Lab.}
\footnotetext[2]{Corresponding authors.}
\renewcommand{\thefootnote}{\arabic{footnote}}

\begin{abstract}

Due to their capacity to generate novel and high-quality samples, diffusion models have attracted significant research interest in recent years. Notably, the typical training objective of diffusion models, i.e., denoising score matching, has a closed-form optimal solution that can only generate training-data replicating samples. This indicates that a memorization behavior is theoretically expected, which contradicts the common generalization ability of state-of-the-art diffusion models, and thus calls for a deeper understanding. Looking into this, we first observe that memorization behaviors tend to occur on smaller-sized datasets, which motivates our definition of effective model memorization (EMM), a metric measuring the maximum size of training data at which a model approximates its theoretical optimum. Then, we quantify the impact of the influential factors on these memorization behaviors in terms of EMM, focusing primarily on data distribution, model configuration, and training procedure. Besides comprehensive empirical results identifying the influential factors, we surprisingly find that conditioning training data on uninformative random labels can significantly trigger the memorization in diffusion models. Our study holds practical significance for diffusion model users and offers clues to theoretical research in deep generative models. Code is available at \url{https://github.com/sail-sg/DiffMemorize}.\looseness=-1

% By incorporating such conditioning design, we show that more than $70\%$ of samples generated by diffusion models trained on $50\text{k}$ CIFAR-10 images are replicas of training data, an obvious contrast to the original $0\%$.

% What Makes Diffusion Models Memorize Training Data?

% An Empirical Study on Memorization in Diffusion Model

% 

% trained diffusion models can memorize more than 70\% of 50,000 CIFAR-10 images, a stark contrast to the nearly 0\% memorization observed in the original implementations.
% Remarkably, such conditioning design 

% By employing such conditioning design, we show that more than 70\% of samples generated by diffusion models trained on 50,000 CIFAR-10 images, a stark constrast to the nearly 0\%

% Title: Deciphering Memorization in Diffusion Models

% It has been found that the training objective (Denoised Score Matching, or DSM) of diffusion models has an optimal solution under the assumption of finite training samples \cite{karras2022elucidating, yi2023generalization}. The optimal diffusion models are non-parametric and show no generalization abilities as they can only reproduce training samples after backward process. The reason why parameterized diffusion models learnt via DSM objective have generalization abilities is still mysterious. In this research, we first demonstrate that with less training samples, parameterized diffusion models reproduce large number of training samples instead of generating novel samples. Afterwards, we build a formal analysis framework to explore factors affecting the generalization of parameterized diffusion models.
\end{abstract}
% \clearpage
% \vspace{-.2cm}
\section{Introduction}
% \vspace{-.2cm}
% Diffusion models, introduced by a series of pioneering works~\citep{sohl2015deep,song2019generative,ho2020denoising,song2021score}, now represent the state-of-the-art techniques in generative modeling. They are not only widely employed in unconditional/conditional image generation~\citep{dhariwal2021diffusion,karras2022elucidating, rombach2022high}, but also in audio/speech synthesis~\citep{huang2023make,kim2022guided}, graph generation~\citep{xu2022geodiff,vignac2022digress}, 3D content generation~\citep{poole2022dreamfusion,lin2023magic3d}, etc. Diffusion models can generate novel samples, which are not included in the training dataset~\citep{ho2020denoising, dhariwal2021diffusion}. Additionally, compared to Generative Adversarial Networks (GANs)~\citep{goodfellow2014generative}, the generated samples of diffusion models are typically more diverse.

In the last few years, diffusion models~\citep{sohl2015deep,song2019generative,ho2020denoising,song2021score} have achieved significant success across diverse domains of generative modeling, including image generation~\citep{dhariwal2021diffusion,karras2022elucidating}, text-to-image synthesis~\citep{rombach2022high,ramesh2022hierarchical}, audio / speech synthesis~\citep{kim2022guided,huang2023make}, graph generation~\citep{xu2022geodiff,vignac2022digress}, and 3D content generation~\citep{poole2022dreamfusion,lin2023magic3d}.
Substantial empirical evidence attests to the ability of diffusion models to generate diverse and novel high-quality samples~\citep{dhariwal2021diffusion,nichol2021improved,nichol2021glide}, underscoring their powerful capability of abstracting and comprehending the characteristics of the training data.

\begin{figure*}
\centering
     \subfigure[]{
         \includegraphics[width=0.46\textwidth]{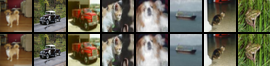}
         \label{fig_optimal}}
        \centering
    \subfigure[]{
         \includegraphics[width=0.46\textwidth]{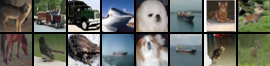}
         \label{fig_edm}}
         \centering
    \subfigure[]{
         \includegraphics[width=0.46\textwidth]{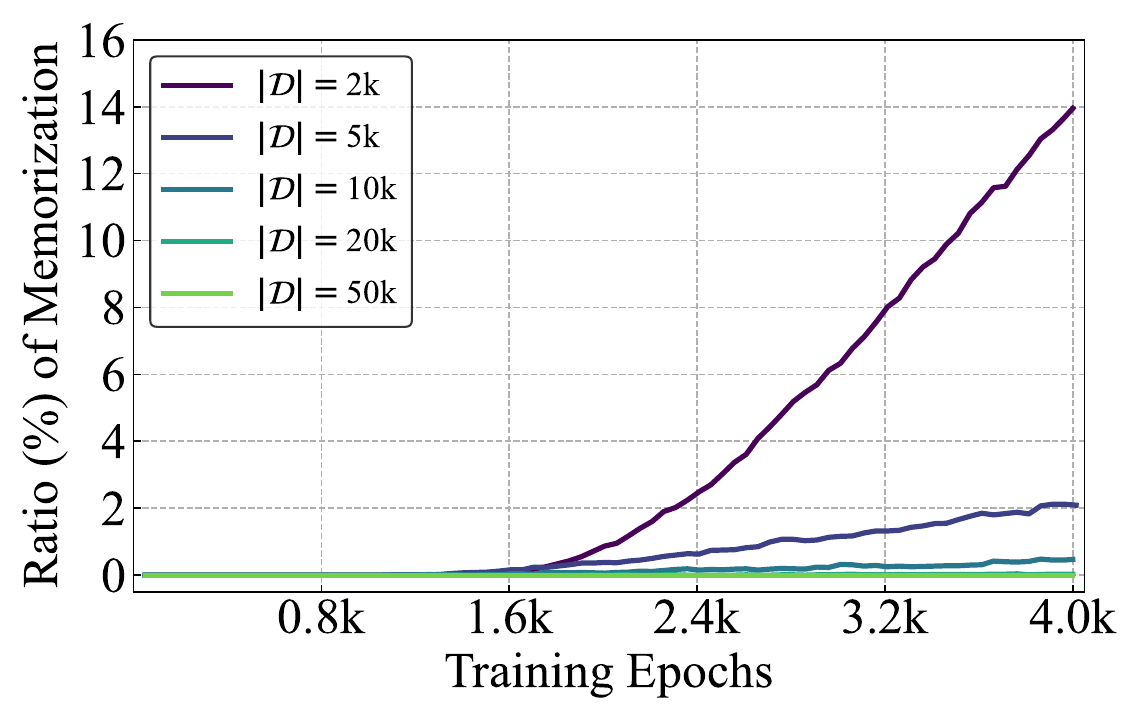}
         \label{fig_data_mem}}
         \centering
    \subfigure[]{
         \includegraphics[width=0.46\textwidth]{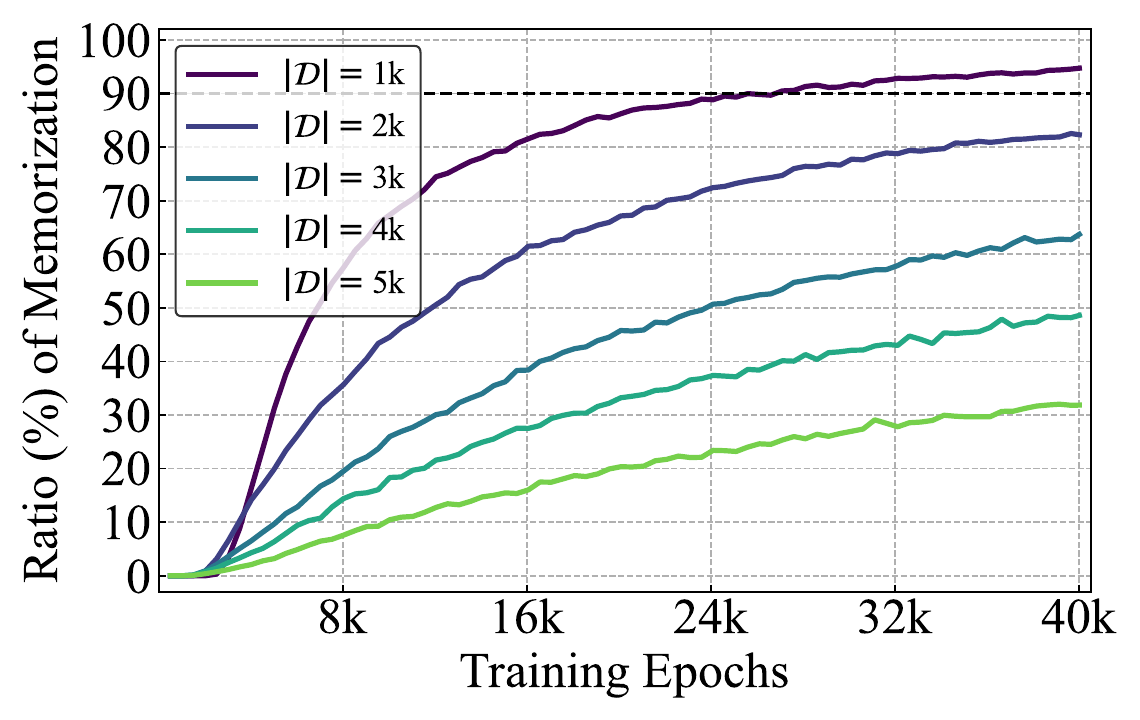}
         \label{fig_data_size}}
\vspace{-0.3cm}
\caption{Overall motivation. Generated images (\emph{top row}) and their $\ell_2$-nearest training samples in $\mathcal{D}$ (\emph{bottom row}) by (a) the theoretical optimum defined in~\eqref{optimal}; (b) EDM~\citep{karras2022elucidating}. \textbf{Memorization Ratios (\%)} of EDM models trained with different $|\mathcal{D}|$; (c) within 4k training epochs; (d) when extending to 40k training epochs.\looseness=-1}
\vspace{-0.3cm}
\end{figure*}

Diffusion models posit a forward diffusion process $\{z_{t}\}_{t}$ ($t\in[0,T]$) that gradually introduces Gaussian noise to a data point $x$, resulting in a transition distribution $q(z_t|x)=\mathcal{N}(z_t|\alpha_t x,\sigma^2_t\mathbf{I})$. The coefficients $\alpha_t$ and $\sigma_t$ are chosen such that the initial distribution $q_0(z_0)$ aligns with the data distribution $\mathbf{P}(x)$ while steering it towards an approximately Gaussian distribution $q_T(z_T)$.
Sampling from the data distribution $\mathbf{P}$ can then be achieved by reversing this process, for which a critical unknown term is the data score $\nabla_{z_t}\log q_t(z_t)$~\citep{song2021score}.
Diffusion models approximate the data scores with a score model $\boldsymbol{s}_\theta(z_t, t)$, which is typically learned via denoising score matching (DSM)~\citep{vincent2011connection}:
% \looseness=-1
\begin{equation}\label{eqn:dsm}
    \mathcal{J}_{\textrm{DSM}}(\theta)\triangleq\frac{1}{2N}\sum_{n=1}^{N}\mathbb{E}_{t, \epsilon}\left\|\boldsymbol{s}_{\theta}(\alpha_{t}x_{n}+\sigma_{t}\epsilon, t)+\frac{\epsilon}{\sigma_{t}}\right\|_{2}^{2}\textrm{,}
\end{equation}
given $\epsilon\sim\mathcal{N}(\mathbf{0},\mathbf{I})$ and a dataset of $N$ training samples $\mathcal{D}\triangleq\{x_{n}|x_n\sim \mathbf{P}(x)\}_{n=1}^{N}$.
% An intriguing fact is that
Interestingly, it is not difficult to identify the optimal solution of~\eqref{eqn:dsm} (assuming sufficient capacity of $\theta$, see proof in Appendix~\ref{appendix_optimal}):
% \looseness=-1
% \begin{equation}\label{optimal}
%     \boldsymbol{s}_\theta^*(z_t, t)=\left(\sum_{n'=1}^N\exp\Big(-\frac{\left\|\alpha_tx_{n'}-z_t\right\|_2^2}{2\sigma_t^2}\Big)\right)^{-1}\cdot\sum_{n=1}^N\exp\left(-\frac{\left\|\alpha_tx_n-z_t\right\|_2^2}{2\sigma_t^2}\right)\frac{\alpha_tx_n-z_t}{\sigma_{t}^2}\textrm{,}
% \end{equation}
% \begin{equation}\label{optimal}
%     \boldsymbol{s}_\theta^*(z_t, t)=\frac{\sum_{n=1}^N\exp\left(-\frac{\left\|\alpha_tx_n-z_t\right\|_2^2}{2\sigma_t^2}\right)\frac{\alpha_tx_n-z_t}{\sigma_{t}^2}}{\sum_{n'=1}^N\exp\Big(-\frac{\left\|\alpha_tx_{n'}-z_t\right\|_2^2}{2\sigma_t^2}\Big)}\textrm{,}
% \end{equation}
\begin{equation}\label{optimal}
    \boldsymbol{s}^*(z_t, t)=\left(\sum_{n'=1}^N\exp\Big(-\frac{\left\|\alpha_tx_{n'}-z_t\right\|_2^2}{2\sigma_t^2}\Big)\right)^{-1}\cdot\sum_{n=1}^N\exp\left(-\frac{\left\|\alpha_tx_n-z_t\right\|_2^2}{2\sigma_t^2}\right)\frac{\alpha_tx_n-z_t}{\sigma_{t}^2}\textrm{,}
\end{equation}
which, however, leads the reverse process towards the empirical data distribution, defined as $\widehat{\mathbf{P}}(x)=$ $\frac{1}{N}\sum_{n=1}^N\delta(x-x_n)$.
Consequently, the optimal score model in~\eqref{optimal} can only produce samples that replicate the training data, as shown in Figure~\ref{fig_optimal}, suggesting a \textit{memorization} behavior~\citep{van2021memorization}.\footnote{We also provide a theoretical analysis from the lens of backward process in Appendix~\ref{appendix_backward}.}
% This suggests that diffusion models trained with Eq.~(\ref{eqn:dsm}) would be prone to \textit{memorize} the training data, 
This evidently contradicts the typical generalization capability exhibited by state-of-the-art diffusion models such as EDM~\citep{karras2022elucidating}, as illustrated in Figure~\ref{fig_edm}.

% The intriguing gap between best-performing diffusion models~\citep{ho2020denoising,karras2022elucidating} and the theoretical optimum (Eq.~(\ref{optimal})) 
Such intriguing gap
prompts inquiries into (i) the conditions under which the learned diffusion models can faithfully approximate the optimum $\boldsymbol{s}^*$ (essentially showing memorization) and (ii) the influential factors governing memorization behaviors in diffusion models.
Besides a clear issue of potential adverse generalization performance~\citep{yoon2023diffusion},
% It then raises a concern of adverse generalization performance~\citep{yoon2023diffusion}.
% \citet{yoon2023diffusion} show that such memorization can cause adverse generalization performance.
% It therefore remains unknown why and when adverse generalization behavior~\citep{yoon2023diffusion} can be avoid when training diffusion models.
it further raises a crucial concern that diffusion models trained with~\eqref{eqn:dsm} might imperceptibly
% exhibit a potential toward memorization of 
memorize
the training data, exposing several risks such as privacy leakage~\citep{somepalli2023understanding} and copyright infringement~\citep{somepalli2023diffusion,zhao2023recipe,wang2024stronger}.
For example, \citet{carlini2023extracting} show that it is possible to extract a few training images from Stable Diffusion~\citep{rombach2022high}, substantiating a tangible hazard.\looseness=-1

% This observation raises a crucial concern that diffusion models trained with Eq.~(\ref{eqn:dsm}) may exhibit a potential toward \textit{memorization} of the training data, which exposes several risks such as privacy leakage~\citep{carlini2023extracting,somepalli2023understanding}, copyright infringement~\citep{zhao2023recipe} and adverse generalization behavior~\citep{yoon2023diffusion}.
% Furthermore, the intriguing disparity between practical diffusion models and the theoretically optimal score model (Eq.~(\ref{optimal})) prompts inquiries into when parameterized diffusion models approximate the optimal version and the influential factors affecting memorization behaviors in diffusion models.
% It therefore appeals a comprehensive understanding of the memorization behavior in diffusion models to enhancing
% Further, as 
% Such intriguing gap motivates us to study the memorization behavior of diffusion models.
% Investigating memorization helps us gain insights into the generalization abilities of diffusion models. Understanding how these models strike a balance between fitting the training data and avoiding overfitting is crucial for their practical utility

In response to these inquiries and concerns, this paper presents a comprehensive empirical study on memorization behavior in widely adopted diffusion models, including EDM~\citep{karras2022elucidating} and Stable Diffusion~\citep{rombach2022high}. We start with an analysis of EDM, noting that memorization tends to occur when trained on smaller-sized datasets, while remaining undetectable on larger datasets. This motivates our definition of \textit{effective model memorization} (EMM), a metric quantifying the maximum number of training data points (sampled from distribution $\mathbf{P}$) at which a diffusion model $\mathcal{M}$ demonstrates the similar memorization behavior as the theoretical optimum after the training procedure $\mathcal{T}$.
% We then present quantitative results that reveal the impact of critical factors on EMMs, considering the three facets of $P$, $\mathcal{M}$, and $\mathcal{T}$.
We then quantify the impact of critical factors on memorization in terms of EMM on the CIFAR-10, FFHQ~\citep{karras2019style}, and Imagenette~\citep{deng2009imagenet,somepalli2023understanding} datasets, considering the three facets of $\mathbf{P}$, $\mathcal{M}$, and $\mathcal{T}$.
Among all illuminating results, we surprisingly observe that the memorization can be triggered by conditioning training data on completely random and uninformative labels.
Specifically, using such conditioning design, we show that more than 65\% of samples generated by diffusion models trained on the 50k CIFAR-10 images are replicas of training data, an obvious contrast to the original 0\%.
Our study holds practical significance for diffusion model users and offers clues to theoretical research in deep generative models.\looseness=-1
\section{Memorization in Diffusion Models}\label{sec2}
\vspace{-.2cm}

We start by examining the memorization in the widely-adopted EDM~\citep{karras2022elucidating}, which is one of the state-of-the-art diffusion models for image generation.
To determine whether a generated image $x$ is a memorized replica from the training data $\mathcal{D}$, we adopt the criteria introduced in \citet{yoon2023diffusion}, which considers $x$ as memorized if its $\ell_2$ distance to the nearest neighbor is smaller than 1/3 of that to the second nearest neighbor in the training data. Here the factor 1/3 is an empirical threshold as it accurately aligns human perception of memorization~\citep{yoon2023diffusion}.
We train an EDM model on the CIFAR-10 dataset without applying data augmentation (to avoid any ambiguity regarding memorization) and evaluate the ratio of memorization among 10k generated images.
Remarkably, we observe a memorization ratio of zero throughout the entire training process, as illustrated by the bottom curve in Figure~\ref{fig_data_mem}.\looseness=-1

Intuitively, we hypothesize that the default configuration of EDM lacks the essential capacity to memorize the 50k training images in CIFAR-10, which motivates our exploration into whether expected memorization behavior will manifest when reducing the training dataset size.
In particular, we generate a sequence of training datasets with different sizes of \{20k, 10k, 5k, 2k\}, by sampling subsets from the original set of 50k CIFAR-10 training images.
We follow the default EDM training procedure~\citep{karras2022elucidating} with consistent training epochs on these smaller datasets.
As shown in Figure~\ref{fig_data_mem}, when $|\mathcal{D}|=\,$20k or 10k, the memorization ratio remains close to zero. However, upon further reducing the training dataset size to $|\mathcal{D}|=\,$5k or 2k, the EDM models exhibit noticeable memorization. This observation indicates the substantial impact of training dataset size on memorization.
Additionally, we notice that the memorization ratio increases with more training epochs. To observe this, we extend the training duration to 40k epochs, ten times longer than that in \citet{karras2022elucidating}. In Figure~\ref{fig_data_size}, when $|\mathcal{D}|=\,$1k, the model achieves over 90\% memorization. However, even with 40k training epochs, the diffusion model still struggles to replicate a large portion of training samples when $|\mathcal{D}|=\,$5k. 

% Intuitively, we hypothesize that it is intractable for EDM to memorize all $50\text{k}$ training images in $\mathcal{D}$ under the current experimental setting, which motivates us to reduce the size of training dataset $\mathcal{D}$, whose samples are from the same real data distribution $P$. Specifically, we sample a series of training datasets $\mathcal{D}$ from $P$ with varying sizes $\{20\text{k}, 10\text{k}, 5\text{k}, 2\text{k}\}$. Afterwards, we follow the same training procedure in \cite{karras2022elucidating} and keep the overall training epochs constant. As depicted in Figure~\ref{fig_data_mem}, when $|\mathcal{D}|=20\text{k}$ or $10\text{k}$, the memorization ratio is still close to zero. However, when further reducing the size of training dataset, e.g., $|\mathcal{D}|=5\text{k}$ or $2\text{k}$, the trained EDM models demonstrate evident memorization. This observation indicates that the size of training dataset plays a significant role in memorization. Additionally, it is noted that the memorization ratio increases with the progression of training epochs. Therefore, we further train the diffusion models until $40\text{k}$ epochs, which are $10$ times of that in \citet{karras2022elucidating}, to observe the ratio of memorization. As present in Figure~\ref{fig_data_size}, when $|\mathcal{D}|=1\text{k}$, the trained EDM model achieves a more than $90\%$ ratio of memorization. Even for $40\text{k}$ training epochs, we observe that it is still difficult for diffusion models to largely replicate training samples of $\mathcal{D}$ with $5\text{k}$ samples.

Based on our findings, we seek to quantify the maximum dataset size at which diffusion models demonstrate behavior similar to the theoretical optimum, which is crucial in understanding memorization behavior.
To formalize this notion, considering a data distribution $\mathbf{P}$, a diffusion model configuration $\mathcal{M}$, and a training procedure $\mathcal{T}$, we introduce the concept of \emph{effective model memorization} with the definition as follows inspired by \citet{yoon2023diffusion}.\looseness=-1 

% Based on the above exploration, given a real data distribution $P$, a diffusion model configuration ${\color{orange}{\mathcal{M}}}$, and a training procedure ${\color{mydarkgreen}{\mathcal{T}}}$, we wonder under which size of training dataset $\mathcal{D}\sim P$, the trained diffusion model can demonstrate similar memorization as its theoretical optimum. Inspired by \citet{nakkiran2020deep}, we define the notation of \textit{effective model memorization} w.r.t. $P$, ${\color{orange}{\mathcal{M}}}$, and ${\color{mydarkgreen}{\mathcal{T}}}$, as the maximum size of training dataset $\mathcal{D}$ (sampled from $P$) on which ${\color{orange}{\mathcal{M}}}$ can only replicate training data. \looseness=-1

\begin{definition}[Effective model memorization]\label{def1}
The effective model memorization (EMM) with respect to $\mathbf{P}$, $\mathcal{M}$, $\mathcal{T}$ and parameter $\zeta > 0$, is defined as
% \begin{equation}
%     \textrm{EMM}(\mathcal{M}, \mathcal{D}, \mathcal{T}) = \max_{|\hat{\mathcal{D}}|} \left\{\hat{\mathcal{D}}\subset\mathcal{D}\textrm{,}\,\,\mathcal{R}_{\textrm{Mem}}(\mathcal{M}, \hat{\mathcal{D}}, \mathcal{T}) \geq 1-\zeta \right\}\textrm{,}
% \end{equation}
\begin{equation}
    \textrm{EMM}_{\zeta}(\mathbf{P}, \mathcal{M}, \mathcal{T})
    = \max_{N} \left\{\mathbb{E}[\mathcal{R}_{\textrm{Mem}}(\mathcal{D}, 
 \mathcal{M}, \mathcal{T})] \geq 1-\zeta \Big|\mathcal{D}\sim \mathbf{P},|\mathcal{D}|=N \right\}\textrm{,}
\end{equation}
where $\mathcal{R}_{\textrm{Mem}}$ refers to the ratio of memorization.
\end{definition}

EMM indicates the condition under which the learned diffusion model approximates the theoretical optimum and reveals how $\mathbf{P}$, $\mathcal{M}$, and $\mathcal{T}$ interact and affect memorization. Our definition assumes that a higher memorization ratio tends to occur on smaller-sized training datasets, which is stated as\looseness=-1

% EMM only serves as a tool to facilitate our analysis on memorization.

\begin{hypothesis}\label{hyp1}
Given $\mathcal{M}$, $\mathcal{T}$ and two training datasets $\mathcal{D}_1$ and $\mathcal{D}_2$, both of which are sampled from the same data distribution $\mathbf{P}$, the ratio of memorization satisfies
\begin{equation}
    \mathcal{R}_{\textrm{Mem}}(\mathcal{D}_1, \mathcal{M}, \mathcal{T}) \geq \mathcal{R}_{\textrm{Mem}}(\mathcal{D}_2, \mathcal{M}, \mathcal{T}) \,\,\,\,\textrm{if}\,\,\mathcal{D}_1\subset\mathcal{D}_2\,\,\textrm{and}\,\, \mathcal{D}_1,\mathcal{D}_2\sim \mathbf{P}\textrm{.}
\end{equation}
\end{hypothesis}

Based on Hypothesis~\ref{hyp1}, we provide a viable way to estimate EMM. Specifically, we sample a series of training datasets $\mathcal{D}_1,\mathcal{D}_2,...,$ with different sizes from the data distribution $\mathbf{P}$, and then train diffusion models with configuration $\mathcal{M}$ following the training procedure $\mathcal{T}$. Afterwards, we evaluate the ratio of memorization $\mathcal{R}_{\textrm{Mem}}(\mathcal{D}_1, \mathcal{M}, \mathcal{T}), \mathcal{R}_{\textrm{Mem}}(\mathcal{D}_2, \mathcal{M}, \mathcal{T}),...,$ and then determine the size of training dataset $\mathcal{D}$ which meets that $\mathcal{R}_{\textrm{Mem}}(\mathcal{D}, \mathcal{M}, \mathcal{T})\approx1-\zeta$. We note that it is computational intractable to determine the accurate value of EMM. Therefore, we interpolate the value of EMM based on two consecutive sampled datasets $\mathcal{D}_i$, $\mathcal{D}_{i+1}$ that $\mathcal{R}_{\textrm{Mem}}(\mathcal{D}_{i}, \mathcal{M}, \mathcal{T})>1-\zeta$ and $\mathcal{R}_{\textrm{Mem}}(\mathcal{D}_{i+1}, \mathcal{M}, \mathcal{T})<1-\zeta$. Therefore, this study is formulated as how the above three factors $\mathbf{P}$, $\mathcal{M}$, and $\mathcal{T}$ affect the measurement of EMM. There is no principled way to select the value of $\zeta$, so we set it as 0.1 based on our experiments in Figure~\ref{fig_data_size} throughout our study. We include an analysis on the sensitivity of $\zeta$ in Appendix~\ref{sensitivity}.

We set CIFAR-10 as our default dataset, and the basic experimental setup is introduced in Appendix~\ref{appendix_setup}, which is a well-adopted recipe for diffusion models. We highlight that compared to \citet{karras2022elucidating}, we run 10 times the number of training epochs. The experiments related to FFHQ~\citep{karras2019style} and Imagenette~\citep{deng2009imagenet,somepalli2023understanding} are shown in Appendix~\ref{appendix ffhq} and Section~\ref{sec cond}, respectively. For evaluation, we also consider the image quality metric, e.g., Fréchet Inception Distance (FID)~\citep{heusel2017gans} in Appendix~\ref{appendix fid}, and alternative memorization metric in Appendix~\ref{appendix criteria}. It is worth mentioning that when diffusion models memorize a significant proportion of training data, the metrics related to image quality (including FID) are also close to optimal. This intuition is in line with our experimental results in Appendix~\ref{appendix fid}. Moreover, when considering alternative memorization metrics, the relationship of various factors and memorization remains consistent.\looseness=-1

\begin{figure*}
\subfigure[Varying data dimensions]{\includegraphics[width=0.32\textwidth]{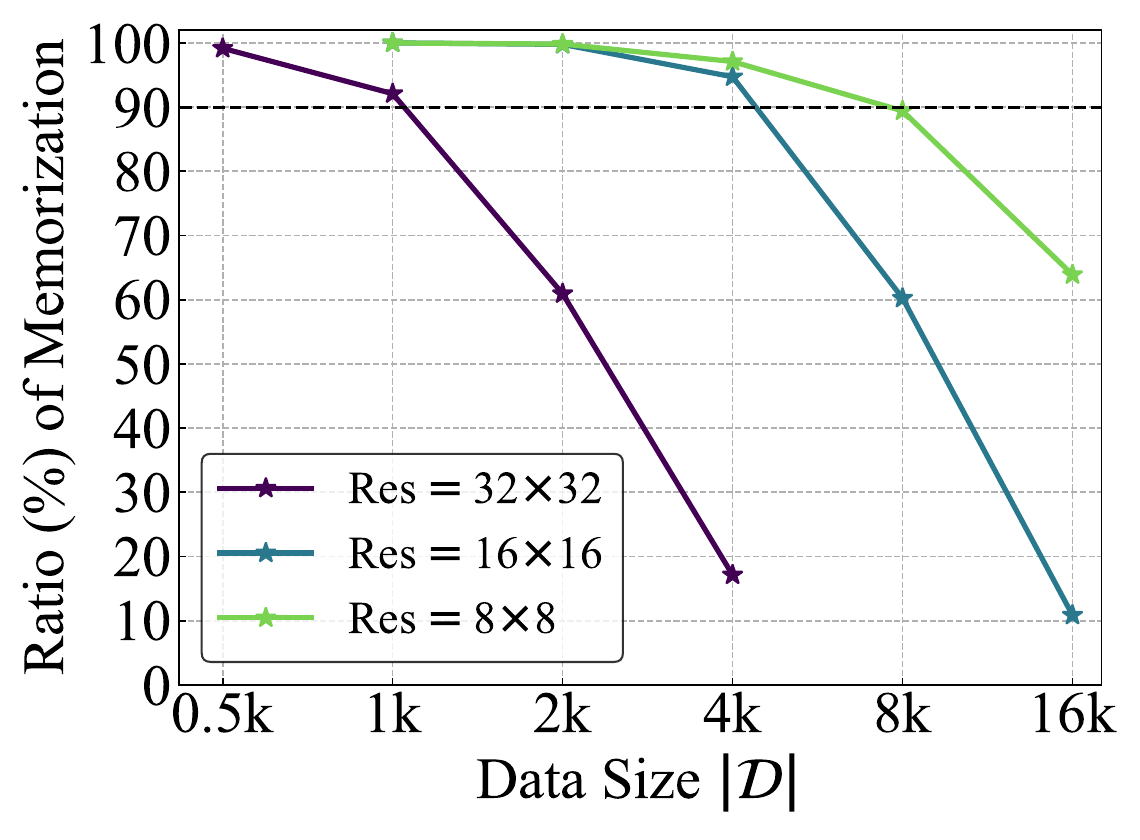}\label{fig_data_dim}}
\subfigure[Varying inter-diversity]{\includegraphics[width=0.32\textwidth]{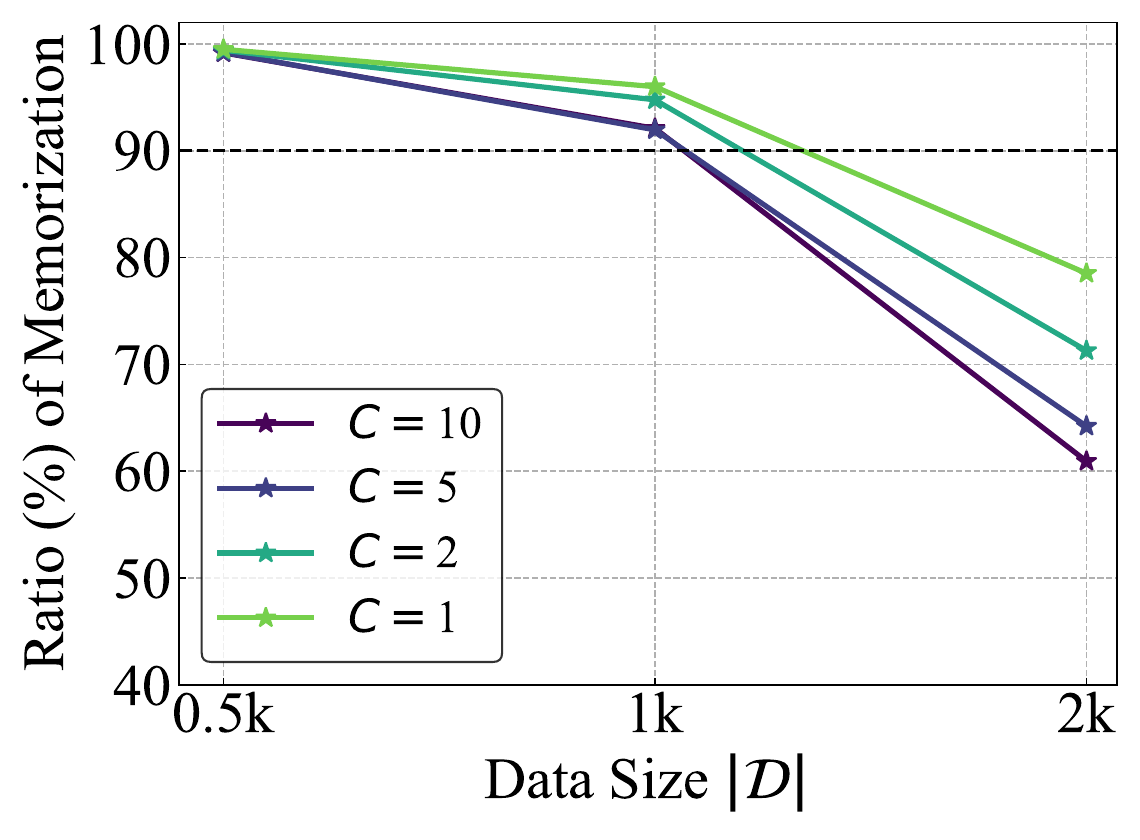}\label{fig_data_inter}}
\subfigure[Varying intra-diversity]{\includegraphics[width=0.32\textwidth]{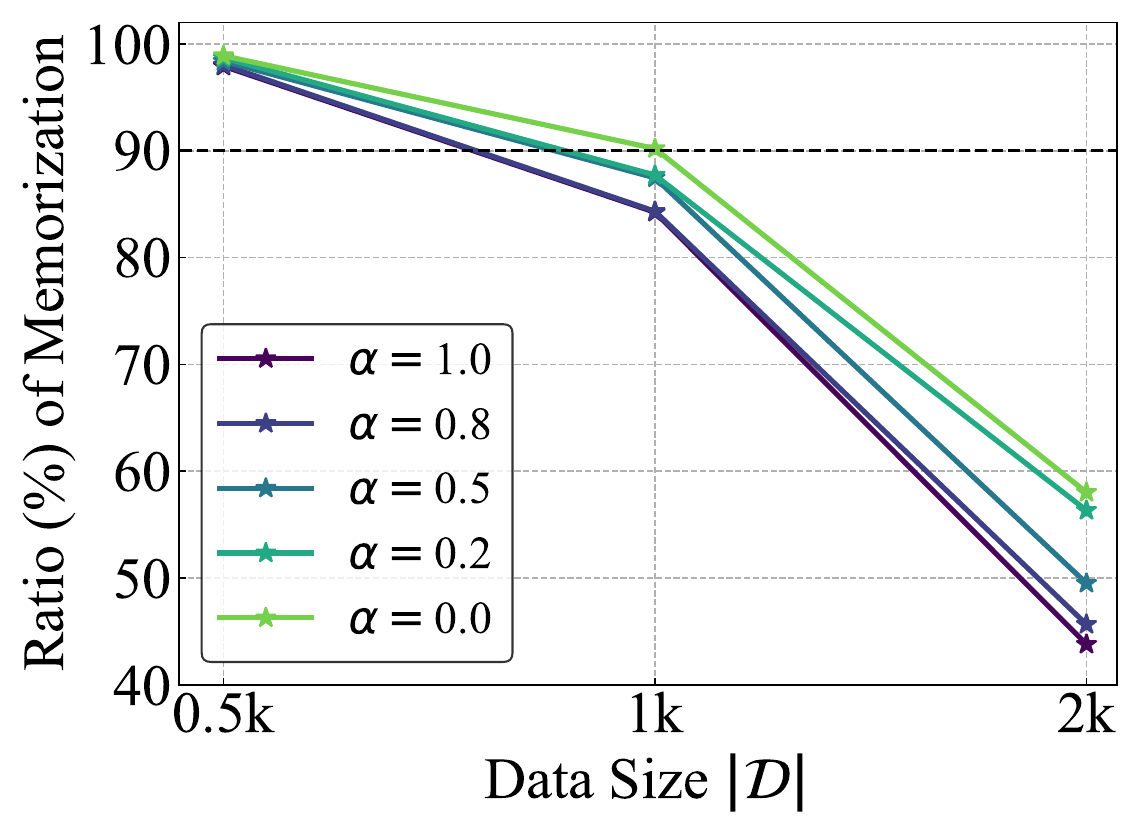}\label{fig_data_intra}}
\vspace{-0.3cm}
        \caption{\textbf{Memorization ratios (\%)} of diffusion models on CIFAR-10 under different factors of data distribution $\mathbf{P}$. The intersections of \emph{dashed line (90\%)} and different curves are the estimates of EMMs.\looseness=-1}
\vspace{-0.3cm}
\end{figure*}

\vspace{-.2cm}
\section{Data Distribution \texorpdfstring{$\mathbf{P}$}{TEXT}} 
\vspace{-.2cm}

In the preceding section, we have illustrated the substantial impact of the size of training data on the memorization in diffusion models and how we evaluate the value of effective model memorization (EMM).
We now proceed to investigate the influence of specific attributes of the data distribution $\mathbf{P}$ on the EMM, focusing primarily on the dimensions and diversity of the data.
We keep both the model configuration $\mathcal{M}$ and the training procedure $\mathcal{T}$ fixed throughout this section.\looseness=-1

\vspace{-.2cm}
\subsection{Data Dimension} 
\vspace{-.2cm}
% Belonging to the family of likelihood-based generative models, diffusion models encounter challenges when fitting high-resolution images~\citep{rombach2022high}. This observation motivates our hypothesis that data dimensionality plays a significant role in the memorization process. Therefore, to validate this hypothesis, we conduct experiments where we modify the original spatial dimension of the CIFAR-10 dataset from 32$\times$32 to 16$\times$16 and 8$\times$8, subsequently training diffusion models. Initially, our diffusion models feature self-attention layers at a resolution of 16$\times$16 to match the original spatial resolution. For input resolutions of 16$\times$16 and 8$\times$8, we adjusted the attention resolutions to 8$\times$8 and 4$\times$4, respectively. Notably, this modification does not alter the number of trainable parameters.

As likelihood-based generative models, diffusion models could face challenges when fitting high-resolution images, stemming from their mode-covering behavior as noted by \citet{rombach2022high}. To explore the influence of data dimensionality on the memorization tendencies of diffusion models, we evaluate the EMMs on CIFAR-10 at various resolutions: 32$\times$32, 16$\times$16 and 8$\times$8, where the latter two are obtained by downsampling. Note that the U-Net~\citep{ronneberger2015u} seamlessly accommodates inputs of different resolutions, requiring no modification of the model configuration $\mathcal{M}$.\looseness=-1

% As depicted in Figure~\ref{fig_data_dim}, our findings reveal that the effective model memorization (EMM) for the input spatial dimension of 32$\times$32 is approximately 1k. For the 16$\times$16 dimension, EMM slightly exceeds 4k, while for the 8$\times$8 dimension, it reaches approximately 8k. Furthermore, when the training dataset size $|\mathcal{D}|$ is set to 16k, the memorization ratio of our diffusion model exceeds 60\% for 8$\times$8 training images. Thus, our results provide strong support for the significant impact of data dimensionality on the memorization process of diffusion models. 

For each resolution, we sample a series of training datasets $\mathcal{D}$ with varying sizes and evaluate the memorization ratios of trained diffusion models. As illustrated in Figure~\ref{fig_data_dim}, we estimate the EMM for each resolution by determining the intersection between the line of 90\% memorization (\textit{dashed line}) and the memorization curve. The results reveal natural insights into the EMM with varying input dimensions. Specifically, for the 32$\times$32 input resolution, we observe an EMM of approximately 1k. Transitioning to a 16$\times$16 resolution, the EMM slightly surpasses 4k, while for the 8$\times$8 resolution, it reaches approximately 8k. Furthermore, even for $|\mathcal{D}|=\,$16k, the ratio of memorization still exceeds 60\% when trained on 8$\times$8 CIFAR-10 images.
These results underscore the profound impact of data dimensionality on the memorization within diffusion models. 

% We also include the results on the FFHQ dataset in Appendix~\ref{ffhq dim}.

\vspace{-.2cm}
\subsection{Data Diversity}
\vspace{-.2cm}
% \paragraph{Inter-class diversity} We explore the influence of inter-class diversity, defined as the number of classes present in the training dataset, on Effective Model Memorization (EMM). Specifically, we vary the number of classes $C$ within a fixed-sized training dataset, considering values from the set $\{1, 2, 5, 10\}$. It is noted that the number of images per class remains consistent across these experiments. To account for potential variations in the impact of training images from different classes on memorization, we ensure that the set of classes in the training dataset with fewer classes is a strict subset of the set with more classes. As depicted in Figure~\ref{fig_data_inter}, our findings indicate that as the number of classes increases, diffusion models tend to exhibit a lower memorization ratio. Notably, this effect is subtle, as evidenced by the nearly identical EMMs observed for $C=5$ and $C=10$.

% Data diversity reflects the number of modes in the data distribution ${\color{blue}{\mathbf{P}}}$.\cd{not true} We explore the influence of following two aspects of data diversity on memorization. 

\textbf{Number of Classes.} We consider four different data distributions by selecting $C\in\{\text{1}, \text{2}, \text{5}, \text{10}\}$ classes of images from CIFAR-10 and then evaluate the EMMs. While varying the data size $|\mathcal{D}|$ during probing the EMMs, we ensure that each class contains an equal share of $|\mathcal{D}|/C$ data instances.
% we first vary the number of classes in the training dataset, denoted as $C$. We select $C$ from the set $\{1, 2, 5, 10\}$, ensuring that each class contains an equal share of $|\mathcal{D}|/C$ data instances.
The results of EMMs for different $C$ are shown in Figure~\ref{fig_data_inter}.
We find that as the number of classes increases, diffusion models tend to exhibit a lower memorization ratio and a lower EMM, which is consistent with the intuition that diverse data is harder to be memorized. Note, however, that this effect is subtle, as evidenced by the nearly identical EMMs observed for $C=\,$5 and $C=\,$10.

% \paragraph{Intra-class diversity} We also investigate the impact of intra-class diversity, which characterizes the variations within each class, on Effective Model Memorization (EMM). To control this diversity, we conduct experiments by mixing images from two datasets, CIFAR-10 and ImageNet~\citep{deng2009imagenet}, wuth a focus on a common class, ``Dog'' to eliminate the influence of inter-class diversity. Upon analyzing ImageNet, we identify that there are 123 sub-classes within the class ``Dog'', suggesting that ImageNet typically exhibits higher intra-class diversity compared to CIFAR-10. To manipulate the mixture of CIFAR-10 and ImageNet data, we introduce an interpolation ratio denoted as $\alpha$, where $0.0\leq\alpha\leq 1.0$ represents the proportion of ImageNet data in the constructed training dataset. In accordance to our assumption, a larger $\alpha$ corresponds to lower intra-class diversity. It is worth noting that the data from ImageNet are scaled to a resolution of $32\times32$ for training.  As illustrated in Figure~\ref{fig_data_intra}, an increased mixture of ImageNet data corresponds to higher EMM in the trained diffusion models. Similar to our experiments concerning inter-class diversity, these results reaffirm that intra-class diversity also makes limited contributions to memorization. 
\textbf{Intra-Class Diversity.} We also explore the impact of intra-class diversity, which measures variations within individual classes. We conduct experiments with $C=\,$1, where only the \textit{Dog} class of CIFAR-10 is used. To control this diversity, we gradually blend images (scaled to a resolution of 32$\times$32) from the \textit{Dog} class in ImageNet~\citep{deng2009imagenet} into the \textit{Dog} class of CIFAR-10. We introduce an interpolation ratio, denoted as $\alpha$ ($\alpha\in[0,1]$), representing the proportion of ImageNet data in the constructed training dataset. Notably, ImageNet's \textit{Dog} class contains 123 sub-classes, indicating higher intra-class diversity compared to CIFAR-10. Consequently, a larger $\alpha$ corresponds to higher intra-class diversity. As shown in Figure~\ref{fig_data_intra}, an increased blend of ImageNet data results in slightly lower EMM in the trained diffusion models. Similar to our experiments concerning the number of classes, these results reaffirm that diversity contributes limitedly to memorization.\looseness=-1

\vspace{-.2cm}
\section{Diffusion Model Configuration \texorpdfstring{$\mathcal{M}$}{TEXT}}
\vspace{-.2cm}
% Apart from the empirical data, we hypothesize the configuration of diffusion model ${\color{orange}{\mathcal{M}}}$ also influences the Effective Model Memorization (EMM), including model size, skip connections in UNet, time embedding approach, model parameterization, and whether to enable class condition input.

In this section, we study the influence of different model configurations on the memorization tendencies of diffusion models. Our evaluation encompasses several aspects of model design, such as model size (width and depth), the way to incorporate time embedding, and the presence of skip connections in the U-Net. Similar to previous sections, we probe the EMMs by training multiple models on training data of different sizes from the same data distribution $\mathbf{P}$ (CIFAR-10), while keeping the training procedure $\mathcal{T}$ fixed.\looseness=-1

\vspace{-.2cm}
\subsection{Model Size}\label{model_size}
\vspace{-.2cm}
% Diffusion models are constructed using U-Net architecture~\citep{ronneberger2015u}. To scale the model size, two distinct approaches are considered: (i) by increasing the channel multiplier, the model's width is augmented. (ii) alternatively, the number of residual blocks per resolution can be increased, as demonstrated by \citet{song2021score}, thereby augmenting the model's depth. These two strategies represent the focus of our investigation into the model scalability and memorization of diffusion models.

Diffusion models are usually constructed using the U-Net architecture~\citep{ronneberger2015u}.
We explore the influence of model size on memorization using two distinct approaches.
First, we increase the channel multiplier, thereby augmenting the width of the model. Alternatively, we raise the number of residual blocks per resolution, as demonstrated by \citep{song2021score}, to increase the model depth.\looseness=-1

% \begin{figure}
%      \begin{subfigure}[t]{0.32\textwidth}
%          \centering
%          \includegraphics[width=\textwidth]{figures/memorize_model_width.pdf}
%          \vspace{-.6cm}
%          \caption{}
%          % \vspace{-.2cm}
%          \label{fig_model_width}
%      \end{subfigure}
%      \hfill
%      \begin{subfigure}[t]{0.32\textwidth}
%          \centering
%          \includegraphics[width=\textwidth]{figures/memorize_model_depth.pdf}
%          \vspace{-.6cm}
%          \caption{}
%          \label{fig_model_depth}
%      \end{subfigure}
%      \hfill
%      \begin{subfigure}[t]{0.32\textwidth}
%          \centering
%          \includegraphics[width=\textwidth]{figures/memorize_model_embed.pdf}
%          \vspace{-.6cm}
%          \caption{}
%          \label{fig_model_embed}
%      \end{subfigure}
%      \vspace{-.3cm}
%         \caption{Memorization ratios (\%) of different (a) model widths; (b) model depths; (c) time embeddings.}
%     \vspace{-.6cm}
% \end{figure}

\begin{figure*}[t]
\centering
\subfigure[Varying model widths]{\includegraphics[width=0.32\textwidth]{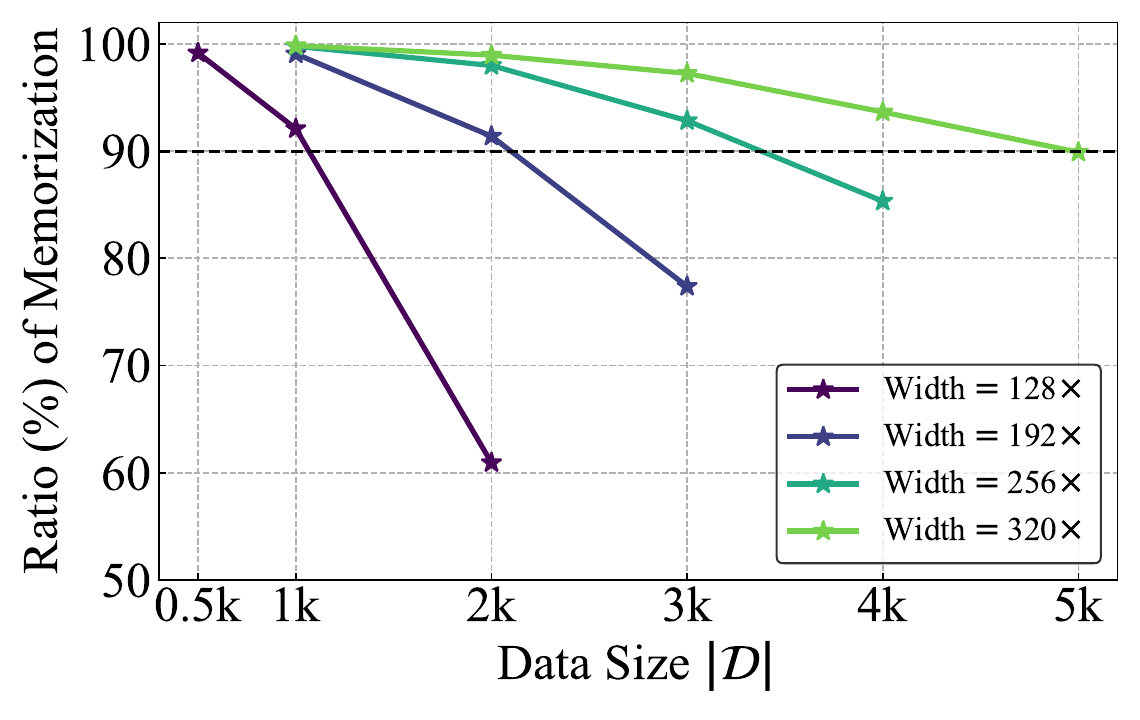}\label{fig_model_width}}
\subfigure[Varying model depths]{\includegraphics[width=0.32\textwidth]{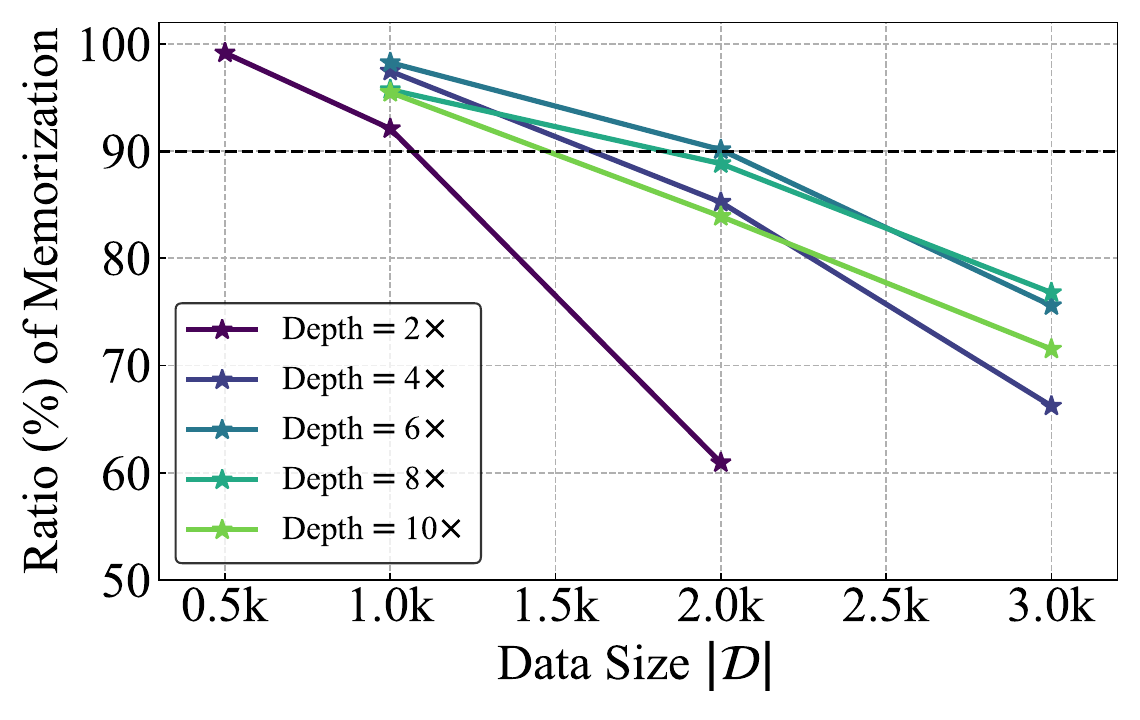}\label{fig_model_depth}}
% \subfigure[Model blocks]{\includegraphics[width=0.45\textwidth]{figures/memorize_model_block.pdf}\label{fig_model_block}}
\subfigure[Varying time embeddings]{\includegraphics[width=0.32\textwidth]{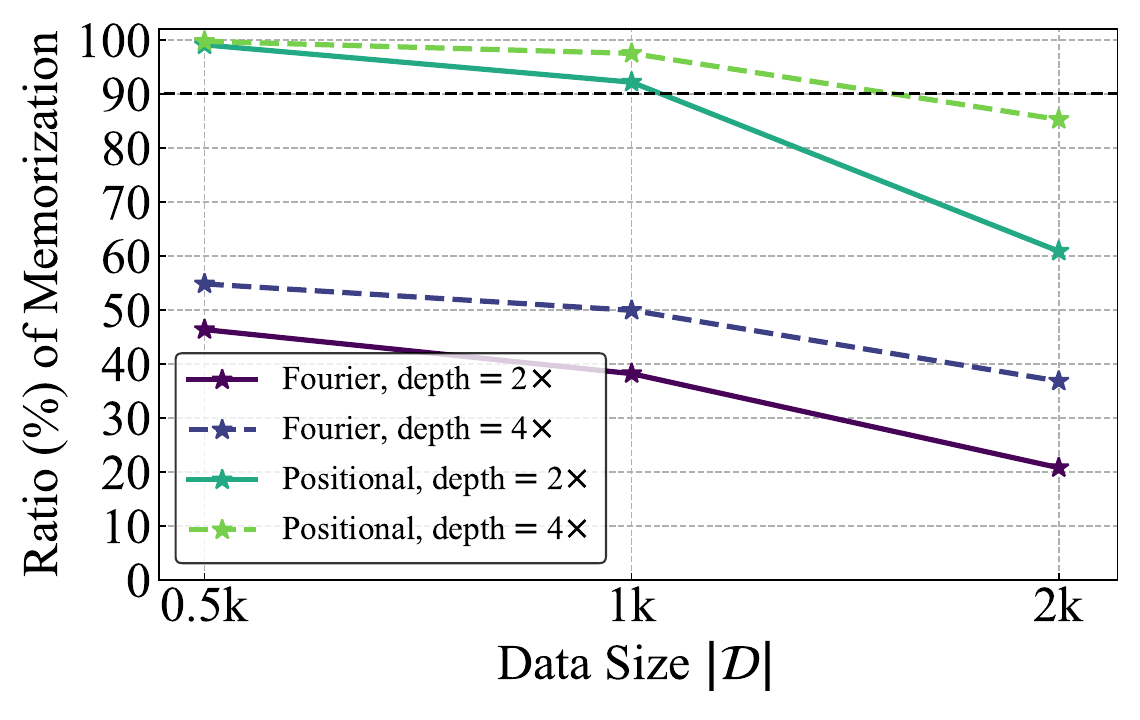}\label{fig_model_embed}}
\vspace{-0.3cm}
\caption{\textbf{Memorization ratios (\%)} of diffusion models on CIFAR-10 under different factors of model configuration $\mathcal{M}$.\looseness=-1}
\vspace{-0.3cm}
\end{figure*}

% \paragraph{Model Width.}In our experiments, we explore different channel multipliers $\{128, 192, 2456, 320\}$, while keeping the number of residual blocks per resolution fixed at 2. As illustrated in Figure~\ref{fig_model_width}, it is evident that as the width increases in diffusion models, the Effective Model Memorization (EMM) exhibits a monotonic rise. Notably, when scaling the channel multiplier to 320, the EMM reaches approximately 5k, representing a four times increase compared to the EMM when the channel multiplier is set to 128. These results underscore the direct relationship between model width and memorization in diffusion models.

\textbf{Model width.} We explore different channel multipliers, specifically $\{\text{128}, \text{192}, \text{256}, \text{320}\}$, while keeping the number of residual blocks per resolution fixed at 2. As illustrated in Figure~\ref{fig_model_width}, it is evident that as the model width increases in diffusion models, the EMMs exhibit a monotonic rise. Notably, scaling the channel multiplier to 320 yields an EMM of approximately 5k, representing a four times increase compared to the EMM observed with a channel multiplier set at 128. These results show the direct relationship between model width and memorization in diffusion models.

% We vary the number of residual blocks per resolution, exploring values ranged from 2 to 12, while keeping the channel multiplier fixed at 128. In contrast to the scenario of varying model width, modifying model depth yields non-monotonic effects on memorization. As present in Figure~\ref{fig_model_depth}, we observe that the EMM initially increases as we scale the number of residual blocks per resolution from 2 to 6. However, when we further increase the model depth, the EMM starts to decrease. To further investigate this non-monotonic behavior, we visualize the relationship between the ratio of memorization and the number of residual blocks per resolution in Figure~\ref{fig_model_block}. For different training data sizes $|\mathcal{D}|=$1k and $|\mathcal{D}|=2$k, the ratio of memorization reaches peaks at about 6 residual blocks per resolution.

\textbf{Model depth.} We vary model depth by adjusting the number of residual blocks per resolution (ranging from 2 to 12), while maintaining a constant channel multiplier of 128. In contrast to the scenario of varying model width, modifying model depth yields non-monotonic effects on memorization.
As present in Figure~\ref{fig_model_depth}, the EMM initially increases as we scale the number of residual blocks per resolution from 2 to 6. However, when further increasing the model depth, the EMM starts to decrease. This non-monotonic trend is further shown in Figure~\ref{fig_model_depth2}.\looseness=-1

% These results demonstrate the non-monotonic relationship between model depth and memorization in diffusion models.

% To further show this non-monotonic behavior, we visualize the relationship between the ratio of memorization and the number of residual blocks per resolution in Figure~\ref{fig_model_block}. For training data with the sizes $|\hat{\mathcal{D}}|=\,$1k and 2k, the ratio of memorization reaches the peak at about 6 residual blocks per resolution.\looseness=-1

% Figure~\ref{fig_model_block}. For training data with the sizes $|\hat{\mathcal{D}}|=1\text{k}$ and $2\text{k}$, the ratio of memorization reaches the peak at about $6$ residual blocks per resolution.\looseness=-1

We assess the above two approaches for scaling the model size of diffusion models. Model size scales linearly when augmenting model depth but quadratically when increasing model width. With a channel multiplier set to 320, the diffusion model encompasses roughly 219M trainable parameters, yielding an EMM of approximately 5k. In contrast, with a residual block number per resolution set to 12, the model contains approximately 138M parameters, resulting in an EMM only ranging between 1k and 2k. Further increasing the model depth may encounter failures in training. Consequently, scaling model width emerges as a more viable approach for increasing the memorization ratio of diffusion models. We conduct more experiments in Appendix~\ref{appendix_size} to further confirm our conclusions.\looseness=-1

% Specifically, we compute the number of model parameters of each scenario and compare the generalization of diffusion models. The model size scales linearly with the increase of model depth while scales quadratically with the increase of model width. As shown in Figure \ref{fig_size}, with the scaling of model parameters, the training data copying of diffusion models scales not monotonically when scaling model depth. However, when scaling model width, the diffusion models can generate more samples which are similar to training data. 

\vspace{-.2cm}
\subsection{Time Embedding}
\vspace{-.2cm}
% In our experimental setup, we employ DDPM++, which incorporates positional embedding~\citep{vaswani2017attention} to encode the time step, as the architecture for our diffusion model.  In addition to positional embedding, \citet{song2021score} leverages the random fourier features, introduced in \citet{tancik2020fourier}, in the diffusion models. In our experiments, we replace the original positional embedding with the fourier features to assess its impact on memorization.  As depicted in Figure~\ref{fig_model_embed}, we consider the diffusion models with 2 and 4 residual blocks per resolution. We observe a significant decrease in the ratio of memorization after the adoption of fourier features in DDPM++, highlighting the noteworthy effect of time embedding on model memorization. 

In our experimental setup (see Appendix~\ref{appendix_setup}), we employ the model architecture of DDPM++~\citep{song2021score,karras2022elucidating}, which incorporates positional embedding~\citep{vaswani2017attention} to encode the diffusion time step.  In addition to positional embedding, \citet{song2021score} used random Fourier features~\citep{tancik2020fourier} in their NCSN++ models. Therefore, we conduct experiments to test both time embedding methods and assess their impact on memorization.  As depicted in Figure~\ref{fig_model_embed}, to further support our conclusion, we consider two diffusion models with 2 and 4 residual blocks per resolution. We observe a significant decrease in the memorization ratio (and thus EMM) when using the Fourier features in DDPM++, highlighting the noteworthy effect of time embedding on memorization. The same trend is observed on FFHQ (see Appendix~\ref{ffhq embedding}).\looseness=-1

% \begin{figure}[t]
% \centering
% \subfigure[Varying resolutions in DDPM++]{\includegraphics[width=0.23\textwidth]{figures/memorize_vp_skip1.pdf}\label{fig_model_skip1_vp}}
% \subfigure[Varying locations in DDPM++]{\includegraphics[width=0.23\textwidth]{figures/memorize_vp_skip2.pdf}\label{fig_model_skip2_vp}}
% \subfigure[Varying resolutions in NCSN++]{\includegraphics[width=0.23\textwidth]{figures/memorize_ve_skip1.pdf}\label{fig_model_skip1_ve}}
% \subfigure[Varying locations in NCSN++]{\includegraphics[width=0.23\textwidth]{figures/memorize_ve_skip2.pdf}\label{fig_model_skip2_ve}}
% \vspace{-0.3cm}
% \caption{\textbf{Memorization ratio (\%)} when retaining (a) skip connections of certain spatial resolution for DDPM++; (b) single skip connection at different locations for DDPM++; (c) skip connections of certain spatial resolution for NCSN++; (d) single skip connection at different locations for NCSN++.}
% \vspace{-0.3cm}
% \end{figure}

\vspace{-.2cm}
\subsection{Skip Connections}
\vspace{-.2cm}
We investigate the impact of skip connections, which are known for their significance in the success of U-Net~\citep{ronneberger2015u}, on the memorization of diffusion models. Specifically, in our experimental setup (see Appendix~\ref{appendix_setup}), the number of skip connections is $3(n+1)$, where $n$ corresponds to the number of residual blocks per resolution. For each resolution, namely, 32$\times$32, 16$\times$16,  8$\times$8, we establish $n+1$ skip connections that bridge the output of the encoder to the input of the decoder. To conduct our experiments, we set the size of training dataset $|\mathcal{D}|$ to 1k, and the value of $n$ to 2, resulting in a total of 9 skip connections in our model architecture. 

Initially, we explore the influence of the quantity of skip connections on memorization. Intriguingly, our observations reveal that even with a limited number of skip connections, the trained diffusion models are capable of maintaining a memorization ratio equivalent to that achieved with full skip connections, as demonstrated in Appendix~\ref{appendix_skip}. This underscores the role of skip connection sparsity as a significant factor influencing memorization in diffusion models, prompting our exploration of the effects of specific skip connections. In the following experiments, we consider both DDPM++ and NCSN++ architectures in \citet{song2021score,karras2022elucidating}. Utilizing full skip connections, we observe that NCSN++ attains only approximately 45\% memorization ratio, whereas DDPM++ achieves a memorization ratio exceeds 90\%, all under the identical training dataset $\mathcal{D}$.\looseness=-1

% In our preliminary experiments, we explore the influence of skip connection quantity on memorization. Specifically, we consider $m=1, 3, 5, 7, 8, 9$ skip connections. To make it computationally tractable, we randomly select 5 different combinations of skip connections for each $m$. Consequently, we observe that for large $m$, the memorization ratio maintains similar levels while demonstrates large variance for small $m$. This highlights the presence of skip connection sparsity as a factor impacting memorization in diffusion models, which motivates us to explore the effects of specific skip connections. We consider both DDPM++ architecture and NCSN++ architecture in \citet{song2021score,karras2022elucidating} to conduct following experiments.

\begin{figure}[t]
\centering
\subfigure[]{\includegraphics[width=0.24\textwidth]{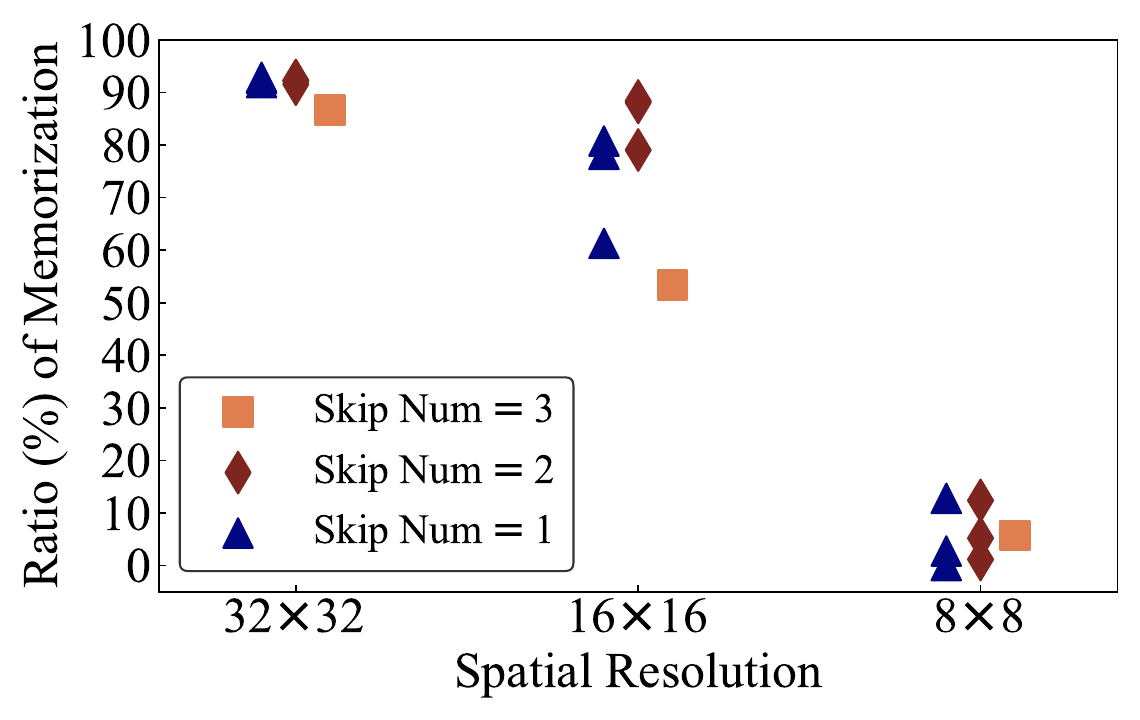}\label{fig_model_skip1_vp}}
\subfigure[]{\includegraphics[width=0.24\textwidth]{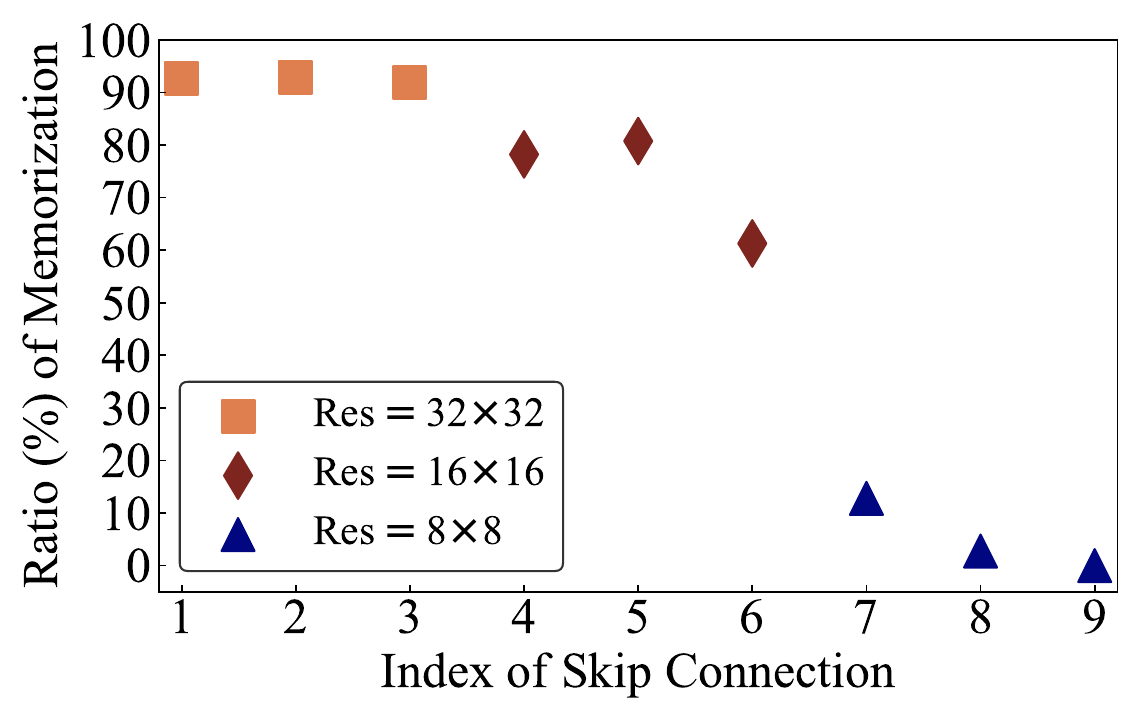}\label{fig_model_skip2_vp}}
\subfigure[]{\includegraphics[width=0.24\textwidth]{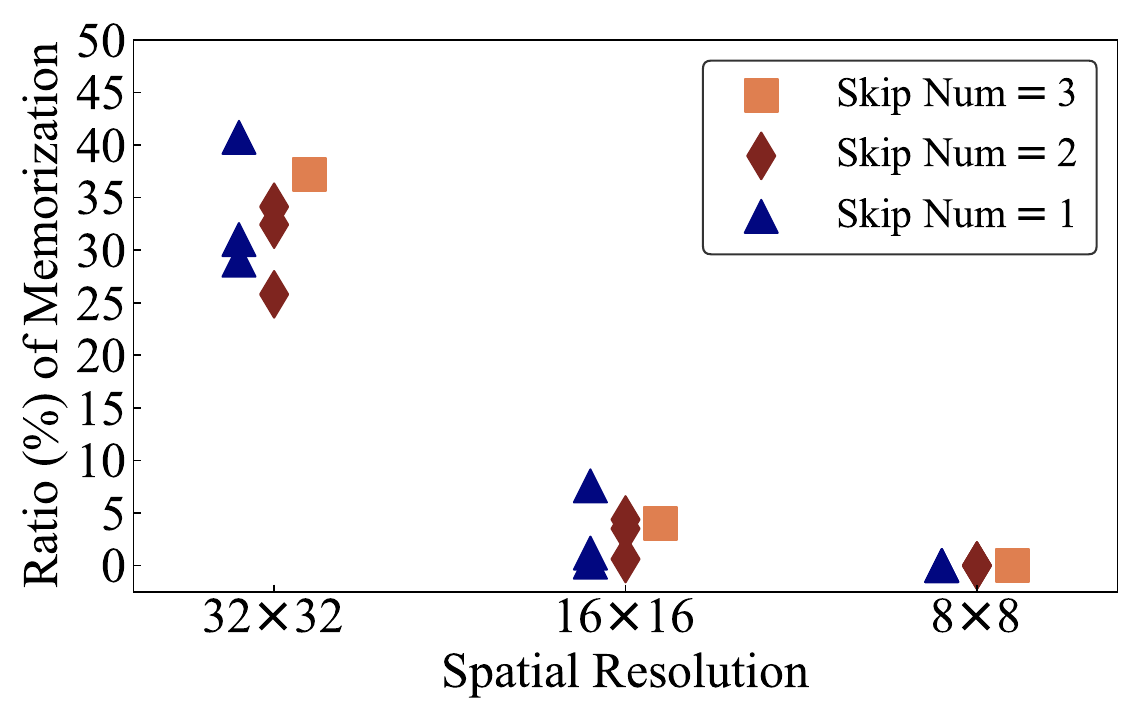}\label{fig_model_skip1_ve}}
\subfigure[]{\includegraphics[width=0.24\textwidth]{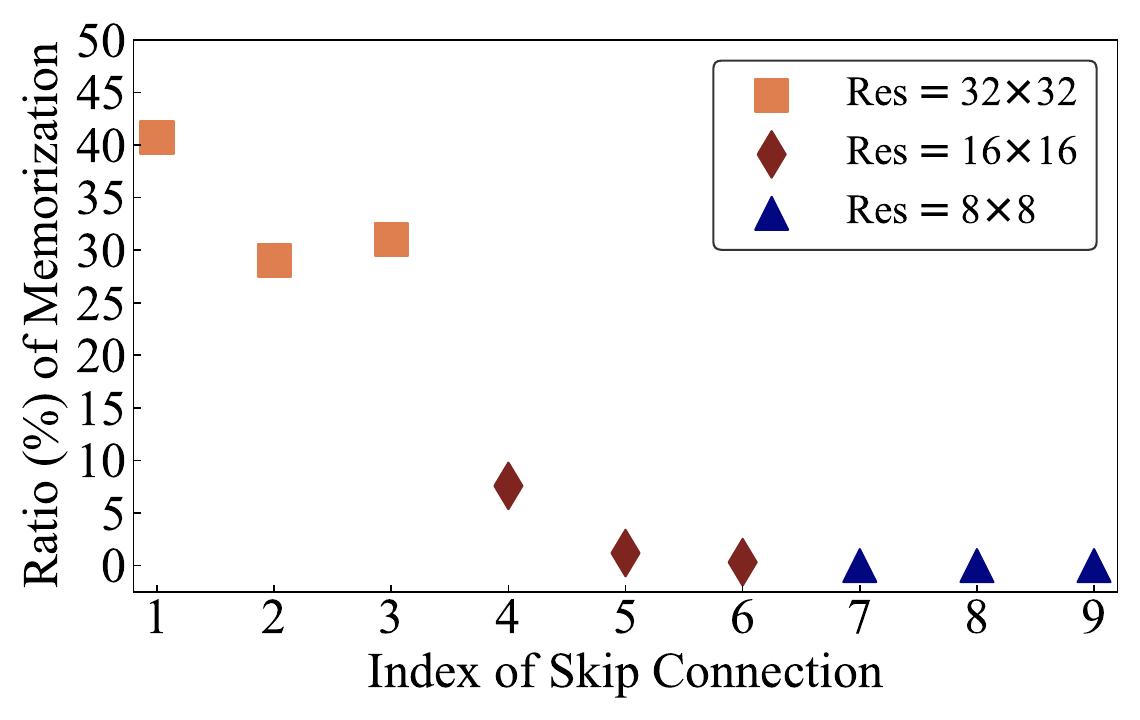}\label{fig_model_skip2_ve}}
\vspace{-0.3cm}
\caption{\textbf{Memorization ratio (\%)} of diffusion models on CIFAR-10 when retaining (a) skip connections of certain spatial resolution for DDPM++; (b) single skip connection at different locations for DDPM++; (c) skip connections of certain spatial resolution for NCSN++; (d) single skip connection at different locations for NCSN++.\looseness=-1}
\vspace{-0.3cm}
\end{figure}

\textbf{Skip connection resolution.} We narrow our examination to the inclusion of a selected number of skip connections, specifically $m=\,$1,2,3, all situated at a particular spatial resolution. The results, illustrated in Figure~\ref{fig_model_skip1_vp} and Figure~\ref{fig_model_skip1_ve}, unveil notable trends. We note that different markers represent distinct quantities of skip connections. Our observations reveal that skip connections situated at higher resolutions contribute more significantly to memorization. Interestingly, we also find that an increase in the number of skip connections does not consistently result in higher memorization ratio. For instance, the DDPM++ model with $m=\,$3 at a spatial resolution of 16$\times$16 exhibits a lower memorization ratio compared to cases where $m=\,$1 or $m=\,$2.\looseness=-1

\textbf{Skip connection location.} Additionally, we retain a single skip connection but alter its placement within the model architecture. As depicted in Figure~\ref{fig_model_skip2_vp} and Figure~\ref{fig_model_skip2_ve}, with the presence of just one skip connection, DDPM++ can achieve a memorization ratio exceeding 90\%, while NCSN++ attains a memorization ratio surpassing 40\%, provided that this single skip connection is positioned at a resolution of 32$\times$32. This further reinforces that skip connections at higher resolutions play a more substantial role in memorization of diffusion models.\looseness=-1

\vspace{-.2cm}
\section{Training Procedure \texorpdfstring{$\mathcal{T}$}{TEXT}}
\vspace{-.2cm}
% \subsection{Training Iterations}

% \subsection{Training Weight Schedule}
% Exploring factors: how to schedule the training weights $\lambda(t)$.

In Section~\ref{sec2}, we have demonstrated that the ratio of memorization increases with the progression of training epochs, indicating the influence of training procedure $\mathcal{T}$ on memorization. Consequently, in this section, we delve into the impact of various training hyperparameters on CIFAR-10.

\textbf{Batch size.} In light of prior research highlighting the substantial role of batch size in the performance of discriminative models~\citep{goyal2017accurate,keskar2016large}, we hypothesize that it may also influence memorization in diffusion models. Therefore, we investigate a range of batch sizes, from 128 to 896. Given that we maintain a constant number of training epochs, we apply the linear scaling rule proposed by \citet{goyal2017accurate} to adjust the learning rate accordingly for each batch size. Based on our basic experimental setup, we ensure a consistent ratio of learning rate to batch size $2\times10^{-4}/512$. As indicated in Table~\ref{tbl_decay_ema}, a larger batch size correlates with a higher memorization ratio.\looseness=-1

% However, it is important to note that the effect of batch size on memorization is not extremely strong in our setting. 

\textbf{Weight decay.} Weight decay is typically adopted to prevent neural networks from overfitting. Memorization can be regarded as an overfitting scenario in terms of diffusion models. Motivated by this, we explore its effect on memorization. In our basic experimental setup (see Appendix~\ref{appendix_setup}), we follow \citet{ho2020denoising,song2021score,karras2022elucidating} to set a zero weight decay. Now we set different values of weight decay and show the memorization results in Table~\ref{tbl_decay_ema}(\emph{left}). Consequently, we find that when weight decay is ranged between $0\sim 1\times10^{-3}$, it has a subtle contribution to memorization. When further increasing the value, the memorization ratio drastically decreases.  

\textbf{Exponential model average.} EMA was shown to effectively stabilize FIDs~\citep{heusel2017gans} and remove artifacts in generated samples~\citep{song2020improved}, as also shown in Table~\ref{fid decay_ema}(\emph{right}). It is widely adopted in current diffusion models~\citep{ho2020denoising,dhariwal2021diffusion,karras2022elucidating}. Motivated by this, we explore its effect on memorization. Previously, we fix the EMA rate as 0.99929 after the warmup following \citet{karras2022elucidating}. As present in Table \ref{tbl_decay_ema}(\emph{right}), we investigate the memorization of diffusion models with varying EMA rates, from 0.0 to 0.99929. Although EMA values are significant for FIDs, they contribute limitedly to memorization.\looseness=-1

% Different from FIDs, we observe that EMA values contribute limitedly to the memorization.\looseness=-1

% \begin{table}[t]
% \begin{minipage}[t]{0.46\linewidth}
% \begin{center}
% \begin{tabular}{c|cc}
% \toprule
% Learning rate     & $|\mathcal{D}|=1$k & $|\mathcal{D}|=2$k \\
% \midrule
% 1e-4 & 88.46& 54.82\\
% 2e-4 & 92.09& 60.93\\
% 5e-4 & \textbf{94.05}& 66.32\\
% 1e-3 & 93.65& \textbf{66.47}\\
% 2e-3 & 93.98& 65.72\\
% \bottomrule
% \end{tabular}
% \end{center}
% \caption{Effects of learning rate on ratio (\%) of memorization.}
% \label{mem_lr}
% \end{minipage}
% \hfill
% \begin{minipage}[t]{0.46\linewidth}
% \begin{center}
% \begin{tabular}{c|cc}
% \toprule
% Batch size     & $|\mathcal{D}|=1$k & $|\mathcal{D}|=2$k \\
% \midrule
% 128 & \textbf{92.85} & \textbf{64.91}\\
% 256 & 92.58& 63.55\\
% 512 & 92.09& 60.93\\
% 768 & 90.13& 59.18\\
% 1000 & 89.30& 57.53\\
% \bottomrule
% \end{tabular}
% \end{center}
% \caption{Effects of batch size on ratio (\%) of memorization.}
% \label{mem_bs}
% \end{minipage}
% \end{table}

\begin{table*}[t]
\begin{center}
\caption{\textbf{Memorization ratios (\%)} of diffusion models on CIFAR-10 under different batch sizes.}
\vspace{-.2cm}
\begin{tabular}{l|c|ccccccc}
\toprule
\multicolumn{2}{l|}{Batch Size} & 128 & 256 & 384 & 512 & 640 & 768 & 896 \\
\multicolumn{2}{l|}{Learning rate ($10^{-4}$)} & 0.5  & 1.0  & 1.5 & 2.0 & 2.5 & 3.0 & 3.5 \\
\midrule
\multirow{2}{*}{$\mathcal{R}_{\text{mem}}$(\%)}  &  $|\mathcal{D}|=$1k & 89.52 & 90.87 & 91.52& 92.09& 91.40& 92.05 & \textbf{92.77}\\
 & $|\mathcal{D}|=$2k & 56.67 & 60.36& 60.31& 60.93& 62.61& \textbf{63.33} & 61.95\\
\bottomrule
\end{tabular}
\end{center}
% \vspace{-.3cm}
\label{tbl_mem_batch}
\vspace{-.35cm}
\end{table*}

% \begin{table}[t]
% \begin{center}
% \begin{tabular}{l|ccccccccc}
% \toprule
% Weight decay & $0$  & $10^{-5}$  & $10^{-4}$ & $10^{-3}$  & $10^{-2}$ & $2\times10^{-2}$ & $5\times10^{-2}$ & $8\times10^{-2}$ & $10^{-1}$ \\
% \midrule
% $\mathcal{R}_{\text{mem}}$ for $|\mathcal{D}|=$1k & 92.09 & 91.67 & 92.47& 92.11& 89.07 & 75.05 & 13.79 & 4.22 & 1.33 \\
% $\mathcal{R}_{\text{mem}}$ for $|\mathcal{D}|=$2k & 56.67 & 60.36& 60.31& 60.93& 62.61& \textbf{63.33} & 61.95\\
% \bottomrule
% \end{tabular}
% \end{center}
% \caption{Effects of weight decay on the ratio (\%) of memorization.}
% \label{tbl_mem_decay}
% \end{table}

\begin{table}[t]
\caption{\textbf{Memorization ratios (\%)} of diffusion models on CIFAR-10 under different (\emph{left}) weight decay values; (\emph{right}) EMA values.}
\begin{minipage}[t]{0.48\linewidth}
\begin{center}
\begin{tabular}{c|cc}
\toprule
\multirow{2}{*}{Weight decay} & \multicolumn{2}{c}{$\mathcal{R}_{\text{mem}}$(\%)} \\
& $|\mathcal{D}|=$1k & $|\mathcal{D}|=$2k\\
\midrule
$0$  & 92.09 & 60.93\\
$1\times10^{-5}$ & 91.67 & \textbf{61.63} \\
$1\times10^{-4}$ & \textbf{92.47}& 61.03\\
$1\times10^{-3}$ & 92.11& 58.39\\
$1\times10^{-2}$ & 89.07& 35.88\\
% $2\times10^{-2}$ & 75.05& \,\,5.72\\
$5\times10^{-2}$ & 13.79& \,\,0.03\\
% $8\times10^{-2}$ & \,\,4.22& \,\,0.00\\
$1\times10^{-1}$ & \,\,1.33& \,\,0.00\\
\bottomrule
\end{tabular}
\end{center}
% \vspace{-.3cm}
% \caption{Memorization ratios (\%) of different values of weight decay.}
% \vspace{-.3cm}
\end{minipage}
\hfill
\begin{minipage}[t]{0.48\linewidth}
\begin{center}
\begin{tabular}{l|cc}
\toprule
\multirow{2}{*}{EMA} & \multicolumn{2}{c}{$\mathcal{R}_{\text{mem}}$(\%)} \\
& $|\mathcal{D}|=$1k & $|\mathcal{D}|=$2k\\
\midrule
0.99929  & \textbf{92.09} & 60.93\\
0.999 & 91.72 & \textbf{61.38}\\
0.99 & 91.45 & 59.27\\
0.9 & 90.16 & 58.31\\
% 0.8 & 90.55 & 58.03\\
0.5 & 90.42 & 57.80\\
% 0.2 & 90.81 & 60.50\\
0.1 & 90.78 & 58.00\\
0.0 & 90.19 & 59.20\\
\bottomrule
\end{tabular}
\end{center}
\end{minipage}
% \vspace{-.2cm}
\label{tbl_decay_ema}
\vspace{-.3cm}
\end{table}

\begin{figure*}[t]
\centering
\subfigure[Unconditional vs conditional]{\includegraphics[width=0.32\textwidth]{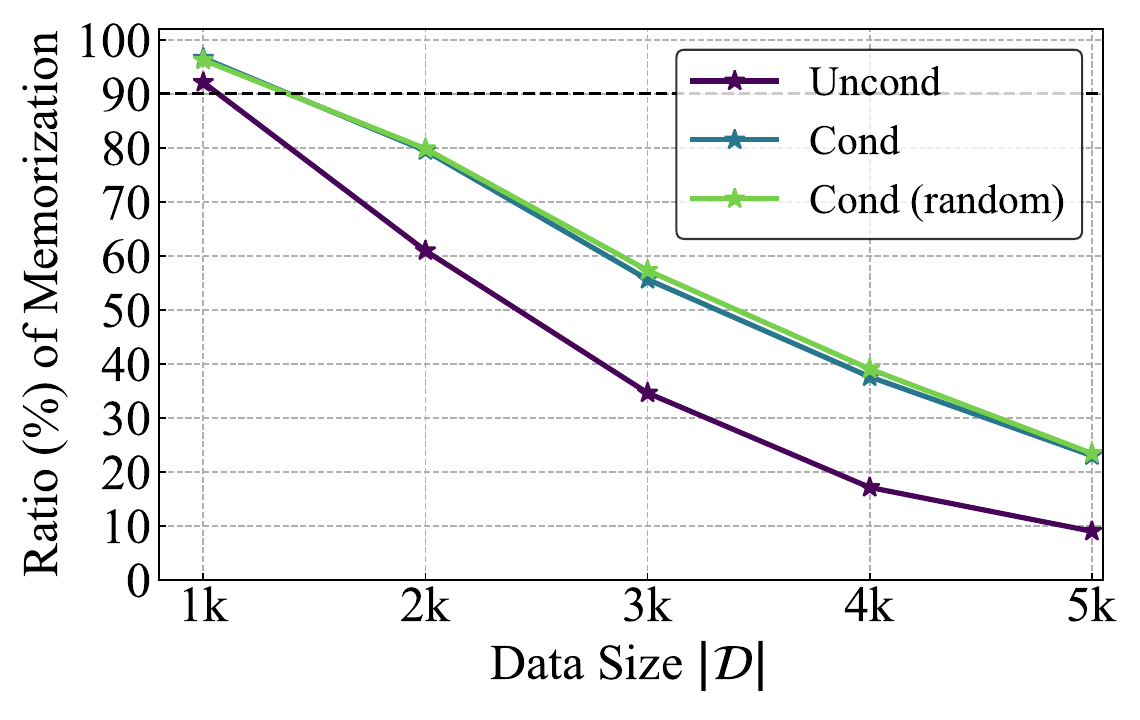}\label{fig_model_cond1}}
\subfigure[Varying $C$ for random labels]{\includegraphics[width=0.32\textwidth]{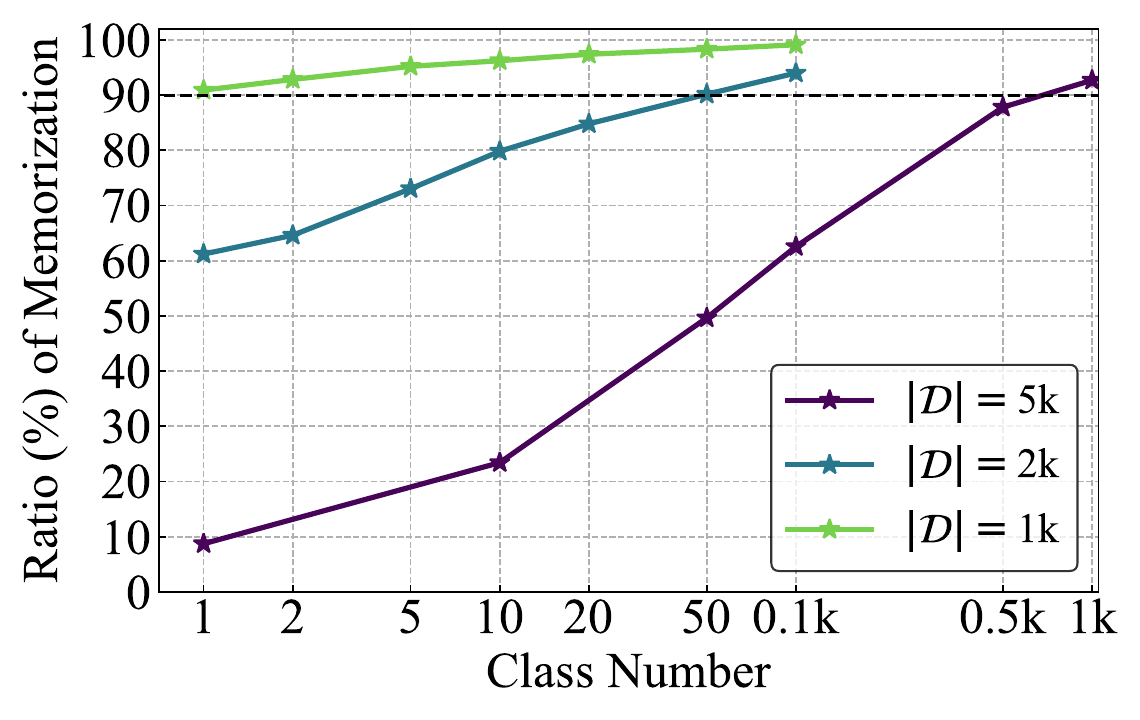}\label{fig_model_cond2}}
% \subfigure[]{\includegraphics[width=0.45\textwidth]{figures/memorize_cond_process.pdf}\label{fig_model_cond3}}
\subfigure[Unconditional vs Unique]{\includegraphics[width=0.32\textwidth]{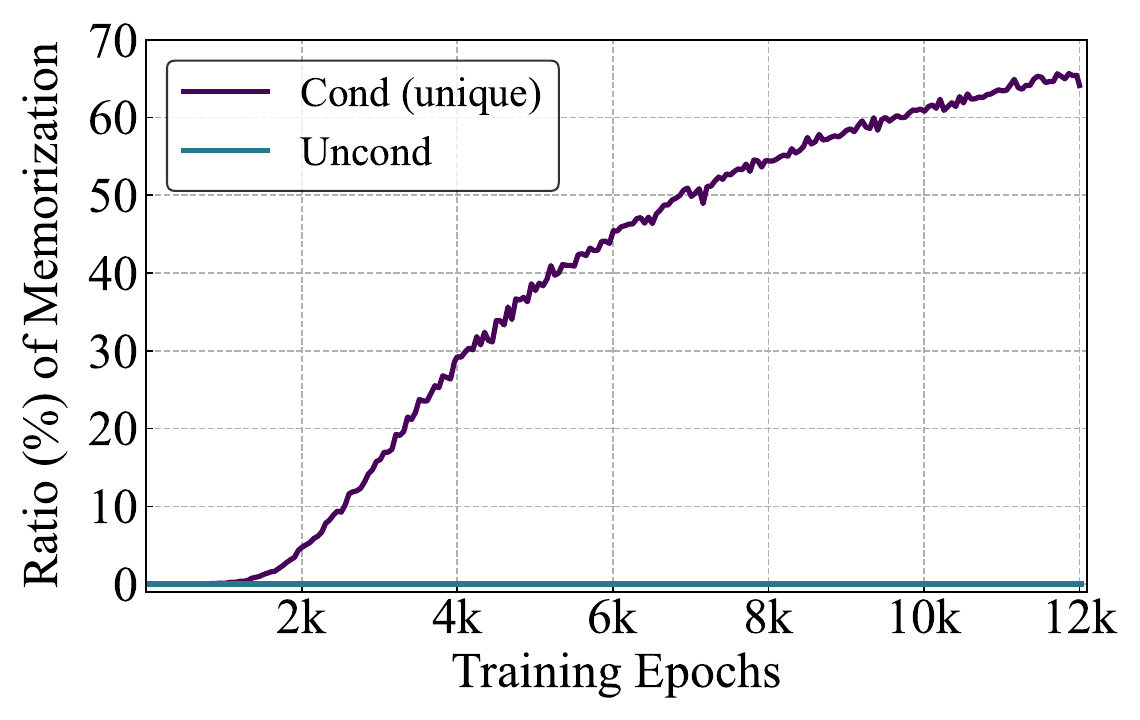}\label{fig_model_cond4}}
\vspace{-0.3cm}
\caption{\textbf{Memorization ratios (\%)} of (a) unconditional diffusion models and conditional diffusion models on CIFAR-10 with true / random labels; (b) conditional diffusion models with $C$ random labels; (c) conditional EDM with unique labels and unconditional EDM at $|\mathcal{D}|=\,$50k during the training.\looseness=-1}
\vspace{-0.3cm}
\end{figure*}

% We choose learning rate from the set $\{1\times10^{-4},2\times10^{-4},5\times10^{-4},1\times10^{-3},2\times10^{-3}\}$ and observe its effects on the memorization in the diffusion models. In our experiments, further increasing the learning rate results in a training failure with high loss values.  As present in Table~\ref{mem_lr}, within our selected range, the effect of the learning rate is not very significant for data size $|\mathcal{D}|=1$k. While for $|\mathcal{D}|=2$k, much smaller learning rate, e.g. $1\times10^{-4}$ results in apparent drop of memorization. Therefore, we conclude that a larger learning rate slightly increases the ratio of memorization. 

% \paragraph{Batch Size} Apart from the learning rate, we also modify the batch size during the training, whose range is $\{128, 256, 512, 768, 1000\}$. As shown in Table~\ref{mem_bs}, we notice that batch size also plays a role in the memorization of diffusion models. From our observation, when training with a smaller batch size, the learned diffusion model tends to have more memorization of training data. 

% \input{tables/lr_and_bs}

\vspace{-.2cm}
\section{Unconditional v.s.\ Conditional Generation}\label{sec cond}
\vspace{-.2cm}

It has shown that conditional diffusion models typically yield lower FIDs than their unconditional counterparts~\citep{karras2022elucidating}. This observation motivates us to investigate whether the input condition also exerts an influence on the memorization of diffusion models. It is worth noting that to train conditional diffusion models, we slightly change $\mathcal{M}$ by incorporating a class embedding layer and adjust $\mathcal{T}$ through modifications of the training objective. We show the training objective of class-conditioned diffusion models and its optimal solution in Appendix~\ref{appendix_optimal_cond}. The experiments are conducted using both EDM~\citep{karras2022elucidating} and Stable Diffusion~\citep{rombach2022high}.\looseness=-1

\textbf{Class conditioning.} As depicted in Figure~\ref{fig_model_cond1}, we note that class-conditional diffusion models exhibit higher memorization ratios compared to their unconditional counterparts (thus larger EMMs). We hypothesize that this observation is attributed to the additional information introduced by these true class labels. To validate the hypothesis put forth above, we substitute the true class labels with random labels. Intriguingly, we find that the memorization of diffusion models with random labels remains at a similar level to that of models with true labels. This contradicts our previous hypothesis, as random labels are uninformative regarding training images. These results align with earlier research in the realm of discriminative models, e.g., \citet{zhang2017understanding}, which suggests that deep neural networks can memorize training data even with randomly assigned labels. It is worth noting that when monitoring the FIDs in Figure~\ref{fid cond}, we observe that diffusion models conditioned on true labels yield lower FIDs than that on random labels. We hypothesize that with true labels as conditioning, similar training samples tend to share the same ``individual model'' for each class, resulting in better ``individual models''.\looseness=-1 

\textbf{Number of classes.} When employing random labels as conditions, the class number $C$ is not restricted to 10 in CIFAR-10. Therefore, we manipulate the choices of $C$ for random labels and then observe their effects on memorization. Initially, we compare Figure~\ref{fig_model_cond1} and Figure~\ref{fig_model_cond2}, revealing that unconditional diffusion models and conditional models with $C=\,$1 maintain similar memorization ratios. However, a discernible trend emerges as we introduce more classes, with diffusion models exhibiting increased memorization (also observed on FFHQ in Figure~\ref{fig_ffhq_cond}(c)). Intriguingly, with a size of training dataset $|\mathcal{D}|=\,$5k, even starting at a modest 10\% memorization ratio for $C=\,$1, conditional diffusion models can attain over 90\% memorization ratio with $C=\,$1k random labels. Based on these observations, we conclude that the number of classes significantly influences EMMs.\looseness=-1

\textbf{Unique class condition.} We consider an extreme scenario where each training sample in $\mathcal{D}$ is assigned a unique class label, which can be regarded as the case of $C=|\mathcal{D}|$. We compare the memorization dynamics of this unique class conditional scenario with true class conditional and unconditional ones during the training, as illustrated in Figure~\ref{fig_appendix_cond}. Notably, within only 8k epochs, the diffusion model with unique labels achieves a memorization ratio of more than 95\%. On the FFHQ dataset, the memorization ratios of unique class conditional diffusion models are also much higher than that of unconditional ones, as present in Figure~\ref{fig_ffhq_cond}(d). Inspired by this, we extend this condition mechanism to encompass the entire CIFAR-10 dataset, which contains $|\mathcal{D}|=\,$50k samples. It is worth noting that we train an EDM model \cite{karras2022elucidating} to facilitate a comparison with our initial observations in Section~\ref{sec2}. As illustrated in Figure~\ref{fig_model_cond4}, the unconditional EDM maintains a memorization ratio of zero throughout the training process even extending it to 12k training epochs. However, upon conditioning on the unique labels, we observe a notable shift on memorization. The trained conditional EDM achieves more than 65\% memorization ratio within 12k training epochs, a significant increase compared to its previous performance. We also visualize the generated images by these two models in Appendix~\ref{appendix_cond} to further validate their memorization behaviors. These results imply that with unique labels, the training samples become strongly associated with their input conditions, rendering them more readily accessible during the generation process when the same conditions are applied.\looseness=-1

\textbf{Extension to text condition.} Beyond unconditional diffusion models and class-conditioned diffusion models, we incorporate the experiments on state-of-the-art text-to-image diffusion models, Stable Diffusion (SD)~\citep{rombach2022high}. As it is infeasibly computationally expensive to train Stable Diffusion from scratch, we opt to fine-tune the U-Net in SD on the Imagenette dataset\footnote{\url{https://github.com/fastai/imagenette}}, which is a realistic scenario. This dataset consists of $C=10$ classes from the ImageNet dataset~\citep{deng2009imagenet} and has also been adopted in \citet{somepalli2023understanding}. We consider two types of text prompts as conditions: (1) \emph{plain conditioning}: using ``a picture'' as text prompt for all images; (2) \emph{class conditioning}: incorporating class labels in the text prompt, i.e., ``a picture of <CLASS\_NAME>''. For fine-tuning, we consider both full fine-tuning and LoRA fine-tuning with different ranks~\citep{hu2021lora}. More experimental setups are detailed in Appendix~\ref{appendix sd}.  As shown in Table~\ref{tbl_mem_stable}, we compare two distinct types of text prompts conditioning under training data with different sizes $|\mathcal{D}|$. Firstly, we observe that although SD was pretrained on billions of images, it is still prone to memorize training data when fully fine-tuning it on customized data. While LoRA fine-tuning has less tendency to memorize training data. Additionally, we find that with higher LoRA rank, SD models demonstrate higher memorization ratios.  Secondly, it is noticed that with the increase of $|\mathcal{D}|$, the memorization ratio drops. Furthermore, the memorization ratios of SDs fine-tuned on class conditioning data is much higher than that on plain conditioning data. To explain this, class conditioning is similar to the case of a number of classes $C=\,$10 in the class-conditioned diffusion model while plain conditioning is similar to the case $C=\,$1. Our results on Stable Diffusion reaffirm the significant influence of class number on memorization within diffusion models.\looseness=-1

\begin{table}[t]
\centering
\caption{\textbf{Memorization ratio (\%)} of \textbf{Stable Diffusion (SD)} on Imagenette with different fine-tuning approaches on the Imagenette dataset under two different text conditioning types.}
\vspace{-0.3cm}
\begin{tabular}{l|c|ccccc}
\toprule
Fine-tuning & Condition  & $|\mathcal{D}|=$400 & $|\mathcal{D}|=$500 & $|\mathcal{D}|=$800 & $|\mathcal{D}|=$1000 & $|\mathcal{D}|=$2000  \\
\midrule
\multirow{2}{*}{LoRA $r=$16} & plain & 0.00 & 0.00 & 0.00 & 0.00 & 0.00\\
& class & 0.00 & 0.00 & 0.00 & 0.00& 0.00\\
\midrule
\multirow{2}{*}{LoRA $r=$256} & plain & 0.08 & 0.01 & 0.00 & 0.00 & 0.00\\
& class & 5.22 & 0.25 & 0.00 & 0.00& 0.00\\
\midrule
\multirow{2}{*}{LoRA $r=$1024} & plain & 0.76 & 0.20 & 0.00 & 0.00 & 0.00\\
& class & 8.77 & 2.07 & 0.04 & 0.00& 0.00\\
\midrule
\multirow{2}{*}{Full fine-tuning} & plain & 21.95 &  20.63 & \,\,7.30 & \,\,2.85 & \,\,0.06\\
& class & 46.89 & 38.79 & 20.46 & 10.67 & \,\,0.21\\
\bottomrule
\end{tabular}
\label{tbl_mem_stable}
\vspace{-0.2cm}
\end{table}

\vspace{-.2cm}
\section{Related Work}
\vspace{-.2cm}

\textbf{Memorization in discriminative models.} For discriminative models, memorization and generalization have interleaving connections. \citet{zhang2017understanding} first demonstrated that deep learning models can memorize the training data, even with random labeling, but generalize well. Additionally, \citet{feldman2020does, feldman2020neural} showed that this memorization is necessary for achieving close-to-optimal generalization under long-tailed assumption of data distribution. In the follow-up works, \citet{baldock2021deep,stephenson2021geometry} showed that memorization predominately happens in the deep layers while \citet{maini2023can} argued that memorization is confined to a few neurons across various layers of deep neural networks. Although discriminative models can largely memorize training data, this phenomenon does not apply to diffusion models. For instance, \citet{zhang2017understanding} showed that the Inception v3 model~\citep{szegedy2016rethinking} with under 25M trainable parameters can almost memorize the ImageNet dataset~\citep{deng2009imagenet} with approximately 1.28M images. However, the EDM model (about 56M parameters)~\citep{karras2022elucidating} can not memorize even the CIFAR-10 dataset with 50k images as present in Section~\ref{sec2}. From another view, \citet{bartlett2020benign, nakkiran2020deep} showed that over-parameterized models generalize well to real data distribution and even perfectly fit to the training dataset, which is called \textit{benign overfitting}. Nevertheless, diffusion models demonstrate adverse generalization performance~\citep{yoon2023diffusion}.\looseness=-1

\textbf{Memorization in generative models.} \citet{somepalli2023diffusion,carlini2023extracting} represent the initial explorations on memorization in diffusion models. They investigated a range of diffusion models and demonstrated that these models memorize a few training samples and replicate them globally or in the object level during the generation. For instance, \citet{carlini2023extracting} identified only 50 memorized training images from 175 million generated images by Stable Diffusion~\citep{rombach2022high} and extracted 200-300 training images from $2^{20}$ generated images by DDPM and its variant~\citep{ho2020denoising, nichol2021improved}. This highlights the memorization gap between empirical diffusion models and the theoretical optimum defined in~\eqref{optimal}. \citet{somepalli2023diffusion, somepalli2023understanding} showed that training data duplication and text conditioning play significant roles in the memorization of diffusion models from empirical studies. Another work~\citep{wen2023detecting} introduced a novel approach to detect memorized prompts and mitigate memorization for Stable Diffusion~\citep{rombach2022high}. However, these conclusions are mostly derived from text-to-image diffusion models. The nature of memorization in diffusion models, especially for unconditional ones, remains unexplored. One recent paper \citet{hintersdorf2024finding} localized the memorization to the specific neurons in text-to-image diffusion models. Another recent paper \citet{liu2024iterative} proposed ensembling model parameters and discarding samples with small loss values in diffusion model training, thus mitigating the memorization. Apart from diffusion models, several studies are researching towards memorization in Generative Adversarial Networks (GANs)~\citep{webster2021person,feng2021gans}, and language models~\citep{carlini2021extracting,carlini2022quantifying}.\looseness=-1

\textbf{Generalization in diffusion models.} \citet{yoon2023diffusion, kadkhodaie2023generalization} analyzed the relationship between memorization and generalization. \citet{yoon2023diffusion} hypothesized that generalization of diffusion models is a failure to memorize the training data when data size is large and model width is comparatively small and this memorization-generalization dichotomy may manifest at the level of classes. Our work focuses on the timing of occurrence and the influential factors regarding memorization through a much more thorough quantitative analysis. \citet{kadkhodaie2023generalization} also showed that diffusion models memorize samples when trained on small data. However, it focused on examining the inductive bias of DNN denoisers and showed that generalization arises from the geometry-adaptive harmonic basis (GAHB) using mathematical formulations. Additionally, \citet{zhang2023emergence} showed that given the same noise input, different diffusion models often generate remarkably similar samples through a deterministic sampler, manifesting in memorization and generalization regimes. This provides a new perspective on understanding memorization and generalization of diffusion models. One recent paper~\citep{kamb2024analytic} shares motivations similar to ours. \citet{kamb2024analytic} identified two inductive biases (locality and equivariance) in common architectures of diffusion models that prevent convergence to the ideal score function. Building on this insight, they theoretically derive equivariant local score (ELS) machines, which mirror the formulation of the ideal score function while exhibiting the generalization behavior of empirically trained models. However, these two inductive biases do not fully capture the behavior of trained models, particularly those with self-attention architectures. Analytical characterization of the exact properties of trained diffusion models remains challenging. Our work, on the empirical side, investigates how data, model designs, training processes, and conditioning prevent trained models from converging to the ideal score function.\looseness=-1

\looseness=-1

% Apart from diffusion models, there are several studies researching towards memorization in Generative Adversarial Networks (GANs)~\citep{webster2021person,feng2021gans}, and language models~\citep{carlini2021extracting,carlini2022quantifying}.\looseness=-1

\vspace{-0.2cm}
\section{Conclusion}
\vspace{-.2cm}

In this study, we first showed that the theoretical optimum of diffusion models can only replicate training data, representing a memorization behavior. This contracts the typical generalization ability demonstrated by state-of-the-art diffusion models. To understand the memorization gap, we found that when trained on smaller-sized datasets, learned diffusion models tend to approximate the theoretical optimum. Motivated by this, we defined the notion of effective model memorization (EMM) to quantify this memorization behavior. Afterwards, we explored the impact of critical factors on memorization through the lens of EMM, from the three facets of data distribution, model configuration, and training procedure. We found that data dimension, model size, time embedding, skip connections, and class conditions play significant roles on memorization. Among all illuminating results, the memorization of diffusion models can be triggered by conditioning training data on completely random and uninformative labels. Intriguingly, when incorporating such conditioning design, more than 65\% of samples generated by diffusion models trained on the entire 50k CIFAR-10 images are replicas of training data, in contrast to the original 0\%. We believe that our study deepens the understanding of memorization in diffusion models and offers clues to theoretical research in the area of generative modeling. \looseness=-1

\subsubsection*{Acknowledgments}
We would like to thank our action editor Kamalika Chaudhuri and the reviewers for their detailed and constructive comments and feedback.

\bibliography{main}

\begin{thebibliography}{58}
\providecommand{\natexlab}[1]{#1}
\providecommand{\url}[1]{\texttt{#1}}
\expandafter\ifx\csname urlstyle\endcsname\relax
  \providecommand{\doi}[1]{doi: #1}\else
  \providecommand{\doi}{doi: \begingroup \urlstyle{rm}\Url}\fi

\bibitem[Baldock et~al.(2021)Baldock, Maennel, and Neyshabur]{baldock2021deep}
Robert Baldock, Hartmut Maennel, and Behnam Neyshabur.
\newblock Deep learning through the lens of example difficulty.
\newblock \emph{Advances in Neural Information Processing Systems (NeurIPS)}, 34:\penalty0 10876--10889, 2021.

\bibitem[Bartlett et~al.(2020)Bartlett, Long, Lugosi, and Tsigler]{bartlett2020benign}
Peter~L Bartlett, Philip~M Long, G{\'a}bor Lugosi, and Alexander Tsigler.
\newblock Benign overfitting in linear regression.
\newblock \emph{Proceedings of the National Academy of Sciences}, 117\penalty0 (48):\penalty0 30063--30070, 2020.

\bibitem[Carlini et~al.(2021)Carlini, Tramer, Wallace, Jagielski, Herbert-Voss, Lee, Roberts, Brown, Song, Erlingsson, et~al.]{carlini2021extracting}
Nicholas Carlini, Florian Tramer, Eric Wallace, Matthew Jagielski, Ariel Herbert-Voss, Katherine Lee, Adam Roberts, Tom Brown, Dawn Song, Ulfar Erlingsson, et~al.
\newblock Extracting training data from large language models.
\newblock In \emph{30th USENIX Security Symposium (USENIX Security 21)}, pp.\  2633--2650, 2021.

\bibitem[Carlini et~al.(2022)Carlini, Ippolito, Jagielski, Lee, Tramer, and Zhang]{carlini2022quantifying}
Nicholas Carlini, Daphne Ippolito, Matthew Jagielski, Katherine Lee, Florian Tramer, and Chiyuan Zhang.
\newblock Quantifying memorization across neural language models.
\newblock \emph{arXiv preprint arXiv:2202.07646}, 2022.

\bibitem[Carlini et~al.(2023)Carlini, Hayes, Nasr, Jagielski, Sehwag, Tramer, Balle, Ippolito, and Wallace]{carlini2023extracting}
Nicholas Carlini, Jamie Hayes, Milad Nasr, Matthew Jagielski, Vikash Sehwag, Florian Tramer, Borja Balle, Daphne Ippolito, and Eric Wallace.
\newblock Extracting training data from diffusion models.
\newblock \emph{arXiv preprint arXiv:2301.13188}, 2023.

\bibitem[Deng et~al.(2009)Deng, Dong, Socher, Li, Li, and Fei-Fei]{deng2009imagenet}
Jia Deng, Wei Dong, Richard Socher, Li-Jia Li, Kai Li, and Li~Fei-Fei.
\newblock Imagenet: A large-scale hierarchical image database.
\newblock In \emph{IEEE Conference on Computer Vision and Pattern Recognition (CVPR)}, 2009.

\bibitem[Dhariwal \& Nichol(2021)Dhariwal and Nichol]{dhariwal2021diffusion}
Prafulla Dhariwal and Alexander Nichol.
\newblock Diffusion models beat gans on image synthesis.
\newblock In \emph{Advances in Neural Information Processing Systems (NeurIPS)}, 2021.

\bibitem[Feldman(2020)]{feldman2020does}
Vitaly Feldman.
\newblock Does learning require memorization? a short tale about a long tail.
\newblock In \emph{Proceedings of the 52nd Annual ACM SIGACT Symposium on Theory of Computing}, pp.\  954--959, 2020.

\bibitem[Feldman \& Zhang(2020)Feldman and Zhang]{feldman2020neural}
Vitaly Feldman and Chiyuan Zhang.
\newblock What neural networks memorize and why: Discovering the long tail via influence estimation.
\newblock \emph{Advances in Neural Information Processing Systems (NeurIPS)}, 33:\penalty0 2881--2891, 2020.

\bibitem[Feng et~al.(2021)Feng, Guo, Benitez-Quiroz, and Martinez]{feng2021gans}
Qianli Feng, Chenqi Guo, Fabian Benitez-Quiroz, and Aleix~M Martinez.
\newblock When do gans replicate? on the choice of dataset size.
\newblock In \emph{Proceedings of the IEEE/CVF International Conference on Computer Vision (ICCV)}, pp.\  6701--6710, 2021.

\bibitem[Goyal et~al.(2017)Goyal, Doll{\'a}r, Girshick, Noordhuis, Wesolowski, Kyrola, Tulloch, Jia, and He]{goyal2017accurate}
Priya Goyal, Piotr Doll{\'a}r, Ross Girshick, Pieter Noordhuis, Lukasz Wesolowski, Aapo Kyrola, Andrew Tulloch, Yangqing Jia, and Kaiming He.
\newblock Accurate, large minibatch sgd: Training imagenet in 1 hour.
\newblock \emph{arXiv preprint arXiv:1706.02677}, 2017.

\bibitem[Heusel et~al.(2017)Heusel, Ramsauer, Unterthiner, Nessler, and Hochreiter]{heusel2017gans}
Martin Heusel, Hubert Ramsauer, Thomas Unterthiner, Bernhard Nessler, and Sepp Hochreiter.
\newblock Gans trained by a two time-scale update rule converge to a local nash equilibrium.
\newblock In \emph{Advances in Neural Information Processing Systems (NeurIPS)}, pp.\  6626--6637, 2017.

\bibitem[Hintersdorf et~al.(2024)Hintersdorf, Struppek, Kersting, Dziedzic, and Boenisch]{hintersdorf2024finding}
Dominik Hintersdorf, Lukas Struppek, Kristian Kersting, Adam Dziedzic, and Franziska Boenisch.
\newblock Finding nemo: Localizing neurons responsible for memorization in diffusion models.
\newblock \emph{arXiv preprint arXiv:2406.02366}, 2024.

\bibitem[Ho et~al.(2020)Ho, Jain, and Abbeel]{ho2020denoising}
Jonathan Ho, Ajay Jain, and Pieter Abbeel.
\newblock Denoising diffusion probabilistic models.
\newblock In \emph{Advances in Neural Information Processing Systems (NeurIPS)}, 2020.

\bibitem[Hu et~al.(2021)Hu, Shen, Wallis, Allen-Zhu, Li, Wang, Wang, and Chen]{hu2021lora}
Edward~J Hu, Yelong Shen, Phillip Wallis, Zeyuan Allen-Zhu, Yuanzhi Li, Shean Wang, Lu~Wang, and Weizhu Chen.
\newblock Lora: Low-rank adaptation of large language models.
\newblock \emph{arXiv preprint arXiv:2106.09685}, 2021.

\bibitem[Huang et~al.(2023)Huang, Huang, Yang, Ren, Liu, Li, Ye, Liu, Yin, and Zhao]{huang2023make}
Rongjie Huang, Jiawei Huang, Dongchao Yang, Yi~Ren, Luping Liu, Mingze Li, Zhenhui Ye, Jinglin Liu, Xiang Yin, and Zhou Zhao.
\newblock Make-an-audio: Text-to-audio generation with prompt-enhanced diffusion models.
\newblock \emph{arXiv preprint arXiv:2301.12661}, 2023.

\bibitem[Kadkhodaie et~al.(2023)Kadkhodaie, Guth, Simoncelli, and Mallat]{kadkhodaie2023generalization}
Zahra Kadkhodaie, Florentin Guth, Eero~P Simoncelli, and St{\'e}phane Mallat.
\newblock Generalization in diffusion models arises from geometry-adaptive harmonic representation.
\newblock \emph{arXiv preprint arXiv:2310.02557}, 2023.

\bibitem[Kamb \& Ganguli(2024)Kamb and Ganguli]{kamb2024analytic}
Mason Kamb and Surya Ganguli.
\newblock An analytic theory of creativity in convolutional diffusion models.
\newblock \emph{arXiv preprint arXiv:2412.20292}, 2024.

\bibitem[Karras et~al.(2019)Karras, Laine, and Aila]{karras2019style}
Tero Karras, Samuli Laine, and Timo Aila.
\newblock A style-based generator architecture for generative adversarial networks.
\newblock In \emph{IEEE International Conference on Computer Vision (CVPR)}, 2019.

\bibitem[Karras et~al.(2022)Karras, Aittala, Aila, and Laine]{karras2022elucidating}
Tero Karras, Miika Aittala, Timo Aila, and Samuli Laine.
\newblock Elucidating the design space of diffusion-based generative models.
\newblock \emph{Advances in Neural Information Processing Systems (NeurIPS)}, 35:\penalty0 26565--26577, 2022.

\bibitem[Keskar et~al.(2016)Keskar, Mudigere, Nocedal, Smelyanskiy, and Tang]{keskar2016large}
Nitish~Shirish Keskar, Dheevatsa Mudigere, Jorge Nocedal, Mikhail Smelyanskiy, and Ping Tak~Peter Tang.
\newblock On large-batch training for deep learning: Generalization gap and sharp minima.
\newblock \emph{arXiv preprint arXiv:1609.04836}, 2016.

\bibitem[Kim et~al.(2022)Kim, Kim, and Yoon]{kim2022guided}
Heeseung Kim, Sungwon Kim, and Sungroh Yoon.
\newblock Guided-tts: A diffusion model for text-to-speech via classifier guidance.
\newblock In \emph{International Conference on Machine Learning (ICML)}, pp.\  11119--11133. PMLR, 2022.

\bibitem[Kingma et~al.(2021)Kingma, Salimans, Poole, and Ho]{kingma2021variational}
Diederik Kingma, Tim Salimans, Ben Poole, and Jonathan Ho.
\newblock Variational diffusion models.
\newblock In \emph{Advances in Neural Information Processing Systems (NeurIPS)}, 2021.

\bibitem[Kingma \& Ba(2014)Kingma and Ba]{kingma2014adam}
Diederik~P Kingma and Jimmy Ba.
\newblock Adam: A method for stochastic optimization.
\newblock \emph{arXiv preprint arXiv:1412.6980}, 2014.

\bibitem[Kingma \& Gao(2023)Kingma and Gao]{kingma2023understanding}
Diederik~P Kingma and Ruiqi Gao.
\newblock Understanding the diffusion objective as a weighted integral of elbos.
\newblock \emph{arXiv preprint arXiv:2303.00848}, 2023.

\bibitem[Lin et~al.(2023)Lin, Gao, Tang, Takikawa, Zeng, Huang, Kreis, Fidler, Liu, and Lin]{lin2023magic3d}
Chen-Hsuan Lin, Jun Gao, Luming Tang, Towaki Takikawa, Xiaohui Zeng, Xun Huang, Karsten Kreis, Sanja Fidler, Ming-Yu Liu, and Tsung-Yi Lin.
\newblock Magic3d: High-resolution text-to-3d content creation.
\newblock In \emph{IEEE/CVF Conference on Computer Vision and Pattern Recognition (CVPR)}, pp.\  300--309, 2023.

\bibitem[Liu et~al.(2024)Liu, Guan, Wu, and Miao]{liu2024iterative}
Xiao Liu, Xiaoliu Guan, Yu~Wu, and Jiaxu Miao.
\newblock Iterative ensemble training with anti-gradient control for mitigating memorization in diffusion models.
\newblock \emph{arXiv preprint arXiv:2407.15328}, 2024.

\bibitem[Maini et~al.(2023)Maini, Mozer, Sedghi, Lipton, Kolter, and Zhang]{maini2023can}
Pratyush Maini, Michael~C Mozer, Hanie Sedghi, Zachary~C Lipton, J~Zico Kolter, and Chiyuan Zhang.
\newblock Can neural network memorization be localized?
\newblock \emph{arXiv preprint arXiv:2307.09542}, 2023.

\bibitem[Nakkiran et~al.(2020)Nakkiran, Kaplun, Bansal, Yang, Barak, and Sutskever]{nakkiran2020deep}
Preetum Nakkiran, Gal Kaplun, Yamini Bansal, Tristan Yang, Boaz Barak, and Ilya Sutskever.
\newblock Deep double descent: Where bigger models and more data hurt.
\newblock In \emph{International Conference on Learning Representations (ICLR)}, 2020.

\bibitem[Nichol et~al.(2021)Nichol, Dhariwal, Ramesh, Shyam, Mishkin, McGrew, Sutskever, and Chen]{nichol2021glide}
Alex Nichol, Prafulla Dhariwal, Aditya Ramesh, Pranav Shyam, Pamela Mishkin, Bob McGrew, Ilya Sutskever, and Mark Chen.
\newblock Glide: Towards photorealistic image generation and editing with text-guided diffusion models.
\newblock \emph{arXiv preprint arXiv:2112.10741}, 2021.

\bibitem[Nichol \& Dhariwal(2021)Nichol and Dhariwal]{nichol2021improved}
Alexander~Quinn Nichol and Prafulla Dhariwal.
\newblock Improved denoising diffusion probabilistic models.
\newblock In \emph{International Conference on Machine Learning (ICML)}, pp.\  8162--8171. PMLR, 2021.

\bibitem[Poole et~al.(2023)Poole, Jain, Barron, and Mildenhall]{poole2022dreamfusion}
Ben Poole, Ajay Jain, Jonathan~T Barron, and Ben Mildenhall.
\newblock Dreamfusion: Text-to-3d using 2d diffusion.
\newblock In \emph{International Conference on Learning Representations (ICLR)}, 2023.

\bibitem[Ramesh et~al.(2022)Ramesh, Dhariwal, Nichol, Chu, and Chen]{ramesh2022hierarchical}
Aditya Ramesh, Prafulla Dhariwal, Alex Nichol, Casey Chu, and Mark Chen.
\newblock Hierarchical text-conditional image generation with clip latents.
\newblock \emph{arXiv preprint arXiv:2204.06125}, 2022.

\bibitem[Rombach et~al.(2022)Rombach, Blattmann, Lorenz, Esser, and Ommer]{rombach2022high}
Robin Rombach, Andreas Blattmann, Dominik Lorenz, Patrick Esser, and Bj{\"o}rn Ommer.
\newblock High-resolution image synthesis with latent diffusion models.
\newblock In \emph{IEEE Conference on Computer Vision and Pattern Recognition (CVPR)}, 2022.

\bibitem[Ronneberger et~al.(2015)Ronneberger, Fischer, and Brox]{ronneberger2015u}
Olaf Ronneberger, Philipp Fischer, and Thomas Brox.
\newblock U-net: Convolutional networks for biomedical image segmentation.
\newblock In \emph{International Conference on Medical Image Computing and Computer-Assisted Intervention (MICCAI)}, pp.\  234--241. Springer, 2015.

\bibitem[Sohl-Dickstein et~al.(2015)Sohl-Dickstein, Weiss, Maheswaranathan, and Ganguli]{sohl2015deep}
Jascha Sohl-Dickstein, Eric Weiss, Niru Maheswaranathan, and Surya Ganguli.
\newblock Deep unsupervised learning using nonequilibrium thermodynamics.
\newblock In \emph{International Conference on Machine Learning (ICML)}, pp.\  2256--2265. PMLR, 2015.

\bibitem[Somepalli et~al.(2023{\natexlab{a}})Somepalli, Singla, Goldblum, Geiping, and Goldstein]{somepalli2023diffusion}
Gowthami Somepalli, Vasu Singla, Micah Goldblum, Jonas Geiping, and Tom Goldstein.
\newblock Diffusion art or digital forgery? investigating data replication in diffusion models.
\newblock In \emph{IEEE/CVF Conference on Computer Vision and Pattern Recognition (CVPR)}, pp.\  6048--6058, 2023{\natexlab{a}}.

\bibitem[Somepalli et~al.(2023{\natexlab{b}})Somepalli, Singla, Goldblum, Geiping, and Goldstein]{somepalli2023understanding}
Gowthami Somepalli, Vasu Singla, Micah Goldblum, Jonas Geiping, and Tom Goldstein.
\newblock Understanding and mitigating copying in diffusion models.
\newblock \emph{arXiv preprint arXiv:2305.20086}, 2023{\natexlab{b}}.

\bibitem[Song et~al.(2021{\natexlab{a}})Song, Meng, and Ermon]{song2021denoising}
Jiaming Song, Chenlin Meng, and Stefano Ermon.
\newblock Denoising diffusion implicit models.
\newblock In \emph{International Conference on Learning Representations (ICLR)}, 2021{\natexlab{a}}.

\bibitem[Song \& Ermon(2019)Song and Ermon]{song2019generative}
Yang Song and Stefano Ermon.
\newblock Generative modeling by estimating gradients of the data distribution.
\newblock In \emph{Advances in Neural Information Processing Systems (NeurIPS)}, pp.\  11895--11907, 2019.

\bibitem[Song \& Ermon(2020)Song and Ermon]{song2020improved}
Yang Song and Stefano Ermon.
\newblock Improved techniques for training score-based generative models.
\newblock In \emph{Advances in Neural Information Processing Systems (NeurIPS)}, 2020.

\bibitem[Song et~al.(2021{\natexlab{b}})Song, Sohl-Dickstein, Kingma, Kumar, Ermon, and Poole]{song2021score}
Yang Song, Jascha Sohl-Dickstein, Diederik~P Kingma, Abhishek Kumar, Stefano Ermon, and Ben Poole.
\newblock Score-based generative modeling through stochastic differential equations.
\newblock In \emph{International Conference on Learning Representations (ICLR)}, 2021{\natexlab{b}}.

\bibitem[Stephenson et~al.(2021)Stephenson, Padhy, Ganesh, Hui, Tang, and Chung]{stephenson2021geometry}
Cory Stephenson, Suchismita Padhy, Abhinav Ganesh, Yue Hui, Hanlin Tang, and SueYeon Chung.
\newblock On the geometry of generalization and memorization in deep neural networks.
\newblock \emph{arXiv preprint arXiv:2105.14602}, 2021.

\bibitem[Szegedy et~al.(2016)Szegedy, Vanhoucke, Ioffe, Shlens, and Wojna]{szegedy2016rethinking}
Christian Szegedy, Vincent Vanhoucke, Sergey Ioffe, Jon Shlens, and Zbigniew Wojna.
\newblock Rethinking the inception architecture for computer vision.
\newblock In \emph{IEEE Conference on Computer Vision and Pattern Recognition (CVPR)}, pp.\  2818--2826, 2016.

\bibitem[Tancik et~al.(2020)Tancik, Srinivasan, Mildenhall, Fridovich-Keil, Raghavan, Singhal, Ramamoorthi, Barron, and Ng]{tancik2020fourier}
Matthew Tancik, Pratul Srinivasan, Ben Mildenhall, Sara Fridovich-Keil, Nithin Raghavan, Utkarsh Singhal, Ravi Ramamoorthi, Jonathan Barron, and Ren Ng.
\newblock Fourier features let networks learn high frequency functions in low dimensional domains.
\newblock \emph{Advances in Neural Information Processing Systems (NeurIPS)}, 33:\penalty0 7537--7547, 2020.

\bibitem[van~den Burg \& Williams(2021)van~den Burg and Williams]{van2021memorization}
Gerrit van~den Burg and Chris Williams.
\newblock On memorization in probabilistic deep generative models.
\newblock \emph{Advances in Neural Information Processing Systems (NeurIPS)}, 34:\penalty0 27916--27928, 2021.

\bibitem[Vaswani et~al.(2017)Vaswani, Shazeer, Parmar, Uszkoreit, Jones, Gomez, Kaiser, and Polosukhin]{vaswani2017attention}
Ashish Vaswani, Noam Shazeer, Niki Parmar, Jakob Uszkoreit, Llion Jones, Aidan~N Gomez, {\L}ukasz Kaiser, and Illia Polosukhin.
\newblock Attention is all you need.
\newblock \emph{Advances in neural information processing systems (NeurIPS)}, 30, 2017.

\bibitem[Vignac et~al.(2022)Vignac, Krawczuk, Siraudin, Wang, Cevher, and Frossard]{vignac2022digress}
Clement Vignac, Igor Krawczuk, Antoine Siraudin, Bohan Wang, Volkan Cevher, and Pascal Frossard.
\newblock Digress: Discrete denoising diffusion for graph generation.
\newblock In \emph{International Conference on Learning Representations (ICLR)}, 2022.

\bibitem[Vincent(2011)]{vincent2011connection}
Pascal Vincent.
\newblock A connection between score matching and denoising autoencoders.
\newblock \emph{Neural computation}, 23\penalty0 (7):\penalty0 1661--1674, 2011.

\bibitem[Wang et~al.(2024)Wang, Shen, Tong, Zhang, and Kawaguchi]{wang2024stronger}
Haonan Wang, Qianli Shen, Yao Tong, Yang Zhang, and Kenji Kawaguchi.
\newblock The stronger the diffusion model, the easier the backdoor: Data poisoning to induce copyright breaches without adjusting finetuning pipeline.
\newblock \emph{arXiv preprint arXiv:2401.04136}, 2024.

\bibitem[Webster et~al.(2021)Webster, Rabin, Simon, and Jurie]{webster2021person}
Ryan Webster, Julien Rabin, Loic Simon, and Frederic Jurie.
\newblock This person (probably) exists. identity membership attacks against gan generated faces.
\newblock \emph{arXiv preprint arXiv:2107.06018}, 2021.

\bibitem[Wen et~al.(2023)Wen, Liu, Chen, and Lyu]{wen2023detecting}
Yuxin Wen, Yuchen Liu, Chen Chen, and Lingjuan Lyu.
\newblock Detecting, explaining, and mitigating memorization in diffusion models.
\newblock In \emph{The Twelfth International Conference on Learning Representations}, 2023.

\bibitem[Xu et~al.(2022)Xu, Yu, Song, Shi, Ermon, and Tang]{xu2022geodiff}
Minkai Xu, Lantao Yu, Yang Song, Chence Shi, Stefano Ermon, and Jian Tang.
\newblock Geodiff: A geometric diffusion model for molecular conformation generation.
\newblock In \emph{International Conference on Learning Representations (ICLR)}, 2022.

\bibitem[Yi et~al.(2023)Yi, Sun, and Li]{yi2023generalization}
Mingyang Yi, Jiacheng Sun, and Zhenguo Li.
\newblock On the generalization of diffusion model.
\newblock \emph{arXiv preprint arXiv:2305.14712}, 2023.

\bibitem[Yoon et~al.(2023)Yoon, Choi, Kwon, and Ryu]{yoon2023diffusion}
TaeHo Yoon, Joo~Young Choi, Sehyun Kwon, and Ernest~K Ryu.
\newblock Diffusion probabilistic models generalize when they fail to memorize.
\newblock In \emph{ICML 2023 Workshop on Structured Probabilistic Inference \& Generative Modeling}, 2023.

\bibitem[Zhang et~al.(2017)Zhang, Bengio, Hardt, Recht, and Vinyals]{zhang2017understanding}
Chiyuan Zhang, Samy Bengio, Moritz Hardt, Benjamin Recht, and Oriol Vinyals.
\newblock Understanding deep learning requires rethinking generalization.
\newblock In \emph{International Conference on Learning Representations (ICLR)}, 2017.

\bibitem[Zhang et~al.(2023)Zhang, Zhou, Lu, Guo, Shen, and Qu]{zhang2023emergence}
Huijie Zhang, Jinfan Zhou, Yifu Lu, Minzhe Guo, Liyue Shen, and Qing Qu.
\newblock The emergence of reproducibility and consistency in diffusion models.
\newblock \emph{arXiv preprint arXiv:2310.05264}, 2023.

\bibitem[Zhao et~al.(2023)Zhao, Pang, Du, Yang, Cheung, and Lin]{zhao2023recipe}
Yunqing Zhao, Tianyu Pang, Chao Du, Xiao Yang, Ngai-Man Cheung, and Min Lin.
\newblock A recipe for watermarking diffusion models.
\newblock \emph{arXiv preprint arXiv:2303.10137}, 2023.

\end{thebibliography}
\bibliographystyle{tmlr}

\appendix
\clearpage
\appendix

\section{Optimal Solution of Diffusion Models}

\subsection{Derivation of the Theoretical Optimum}\label{appendix_optimal}

In this section, we prove the close form of the optimal score model defined in~\eqref{optimal}. Firstly, the empirical denoising score matching (DSM) objective~\citep{vincent2011connection} is: 
\begin{align}\label{dsm}
    \mathcal{J}_{\textrm{DSM}}(\theta)&\triangleq\frac{1}{2N}\sum_{n=1}^{N}\mathbb{E}_{t\sim[0, T]}\mathbb{E}_ {\epsilon\sim \mathcal{N}(\mathbf{0},\mathbf{I})}\left[\lambda(t)\left\|\boldsymbol{s}_{\theta}(z_t, t)+\frac{\epsilon}{\sigma_{t}}\right\|_{2}^{2}\right]\\
    &= \mathbb{E}_{t\sim[0, T]}\lambda(t)\mathbb{E}_ {\epsilon\sim \mathcal{N}(\mathbf{0},\mathbf{I})}\left[\frac{1}{2N}\sum_{n=1}^N\left\|\boldsymbol{s}_{\theta}(z_t, t)+\frac{\epsilon}{\sigma_{t}}\right\|_{2}^{2}\right] \\
    &=\mathbb{E}_{t\sim[0, T]}\left[\lambda(t) \int \frac{1}{2N}\sum_{n=1}^N\left\|\boldsymbol{s}_{\theta}(z_t, t)+\frac{\epsilon}{\sigma_{t}}\right\|_{2}^{2}\mathcal{N}(\epsilon;\mathbf{0}, \mathbf{I}) d\epsilon\right]\textrm{.}
\end{align}

Compared to~\eqref{eqn:dsm}, we add a positive weighting function $\lambda(t)>0$, which is normally used in the training of diffusion models~\citep{song2021score}. Since $z_t=\alpha_{t}x_{n}+\sigma_{t}\epsilon$, we have $\epsilon=-\frac{\alpha_{t}x_{n}-z_t}{\sigma_{t}}$. Therefore, the derivative of $\epsilon$ w.r.t. $z_t$ is $d\epsilon=\frac{d z_t}{\sigma_t}$. Then
\begin{align}
    \mathcal{J}_{\textrm{DSM}}(\theta)&=\mathbb{E}_{t\sim[0, T]}\left[\lambda(t) \int \frac{1}{2N}\sum_{n=1}^N\left\|\boldsymbol{s}_{\theta}(z_t, t)-\frac{\alpha_tx_n-z_t}{\sigma_{t}^2}\right\|_{2}^{2} \mathcal{N}(z_t;\alpha_tx_n, \sigma_t^2\mathbf{I})\sigma_t d z_t\right]\\
    &=\mathbb{E}_{t\sim[0, T]}\left[\lambda(t) \int \mathcal{J}_{\textrm{DSM}}(\theta, z_t, t)dz_t\right]\textrm{.}
\end{align}
To minimize the empirical DSM objective $\mathcal{J}_{\textrm{DSM}}(\theta)$, we can minimize $\mathcal{J}_{\textrm{DSM}}(\theta, z_t, t)$ given each $z_t$ since $\lambda(t)>0$. The minimization of $\mathcal{J}_{\textrm{DSM}}(\theta, z_t, t)$ is a convex optimization problem w.r.t. the score $\boldsymbol{s}_\theta(z_t, t)$, which can be solved by taking the gradient w.r.t. $\boldsymbol{s}_\theta(z_t, t)$:
\begin{align}
    \mathbf{0} &= \nabla_{\boldsymbol{s}_\theta(z_t, t)}\left[\frac{1}{2N}\sum_{n=1}^N\left\|\boldsymbol{s}_{\theta}(z_t, t)-\frac{\alpha_tx_n-z_t}{\sigma_{t}^2}\right\|_{2}^{2}\mathcal{N}(z_t;\alpha_tx_n, \sigma_t^2\mathbf{I})\sigma_t\right]\\
    &=\sum_{n=1}^N2\left[\boldsymbol{s}(z_t, t)-\frac{\alpha_tx_n-z_t}{\sigma_{t}^2}\right]\mathcal{N}(z_t;\alpha_tx_n, \sigma_t^2\mathbf{I})\\
    &=\left[\sum_{n=1}^N\mathcal{N}(z_t;\alpha_tx_n, \sigma_t^2\mathbf{I})\right]\boldsymbol{s}_\theta(z_t, t)-\sum_{n=1}^N\mathcal{N}(z_t;\alpha_tx_n, \sigma_t^2\mathbf{I})\frac{\alpha_tx_n-z_t}{\sigma_{t}^2}\textrm{.}
\end{align}
The optimal diffusion model can be written
\begin{align}
    \boldsymbol{s}^*(z_t, t)&=\frac{\sum_{n=1}^N\mathcal{N}(z_t;\alpha_tx_n, \sigma_t^2\mathbf{I})\frac{\alpha_tx_n-z_t}{\sigma_{t}^2}}{\sum_{n'=1}^N\mathcal{N}(z_t;\alpha_tx_{n'}, \sigma_t^2\mathbf{I})}\\
    &=\frac{\sum_{n=1}^N\exp\left(-\frac{\left\|\alpha_tx_n-z_t\right\|_2^2}{2\sigma_t^2}\right)\frac{\alpha_tx_n-z_t}{\sigma_{t}^2}}{\sum_{n'=1}^N\exp\left(-\frac{\left\|\alpha_tx_{n'}-z_t\right\|_2^2}{2\sigma_t^2}\right)}\\
    &=\sum_{n=1}^{N}\mathbb{S}\left(-\frac{\left\|\alpha_tx_n-z_t\right\|_{2}^{2}}{2\sigma_{t}^{2}}\right)\cdot\frac{\alpha_tx_n-z_t}{\sigma_t^2}\textrm{,}
\end{align}
where $\mathbb{S}$ refers to the softmax operation.

It is noted that when $\boldsymbol{s}_\theta(z_t, t)$ is parameterized using a neural network $\theta$, the objective in equation~\ref{dsm} is typically highly non-convex w.r.t. $\theta$. Therefore, achieving this theoretical optimum necessitates a model $\theta$ with sufficient capacity. Firstly, this theoretical optimum may lie outside the solution space of $\theta$'s model family. Even if the solution space encompasses the theoretical optimum, reaching it remains challenging due to the non-convex nature of the optimization in equation~\ref{dsm} w.r.t. model parameters $\theta$. The optimization may get stuck in the local optimum.

Given the existence of theoretical optimum, we find that the empirical DSM objective can be rewritten
\begin{align}
    \mathcal{J}_{\textrm{DSM}}(\theta)=\frac{1}{2N}\sum_{n=1}^{N}\mathbb{E}_{t,\epsilon}\left[\lambda(t)\left\|\boldsymbol{s}_{\theta}(\alpha_{t}x_{n}+\sigma_{t}\epsilon, t)-\boldsymbol{s}^*(\alpha_{t}x_{n}+\sigma_{t}\epsilon, t)\right\|_{2}^{2}\right]+C\textrm{,}
\end{align}
where $C=\mathbb{E}_{t}\left[\lambda(t) \int \frac{1}{2N}\sum_{n=1}^N\left\|\boldsymbol{s}^*(z_t, t)-\frac{\alpha_tx_n-z_t}{\sigma_{t}^2}\right\|_{2}^{2}\mathcal{N}(z_t;\alpha_tx_n, \sigma_t^2\mathbf{I})\sigma_t d z_t\right]$ is a constant value without involvement of the trained diffusion model $\theta$. This equivalence shows that diffusion models are trained to approximate the theoretical optimum. The proof is shown as below
\begin{align}
    &\mathcal{J}_{\textrm{DSM}}(\theta)\\
    =&\mathbb{E}_{t}\left[\lambda(t) \int \frac{1}{2N}\sum_{n=1}^N\left\|\boldsymbol{s}_{\theta}(z_t, t)-\frac{\alpha_tx_n-z_t}{\sigma_{t}^2}\right\|_{2}^{2}\mathcal{N}(z_t;\alpha_tx_n, \sigma_t^2\mathbf{I})\sigma_t d z_t\right]\\
    =&\mathbb{E}_{t}\left[\lambda(t) \int \frac{1}{2N}\sum_{n=1}^N\left\|\boldsymbol{s}_{\theta}(z_t, t)-\boldsymbol{s}^*(z_t, t)+\boldsymbol{s}^*(z_t, t)-\frac{\alpha_tx_n-z_t}{\sigma_{t}^2}\right\|_{2}^{2}\mathcal{N}(z_t;\alpha_tx_n, \sigma_t^2\mathbf{I})\sigma_t d z_t\right]\\
    =&\mathbb{E}_{t}\left[\lambda(t) \int \frac{1}{2N}\sum_{n=1}^N\left\|\boldsymbol{s}_{\theta}(z_t, t)-\boldsymbol{s}^*(z_t, t)\right\|_{2}^{2}\mathcal{N}(z_t;\alpha_tx_n, \sigma_t^2\mathbf{I})\sigma_t d z_t\right]\\\nonumber
    -&\mathbb{E}_{t}\left[\lambda(t) \int \frac{1}{2N}\sum_{n=1}^N2\left(\boldsymbol{s}_{\theta}(z_t, t)-\boldsymbol{s}^*(z_t, t)\right)\left(\boldsymbol{s}^*(z_t, t)-\frac{\alpha_tx_n-z_t}{\sigma_{t}^2}\right)\mathcal{N}(z_t;\alpha_tx_n, \sigma_t^2\mathbf{I})\sigma_t d z_t\right]\\\nonumber
    +&\mathbb{E}_{t}\left[\lambda(t) \int \frac{1}{2N}\sum_{n=1}^N\left\|\boldsymbol{s}^*(z_t, t)-\frac{\alpha_tx_n-z_t}{\sigma_{t}^2}\right\|_{2}^{2}\mathcal{N}(z_t;\alpha_tx_n, \sigma_t^2\mathbf{I})\sigma_t d z_t\right]\\
    =&\mathbb{E}_{t}\left[\lambda(t) \int \frac{1}{2N}\sum_{n=1}^N\left\|\boldsymbol{s}_{\theta}(z_t, t)-\boldsymbol{s}^*(z_t, t)\right\|_{2}^{2}\mathcal{N}(z_t;\alpha_tx_n, \sigma_t^2\mathbf{I})\sigma_t d z_t\right] + C\\
    =&\frac{1}{2N}\sum_{n=1}^{N}\mathbb{E}_{t,\epsilon}\left[\lambda(t)\left\|\boldsymbol{s}_{\theta}(\alpha_{t}x_{n}+\sigma_{t}\epsilon, t)-\boldsymbol{s}^*(\alpha_{t}x_{n}+\sigma_{t}\epsilon, t)\right\|_{2}^{2}\right]+C\textrm{.}
\end{align}
The above equation holds since
\begin{align}
    &\int \frac{1}{2N}\sum_{n=1}^N2\left(\boldsymbol{s}_{\theta}(z_t, t)-\boldsymbol{s}^*(z_t, t)\right)\left(\boldsymbol{s}^*(z_t, t)-\frac{\alpha_tx_n-z_t}{\sigma_{t}^2}\right)\mathcal{N}(z_t;\alpha_tx_n, \sigma_t^2\mathbf{I})\sigma_t d z_t\\
    =&\frac{1}{N}\left(\boldsymbol{s}_{\theta}(z_t, t)-\boldsymbol{s}^*(z_t, t)\right)\int \sum_{n=1}^N \mathcal{N}(z_t;\alpha_tx_n, \sigma_t^2\mathbf{I})\sigma_t \left(\boldsymbol{s}^*(z_t, t)-\frac{\alpha_tx_n-z_t}{\sigma_{t}^2}\right) d z_t\\
    =& 0\textrm{.}
\end{align}

% \begin{align}
%     &\int \frac{1}{2N}\sum_{n=1}^N2\left(\boldsymbol{s}_{\theta}(z_t, t)-\boldsymbol{s}^*(z_t, t)\right)\left(\boldsymbol{s}^*(z_t, t)-\frac{\alpha_tx_n-z_t}{\sigma_{t}^2}\right)\mathcal{N}(z_t;\alpha_tx_n, \sigma_t^2\mathbf{I})\sigma_t d z_t\\
%     =&\int \frac{1}{N}\sum_{n=1}^N\left(\boldsymbol{s}_{\theta}(z_t, t)-\boldsymbol{s}^*(z_t, t)\right)\left(\frac{\sum_{n=1}^N\mathcal{N}(z_t;\alpha_tx_n, \sigma_t^2\mathbf{I})\frac{\alpha_tx_n-z_t}{\sigma_{t}^2}}{\sum_{n'=1}^N\mathcal{N}(z_t;\alpha_tx_{n'}, \sigma_t^2\mathbf{I})}-\frac{\alpha_tx_n-z_t}{\sigma_{t}^2}\right)\mathcal{N}(z_t;\alpha_tx_n, \sigma_t^2\mathbf{I})\sigma_t d z_t\\
%     =&\frac{1}{N}\left(\boldsymbol{s}_{\theta}(z_t, t)-\boldsymbol{s}^*(z_t, t)\right)\int \sum_{n=1}^N \mathcal{N}(z_t;\alpha_tx_n, \sigma_t^2\mathbf{I})\sigma_t \left(\frac{\sum_{n=1}^N\mathcal{N}(z_t;\alpha_tx_n, \sigma_t^2\mathbf{I})\frac{\alpha_tx_n-z_t}{\sigma_{t}^2}}{\sum_{n=1}^N\mathcal{N}(z_t;\alpha_tx_n, \sigma_t^2\mathbf{I})}-\frac{\alpha_tx_n-z_t}{\sigma_{t}^2}\right) d z_t\\
%     =& \frac{1}{N}\left(\boldsymbol{s}_{\theta}(z_t, t)-\boldsymbol{s}^*(z_t, t)\right)\int \left(\sum_{n=1}^N\mathcal{N}(z_t;\alpha_tx_n, \sigma_t^2\mathbf{I})\frac{\alpha_tx_n-z_t}{\sigma_{t}} - \sum_{n=1}^N\mathcal{N}(z_t;\alpha_tx_n, \sigma_t^2\mathbf{I})\frac{\alpha_tx_n-z_t}{\sigma_{t}} \right) d z_t \\
%     =& 0
% \end{align}

In \citet{karras2022elucidating}, the authors have derived the optimal denoised function, while in \citet{yi2023generalization}, the authors provided the closed form of the optimal DDPM~\citep{ho2020denoising}. Therefore, we show the equivalence of our~\eqref{optimal} with the above two forms. 

For DDPM, which is trained under noise prediction objective, \citet{kingma2023understanding} gave the transformation between the two parameterizations:
\begin{equation}
    \boldsymbol{s}_\theta(z_t, t)=-\frac{\boldsymbol{\epsilon}_\theta(z_t, t)}{\sigma_t}\textrm{.}
\end{equation}

Therefore, the optimal DDPM can be represented as
\begin{align}
    \boldsymbol{\epsilon}^*(z_t, t)=-\sigma_t\boldsymbol{s}^*(z_t, t)&=\frac{\sum_{n=1}^N\mathcal{N}(z_t;\alpha_tx_n, \sigma_t^2\mathbf{I})\frac{z_t-\alpha_tx_n}{\sigma_{t}}}{\sum_{n'=1}^N\mathcal{N}(z_t;\alpha_tx_{n'}, \sigma_t^2\mathbf{I})}\\
    &=\frac{z_t}{\sigma_t}-\frac{\alpha_t}{\sigma_t}\sum_{n=1}^N\frac{\exp\left(-\frac{\left\|\alpha_tx_n-z_t\right\|_2^2}{2\sigma_t^2}\right)x_n}{\sum_{n'=1}^N \exp\left(-\frac{\left\|\alpha_tx_{n'}-z_t\right\|_2^2}{2\sigma_t^2}\right)}\textrm{.}
\end{align}
In DDPM~\citep{ho2020denoising}, the forward process is defined as  $z_t=\sqrt{\bar{\alpha}_t}x+\sqrt{1-\bar{\alpha}_t}\epsilon$, so the closed form for optimal DDPM is
\begin{equation}
    \boldsymbol{\epsilon}^*(z_t, t)=\frac{z_t}{\sqrt{1-\bar{\alpha}_t}}-\frac{\bar{\alpha}_t}{\sqrt{1-\bar{\alpha}_t}}\sum_{n=1}^N\frac{\exp\left(-\frac{\left\|z_t-\bar{\alpha}_tx_n\right\|_2^2}{2(1-\bar{\alpha}_t)}\right)x_n}{\sum_{n'=1}^N \exp\left(-\frac{\left\|z_t-\bar{\alpha}_tx_{n'}\right\|_2^2}{2(1-\bar{\alpha}_t)}\right)}\textrm{,}
\end{equation}
which is the same as Theorem~2 in \citet{yi2023generalization}. 

For denoised function, which is trained under data prediction objective, \citet{kingma2023understanding} gave:
\begin{equation}
    \boldsymbol{s}_\theta(z_t, t)=-\frac{z_t-\alpha_tD_{\theta}(z_t, t)}{\sigma_t^2}\textrm{.}
\end{equation}
Therefore, the optimal denoised function can be represented as
\begin{align}
    D^*(z_t, t)=\frac{\sigma_t^2\boldsymbol{s}^*(z_t, t)+z_t}{\alpha_t}=\frac{\sum_{n=1}^N\mathcal{N}(z_t;\alpha_tx_n, \sigma_t^2\mathbf{I})x_n}{\sum_{n'=1}^N\mathcal{N}(z_t;\alpha_tx_{n'}, \sigma_t^2\mathbf{I})}\textrm{.}
\end{align}
In EDM~\citep{karras2022elucidating}, the forward process is defined as $z_t=x+\sigma_t\epsilon$, so the closed form for optimal denoised function is:
\begin{equation}
    D^*(z_t, t)=\frac{\sum_{n=1}^N\mathcal{N}(z_t;x_n, \sigma_t^2\mathbf{I})x_n}{\sum_{n'=1}^N\mathcal{N}(z_t;x_{n'}, \sigma_t^2\mathbf{I})}\textrm{.}
\end{equation}

which is the same as Eq.~(57) in \citet{karras2022elucidating}. 

% Besides DSM, the diffusion models can be trained by other objectives, including noise prediction, data prediction, velocity prediction, etc. The resulting diffusion models though have different parameterizations, their optimal solutions are equivalent in essence.

% \cite{kingma2023understanding} showed the transformation among different paramerization

 % &= \mathbb{E}_{t\sim[0, T]}\lambda(t) \widehat{\mathcal{J}_{\textrm{DSM}}(\theta, t)}, 
 % \mathbb{E}_t\lambda(t)\left[\mathbb{E}_{\epsilon}\frac{1}{2N}\left\|\boldsymbol{s}_{\theta}(\alpha_{t}x_{n}+\sigma_{t}\epsilon, t)+\frac{\epsilon}{\sigma_{t}}\right\|_{2}^{2}\right]

\subsection{Backward Process of the Optimal Diffusion Model}\label{appendix_backward}

We analyze the memorization behavior of the optimal diffusion model $\boldsymbol{s}^*(z_t, t)$ defined in Eq.~\eqref{optimal} through the lens of backward process. As shown in \citet{kingma2021variational}, the backward process of our defined diffusion models is governed by the following stochastic differential equation (SDE):

\begin{equation}
    dz_t=\left[f(t)z_t-g^2(t)\boldsymbol{s}_\theta(z_t, t)\right]dt+g(t)dW_t\textrm{,}
\end{equation}
where $W_t$ is a standard Brownian motion, and $f(t)$ and $g(t)$ follow
\begin{equation}
    f(t)=\frac{d\log \alpha_t}{dt}\textrm{,}\,\,\,\,\,g^2(t)=\frac{d\sigma_t^2}{dt}-2\frac{d\log\alpha_t}{dt}\sigma_t^2\textrm{.}
\end{equation}

Besides the SDE, \citet{song2021score} showed that there exists an ordinary differential equation (ODE) for deterministic backward process

\begin{equation}
    dz_t=\left[f(t)z_t-\frac{1}{2}g^2(t)\boldsymbol{s}_\theta(z_t, t)\right]dt\textrm{.}
\end{equation}

We first show how to adopt the above ODE to generate samples using the optimal score model defined in~\eqref{optimal}. Specifically, we sample multiple time steps $0=t_0<\xi=t_1<...<t_n=T$, where $\xi$ refers to a small value close to 0 and $T>0$ represents the maximum time step. For simplicity, we consider Euler solver, and then we have the following update rule
\begin{align}
    z_{t_n}=z_{t_{n+1}}+\left(\frac{d\log\alpha_t}{dt}z_{t}-\frac{1}{2}\left(\frac{d\sigma_t^2}{dt}-2\frac{d\log\alpha_t}{dt}\sigma_t^2\right)\boldsymbol{s}_\theta(z_t, t)\right)\bigg|_{t=t_{n+1}}(t_{n}-t_{n+1})\textrm{.}
\end{align}

We also use Euler method to approximate $\frac{d\log\alpha_t}{dt}$ and $\frac{d\sigma_t^2}{dt}$ considering
\begin{equation}
    \lim_{t_{n}-{t_{n+1}}\rightarrow 0^{-}}\frac{d\log\alpha_t}{dt}\bigg|_{t=t_{n+1}}=\lim_{t_{n}-{t_{n+1}}\rightarrow 0^{-}}\frac{1}{\alpha_t}\frac{d\alpha_t}{dt}\bigg|_{t=t_{n+1}}=\lim_{t_{n}-{t_{n+1}}\rightarrow 0^{-}}\frac{1}{\alpha_{t_{n+1}}}\frac{\alpha_{t_n}-\alpha_{t_{n+1}}}{t_n-t_{n+1}}\textrm{,}
\end{equation}
\begin{equation}
    \lim_{t_{n}-{t_{n+1}}\rightarrow 0^{-}}\frac{d\sigma_t^2}{dt}\bigg|_{t=t_{n+1}}=\lim_{t_{n}-{t_{n+1}}\rightarrow 0^{-}}2\sigma_t\frac{d\sigma_t}{dt}\bigg|_{t=t_{n+1}}=\lim_{t_{n}-{t_{n+1}}\rightarrow 0^{-}}2\sigma_{t_{n+1}}\frac{\sigma_{t_{n}}-\sigma_{t_{n+1}}}{t_n-t_{n+1}}\textrm{.}
\end{equation}

Then we have
\begin{align}
    z_{t_n}&=\frac{\alpha_{t_n}}{\alpha_{t_{n+1}}}z_{t_{n+1}}-\left(\sigma_{t_{n+1}}(\sigma_{t_{n}}-\sigma_{t_{n+1}})-\frac{(\alpha_{t_{n}}-\alpha_{t_{n+1}})\sigma^2_{t_{n+1}}}{\alpha_{t_{n+1}}}\right)\boldsymbol{s}_\theta(z_{t_{n+1}}, t_{n+1})\\
    &=\frac{\alpha_{t_n}}{\alpha_{t_{n+1}}}z_{t_{n+1}}-\left(\sigma_{t_{n+1}}\sigma_{t_{n}}-\frac{\alpha_{t_{n}}\sigma^2_{t_{n+1}}}{\alpha_{t_{n+1}}}\right)\boldsymbol{s}_\theta(z_{t_{n+1}}, t_{n+1})\textrm{.}
\end{align}
For $t_{0}=0$, we know that $\alpha_{0}=1$ and $\sigma_{0}=0$. $z_0$ refers to the generated samples, and we have
\begin{align}
    z_0&=\frac{z_{\xi}}{\alpha_{\xi}}+\frac{\sigma^2_{\xi}}{\alpha_{\xi}}\boldsymbol{s}_\theta(z_{\xi}, \xi)\\
    &=\frac{z_{\xi}}{\alpha_{\xi}}+\frac{\sigma^2_{\xi}}{\alpha_{\xi}} \sum_{n=1}^N\frac{\exp\left(-\frac{\left\|\alpha_\xi x_n-z_\xi\right\|_2^2}{2\sigma_\xi^2}\right)}{\sum_{n'=1}^N\exp\left(-\frac{\left\|\alpha_\xi x_{n'}-z_\xi\right\|_2^2}{2\sigma_\xi^2}\right)}\cdot\frac{\alpha_\xi x_n-z_\xi}{\sigma_{\xi}^2}\\
    &=\sum_{n=1}^N\frac{\exp\left(-\frac{\left\|\alpha_\xi x_n-z_\xi\right\|_2^2}{2\sigma_\xi^2}\right)}{\sum_{n'=1}^N\exp\left(-\frac{\left\|\alpha_\xi x_{n'}-z_\xi\right\|_2^2}{2\sigma_\xi^2}\right)}\cdot x_n\textrm{.}
\end{align}

% In the backward process, we consider the probabilistic ordinary differential equation (ODE) in EDM \cite{karras2022elucidating}
% % \begin{equation}
% %     dz=\left[f(t)z-\frac{1}{2}g(t)^2\nabla_z\log p(z, t)\right]dt
% % \end{equation}
% \begin{equation}
%     dz=-t\nabla_{z_t}\log p(z_t, t)dt=-t\boldsymbol{s}^*(z_t, t)dt\textrm{,}
% \end{equation}
% where $z_t=x+t\epsilon$. When using Euler solver, we have the following update rule
% \begin{equation}
%     z_{t_n}=z_{t_{n+1}}-(t_n-t_{n+1})t_{n+1}\boldsymbol{s}^*(z_{t_{n+1}}, t_{n+1})\textrm{.}
% \end{equation}
% The sampled time steps meet $0=t_0<\xi=t_1<...<t_n=T$, where $\xi$ refers to a small value closed to 0 and $T>0$ represents the maximum time step. For $t_0=0$, $z_0$ refers to the generated samples, and we have

% \begin{align}
%     z_0&=z_{t_1}+t_1^2\boldsymbol{s}^*(z_{t_1}, t_1)\\
%     &=z_\xi+\xi^2\frac{\sum_{n=1}^N\mathcal{N}(z_\xi;x_n, \xi^2\mathbf{I})\frac{x_n-z_\xi}{\xi^2}}{\sum_{n=1}^N\mathcal{N}(z_\xi;x_n, \xi^2\mathbf{I})}\\
%     &=\frac{\sum_{n=1}^N\exp\left(-\frac{\left\|z_\xi-x_n\right\|_2^2}{2\xi^2}\right)x_n}{\sum_{n=1}^N\exp\left(-\frac{\left\|z_\xi-x_n\right\|_2^2}{2\xi^2}\right)}\textrm{.}
% \end{align}

From the above equation, we conclude that the generated samples by the optimal diffusion model are the linear combinations of training samples in $\mathcal{D}$.

Next, we consider a discrete distribution, and suppose $z=\frac{z_{\xi}}{\alpha_{\xi}}$, then
\begin{equation}
    p(z=x_n, \xi)=\frac{\exp\left(-\frac{\left\|z-x_n\right\|_2^2}{2\alpha_{\xi}^2\sigma_{\xi}^2}\right)}{\sum_{n'=1}^N\exp\left(-\frac{\left\|z-x_{n'}\right\|_2^2}{2\alpha_{\xi}^2\sigma_{\xi}^2}\right)}\textrm{,} \,\,n=1\textrm{,}2\textrm{,}...\textrm{,}N\textrm{.}
\end{equation}

Suppose $x_{m}=\textrm{NN}_1(z, \mathcal{D})=\arg\min_{x\in\mathcal{D}} ||z-x||_2^2$, then we have
\begin{equation}
    \left\|z-x_m\right\|_2^2-\left\|z-x_{n\neq m}\right\|_2^2<0\textrm{.}
\end{equation}

When $\xi\rightarrow0^{+}$, $\eta = \alpha_\xi \sigma_\xi \to 0^+$, then we have
\begin{align}
\lim_{\xi\rightarrow0^{+}}p(z=x_m,\xi)&=\lim_{\eta\rightarrow0^{+}}\frac{\exp\left(-\frac{\left\|z-x_m\right\|_2^2}{2\eta^2}\right)}{\sum_{n'=1}^N\exp\left(-\frac{\left\|z-x_{n'}\right\|_2^2}{2\eta^2}\right)}\\
    &=\lim_{\eta\rightarrow0^{+}}\frac{1}{\sum_{n'\neq m}^N\exp\left(-\frac{\left\|z-x_{n'}\right\|_2^2-\left\|z-x_m\right\|_2^2}{2\eta^2}\right) + 1}\\
    &=\frac{1}{\sum_{n'\neq m}^N\lim_{\eta\rightarrow0^{+}}\exp\left(-\frac{\left\|z-x_{n'}\right\|_2^2-\left\|z-x_m\right\|_2^2}{2\eta^2}\right) + 1}\\
    &=\frac{1}{\sum_{n'\neq m}^N\lim_{\eta\rightarrow+\infty}\exp\left(\frac{\eta^2}{2}\left(\left\|z-x_m\right\|_2^2-\left\|z-x_{n'}\right\|_2^2\right)\right) + 1}\\
    &=1\textrm{.}
\end{align}
Then
\begin{equation}
    \lim_{\xi\rightarrow0^{+}}\left(p(z=x_m,\xi)+\sum_{n'\neq m}^Np(z=x_{n'},\xi)\right)=1\textrm{,}
\end{equation}
\begin{equation}
    \sum_{n'\neq m}^N\lim_{\xi\rightarrow0^{+}}p(z=x_{n'},\xi)=0\textrm{.}
\end{equation}
Given $p(z=x_i,\xi)\geq0$,
\begin{equation}
    \lim_{\xi\rightarrow0^{+}}p(z=x_{n\neq m},\xi)=0\textrm{.}
\end{equation}
\begin{equation}
\lim_{\xi\rightarrow0^{+}}z_0=\lim_{\xi\rightarrow0^{+}}\mathbb{E}_{p(z=x_n, \xi)}\left[z\right]=\textrm{NN}_1(z, \mathcal{D})\textrm{.}
\end{equation}
From the above analysis, we conclude that when $t_1=\xi$ is closed to 0, the probabilistic ODE solver returns a training sample in $\mathcal{D}$.

Next, we consider an SDE solver, then the update rule is

\begin{align}
    z_{t_n}&=z_{t_{n+1}}+\left(\frac{d\log\alpha_t}{dt}z_{t}-\left(\frac{d\sigma_t^2}{dt}-2\frac{d\log\alpha_t}{dt}\sigma_t^2\right)\boldsymbol{s}_\theta(z_t, t)\right)\bigg|_{t=t_{n+1}}(t_{n}-t_{n+1})\\\nonumber
    &+\sqrt{\frac{d\sigma_t^2}{dt}-2\frac{d\log\alpha_t}{dt}\sigma_t^2}\bigg|_{t=t_{n+1}}(t_n-t_{n+1})\epsilon\\
    &=\frac{\alpha_{t_n}}{\alpha_{t_{n+1}}}z_{t_{n+1}}-2\left(\sigma_{t_{n+1}}\sigma_{t_{n}}-\frac{\alpha_{t_{n}}\sigma^2_{t_{n+1}}}{\alpha_{t_{n+1}}}\right)\boldsymbol{s}_\theta(z_{t_{n+1}}, t_{n+1})\\\nonumber
    &+\sqrt{2\left(\sigma_{t_{n+1}}\sigma_{t_{n}}-\frac{\alpha_{t_{n}}\sigma^2_{t_{n+1}}}{\alpha_{t_{n+1}}}\right)\left(t_n-t_{n+1}\right)}\epsilon\textrm{,}
\end{align}
where $\epsilon\sim\mathcal{N}(\mathbf{0},\mathbf{I})$ is a Gaussian noise. Similarly, we consider the update step at $t_0=0$
\begin{align}
    z_0=\frac{z_{\xi}}{\alpha_{\xi}}+2\frac{\sigma^2_{\xi}}{\alpha_{\xi}}\boldsymbol{s}_\theta(z_{\xi}, \xi)+\sqrt{2\frac{\sigma_{\xi}^2\xi}{\alpha_{\xi}}}\epsilon=2\sum_{n=1}^N\frac{\exp\left(-\frac{\left\|\alpha_\xi x_n-z_\xi\right\|_2^2}{2\sigma_\xi^2}\right)}{\sum_{n'=1}^N\exp\left(-\frac{\left\|\alpha_\xi x_{n'}-z_\xi\right\|_2^2}{2\sigma_\xi^2}\right)}\cdot x_n-\frac{z_{\xi}}{\alpha_{\xi}}+\sqrt{2\frac{\sigma_{\xi}^2\xi}{\alpha_{\xi}}}\epsilon\textrm{.}
\end{align}
When $\xi\rightarrow 0^{+}$
\begin{align}
    \lim_{\xi\rightarrow 0^{+}}z_0=2\textrm{NN}_1\left(\frac{z_{\xi}}{\alpha_{\xi}}, \mathcal{D}\right)-\lim_{\xi\rightarrow 0^{+}}\frac{z_{\xi}}{\alpha_\xi}\textrm{.}
\end{align}
Consider $\lim_{\xi\rightarrow 0^{+}}\frac{z_{\xi}}{\alpha_\xi}=\lim_{\xi\rightarrow 0^{+}}z_0$
\begin{equation}
    \lim_{\xi\rightarrow 0^{+}}z_0=\textrm{NN}_1\left(\frac{z_{\xi}}{\alpha_{\xi}}, \mathcal{D}\right)\textrm{.}
\end{equation}

To summarize, through our analysis, we find that the optimal diffusion model always replicates training data through the backward process.

\subsection{Optimal Class-conditioned Diffusion Model}\label{appendix_optimal_cond}

In the above, we provide the derivation of the optimal diffusion model for unconditional generation under the assumption of empirical data distribution $x\sim\frac{1}{N}\sum_{n=1}^N\delta(x-x_n)$. Next, we consider the scenario of class-conditional generation. The dataset can be represented as $\mathcal{D}\triangleq\{x_{n}, y_{n}\}_{n=1}^{N}, y_n\in\{1,2,...,C\}$, where $C$ is the number of classes. Then the empirical joint distribution of data and label is $x,y\sim\frac{1}{N}\sum_{n=1}^N\delta(x-x_n)\delta(y-y_n)$. 

For class-conditional generation, the empirical DSM objective is written
\begin{align}
    \mathcal{J}_{\textrm{DSM}}(\theta)&=\frac{1}{2N}\sum_{n=1}^{N}\mathbb{E}_{t, \epsilon}\left[\lambda(t)\left\|\boldsymbol{s}_{\theta}(\alpha_{t}x_{n}+\sigma_{t}\epsilon, y_n, t)+\frac{\epsilon}{\sigma_{t}}\right\|_{2}^{2}\right]\\\nonumber
    &=\frac{1}{2N}\sum_{c=1}^{C}\sum_{n=1}^{N_c}\mathbb{E}_{t, \epsilon}\left[\lambda(t)\left\|\boldsymbol{s}_{\theta}(\alpha_{t}x_{n}^c+\sigma_{t}\epsilon, c, t)+\frac{\epsilon}{\sigma_{t}}\right\|_{2}^{2}\right]\textrm{,}
\end{align}
where $x_n^c$ refers to the $n$-th sample with class label $c$, and $N_c$ represents the number of class label $c$. 

Similarly, by taking the gradient w.r.t. $\boldsymbol{s}_{\theta}(z_t, c, t)$, we derive the optimal class-conditioned diffusion model for each class condition $c$

\begin{equation}\label{optimal_cond}
    \boldsymbol{s}_\theta^*(z_t, c, t)=\frac{\sum_{n=1}^{N_c}\exp\left(-\frac{\left\|\alpha_tx_n^c-z_t\right\|_2^2}{2\sigma_t^2}\right)\frac{\alpha_tx_n^c-z_t}{\sigma_{t}^2}}{\sum_{n'=1}^{N_c}\exp\left(-\frac{\left\|\alpha_tx_{n'}^c-z_t\right\|_2^2}{2\sigma_t^2}\right)}=\sum_{n=1}^{N_c}\mathbb{S}\left(-\frac{\left\|\alpha_tx_n^c-z_t\right\|_{2}^{2}}{2\sigma_{t}^{2}}\right)\cdot\frac{\alpha_tx_n^c-z_t}{\sigma_t^2}\textrm{.}
\end{equation}
\section{Implementation Details of Our Basic Experimental setup}\label{appendix_setup}

% We introduce our basic experimental setup as follows. 

\begin{figure*}
\centering
\subfigure{\includegraphics[width=0.46\textwidth]{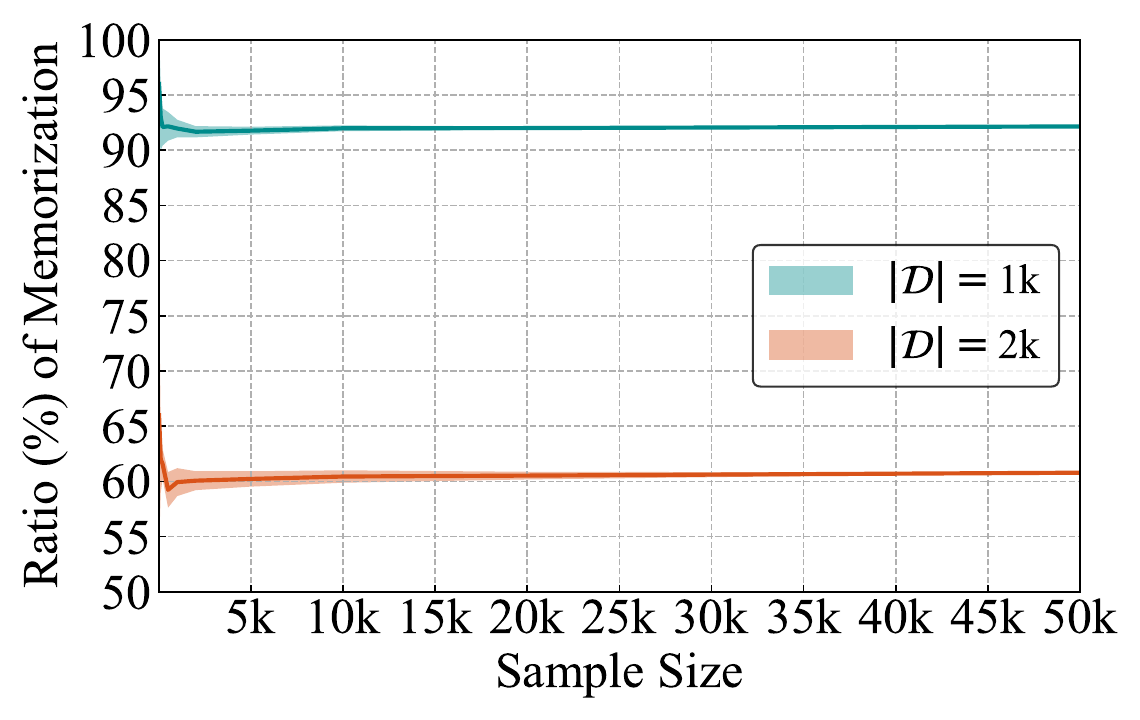}}
\subfigure{\includegraphics[width=0.46\textwidth]{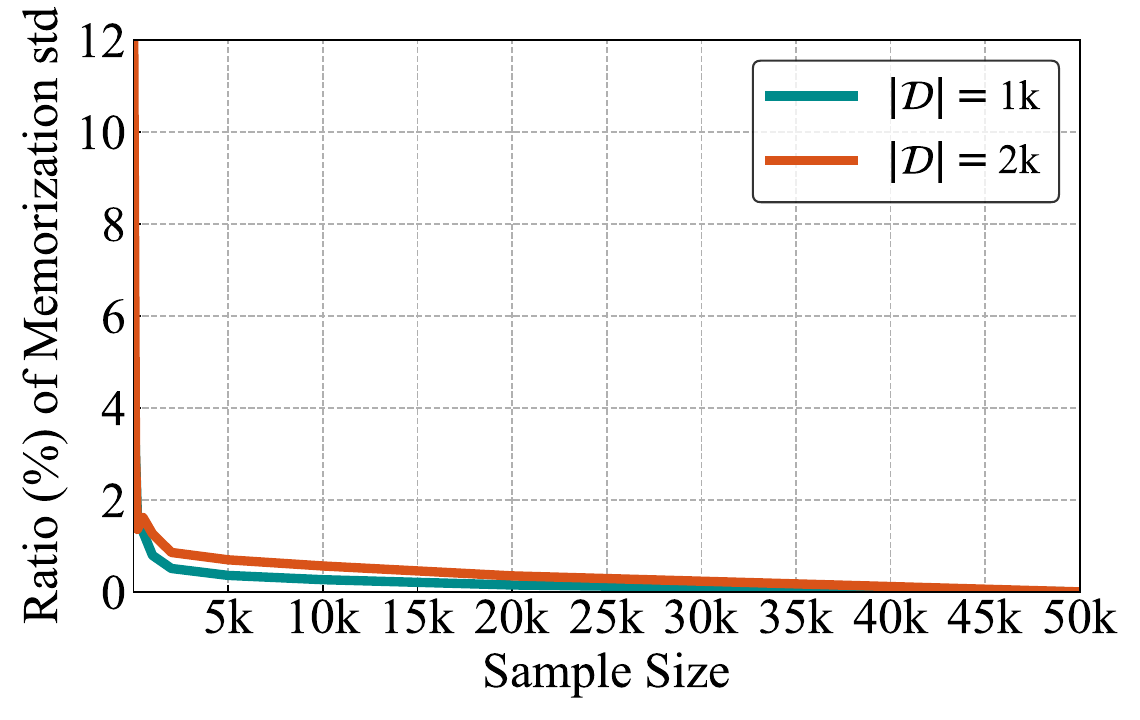}}
% \vspace{-0.2cm}
\caption{(\emph{Left}) Memorization ratios (\%) (\emph{Right}) Standard deviations of memorization ratios (\%) of diffusion models with different sample sizes.}
\label{fig_bootstrap}
% \vspace{-0.2cm}
\end{figure*}

\textbf{Data distribution.} Most of our experiments are conducted on the CIFAR-10 dataset, which consists of 50k RGB images with a spatial resolution of 32$\times$32. CIFAR-10 has 10 classes, each of which has 5k images. We also incorporate the experiments on the FFHQ dataset~\citep{karras2019style} to emphasize our findings. The FFHQ dataset comprises 70k RGB images of human faces with an original resolution of 1024$\times$1024. We down-sample these images to a resolution of 64$\times$64 following \citet{karras2022elucidating} to conduct our experiments. When modifying the intra-diversity of data distribution, we blend several images from the ImageNet dataset~\citep{deng2009imagenet} to construct training datasets. In our study, we disable the data augmentation to prevent any ambiguity regarding memorization. 

\textbf{Model configuration.} We consider the baseline VP configuration in \citet{karras2022elucidating}.\footnote{We run the training configuration C in \citet{karras2022elucidating} after adjusting hyper-parameters and redistributing capacity compared to \citet{song2021score}.} The model architecture is DDPM++, which is based on U-Net~\citep{ronneberger2015u}. As our basic model, we select the number of residual blocks per resolution in the U-Net as 2 instead of 4 in original implementations. For diffusion models trained on CIFAR-10, the channel multiplier is 128, resulting in 256 channels at all resolutions of 32$\times$32, 16$\times$16, and 8$\times$8. While for diffusion models trained on FFHQ, the employed U-Net has 128 channels at the resolution of 64$\times$64, and  256 channels at the resolutions of 32$\times$32, 16$\times$16, and 8$\times$8. The time embedding is positional encoding~\citep{vaswani2017attention}.  

% Consequently, our basic model has 35M trainable parameters, which is close to that in DDPM~\citep{ho2020denoising}.

\textbf{Training procedure.} Our diffusion models are trained using Adam optimizer~\citep{kingma2014adam} with a learning rate of $2\times10^{-4}$ and a batch size of 512 for CIFAR-10 while 256 for FFHQ. The training duration is 40k epochs, while it is 4k epochs in \citet{karras2022elucidating}.  It is worth mentioning that for different training sizes, the number of training epochs is the same. Therefore, for smaller training datasets, the number of total training steps will be smaller. This setup ensures that the frequency of each image being drawn during the training procedure is the same. We schedule the learning rate and EMA rate similar to \citet{karras2022elucidating} but in an epoch-wise manner. In the first 200 epochs, we warm-up the learning rate and the EMA rate with the increase of training iterations. Afterwards, the learning rate is fixed to $2\times10^{-4}$ and the EMA rate is fixed to 0.99929.

\textbf{Generation and evaluation.} We follow the backward process in \citet{karras2022elucidating} to generate images to compute the memorization metric. To decide appropriate sample size, we first train two diffusion models on CIFAR-10 according to our basic experimental setup at $|\mathcal{D}|=\,$1k and $|\mathcal{D}|=\,$2k. Afterwards, we generate 50k images by each model and then bootstrap different number of samples to compute memorization ratios. As present in Figure~\ref{fig_bootstrap}, we find that the ratio of memorization has a negligible variance when sample size more than 10k images. Therefore, we generate 10k images throughout this study. We report the highest memorization ratio during the training process. To ensure the consistency in our analysis, we maintain the same set of 10k samples for computing FIDs and alternative memorization metrics. Specifically, we compute FIDs using 10k generated samples and the training dataset $\mathcal{D}$ (with varying sizes $|\mathcal{D}|$). 

\textbf{Experimental compute resource.} All experiments were run on 8 NVIDIA A100 GPUs, each with 80GB of memory. The running time of each experiment highly depends on the model configuration and data size. For example, it takes about 120 hours to train an unconditional model or conditional model with unique labels in Figure~\ref{fig_model_cond4}. 
 
\textbf{Experimental statistical significance.} We also evaluate the variance of memorization ratios across multiple trials. Specifically, on CIFAR-10, we run a pair of experiments focused on data distribution, primarily by altering the number of classes. We set the training data size as $|\mathcal{D}|=\,$1k and the number of classes as $C=\,$2, 5, and then execute each experiment over three distinct random seeds. Consequently, the memorization ratio for $C=\,$2 stands at 94.59$\pm$0.19\%, whereas for $C=\,$5, it is recorded at 92.32$\pm$0.14\%. It is noticed that the variance of memorization ratios is less than 0.2\%, which is insignificant. Considering this minimal fluctuation and difficulty for repeating all experiments, we run each experiment once in the main paper.

\section{More Empirical Results on CIFAR-10}

\subsection{Model Size}\label{appendix_size}
% Apart from the experiments in Sec.~\ref{model_size}, we also conduct the following experiments to further validate the effects of model size on memorization.

\begin{figure}[t]
\centering
\subfigure[]{\includegraphics[width=0.46\textwidth]{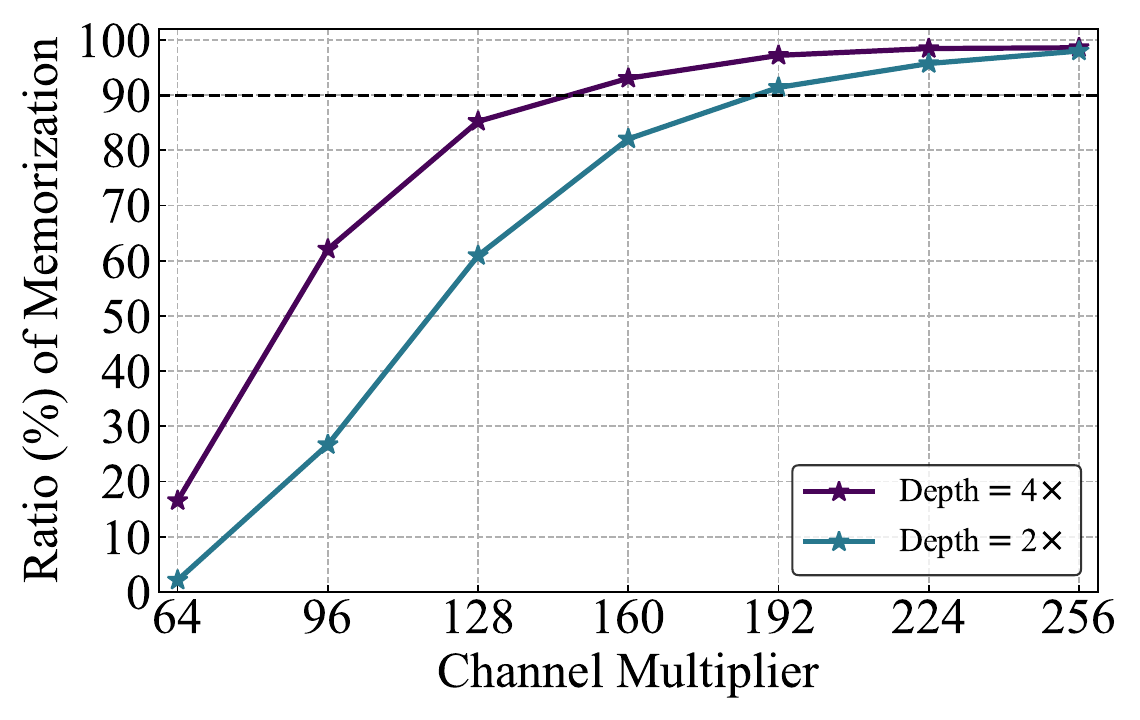}\label{fig_model_width2}}
\subfigure[]{\includegraphics[width=0.46\textwidth]{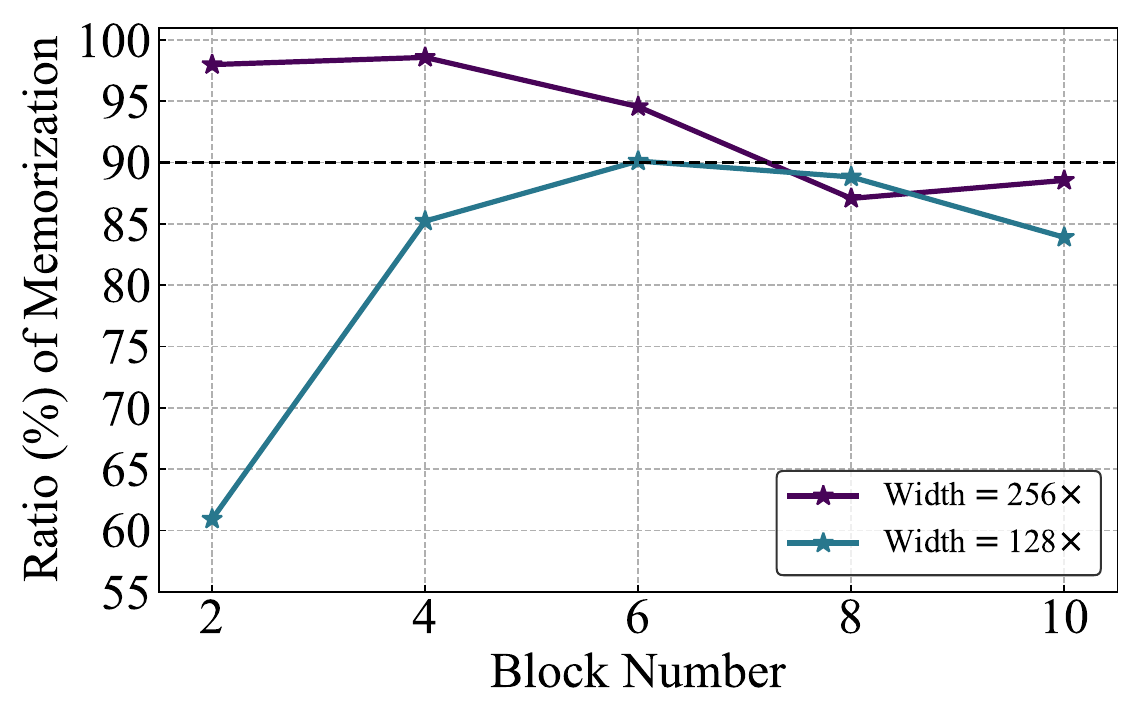}\label{fig_model_depth2}}
\subfigure[]{\includegraphics[width=0.46\textwidth]{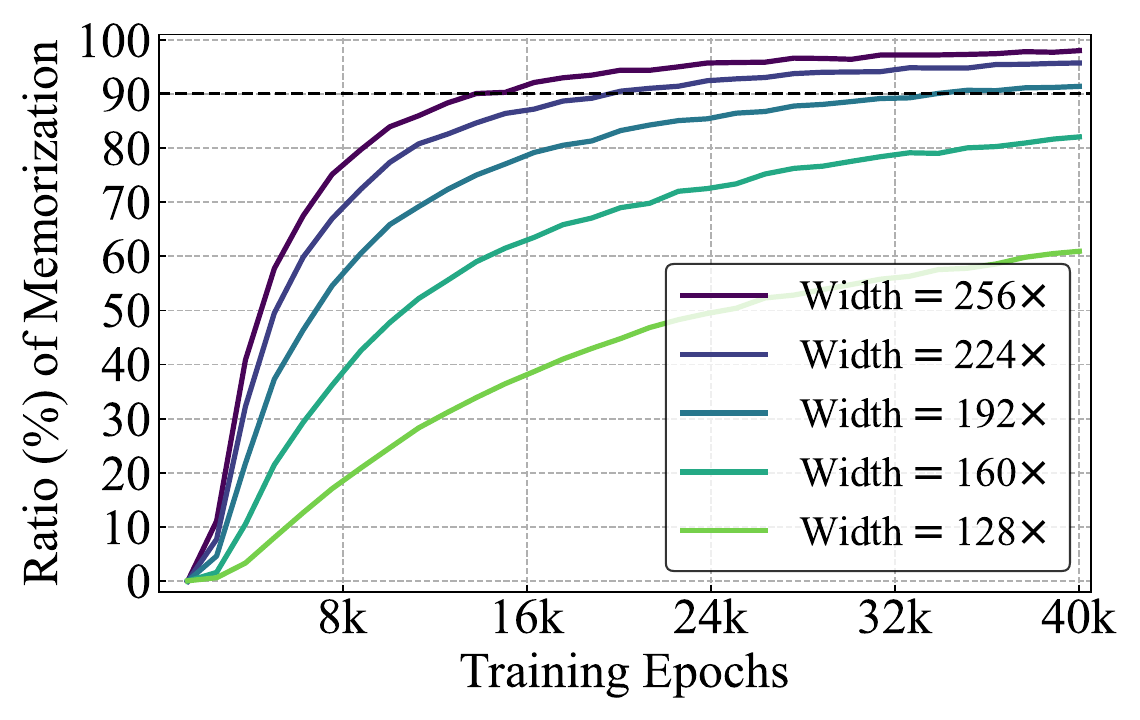}\label{fig_width_process2}}
\subfigure[]{\includegraphics[width=0.46\textwidth]{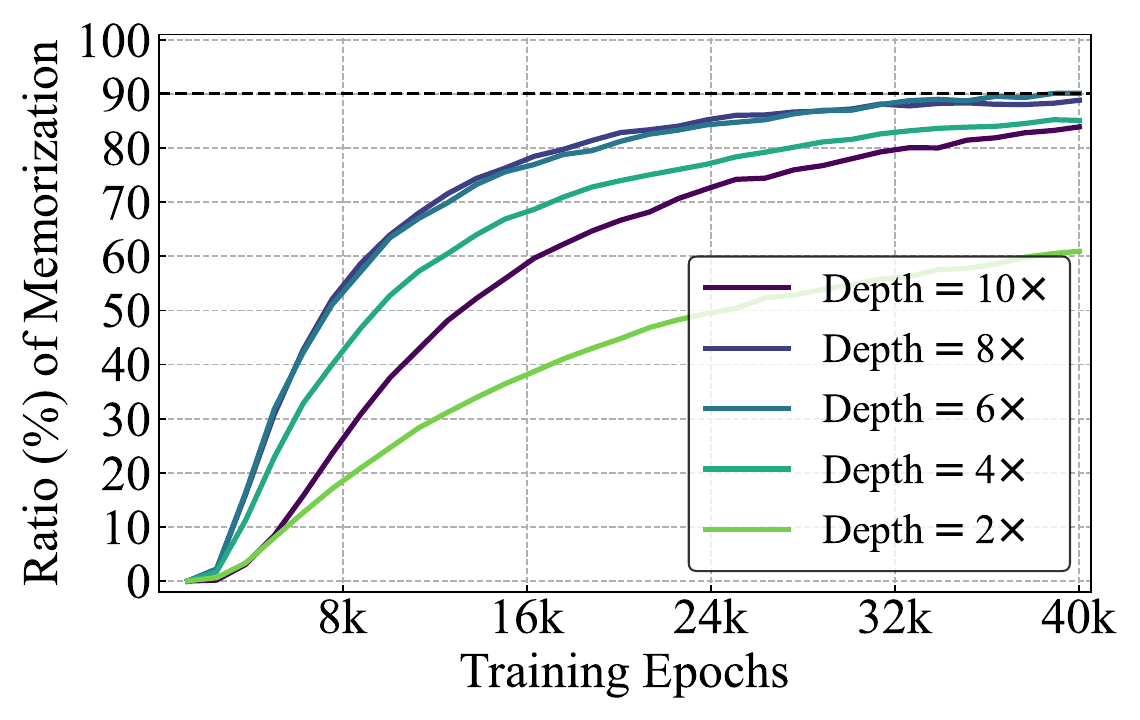}\label{fig_depth_process128}}
\subfigure[]{\includegraphics[width=0.46\textwidth]{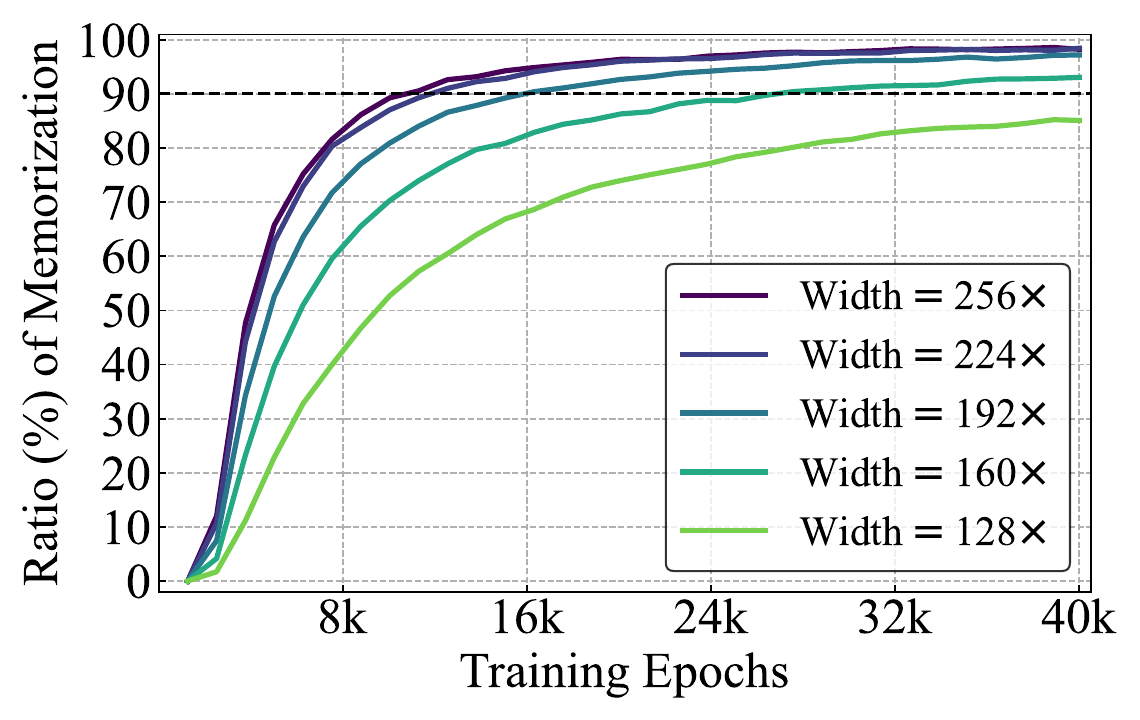}\label{fig_width_process4}}
\subfigure[]{\includegraphics[width=0.46\textwidth]{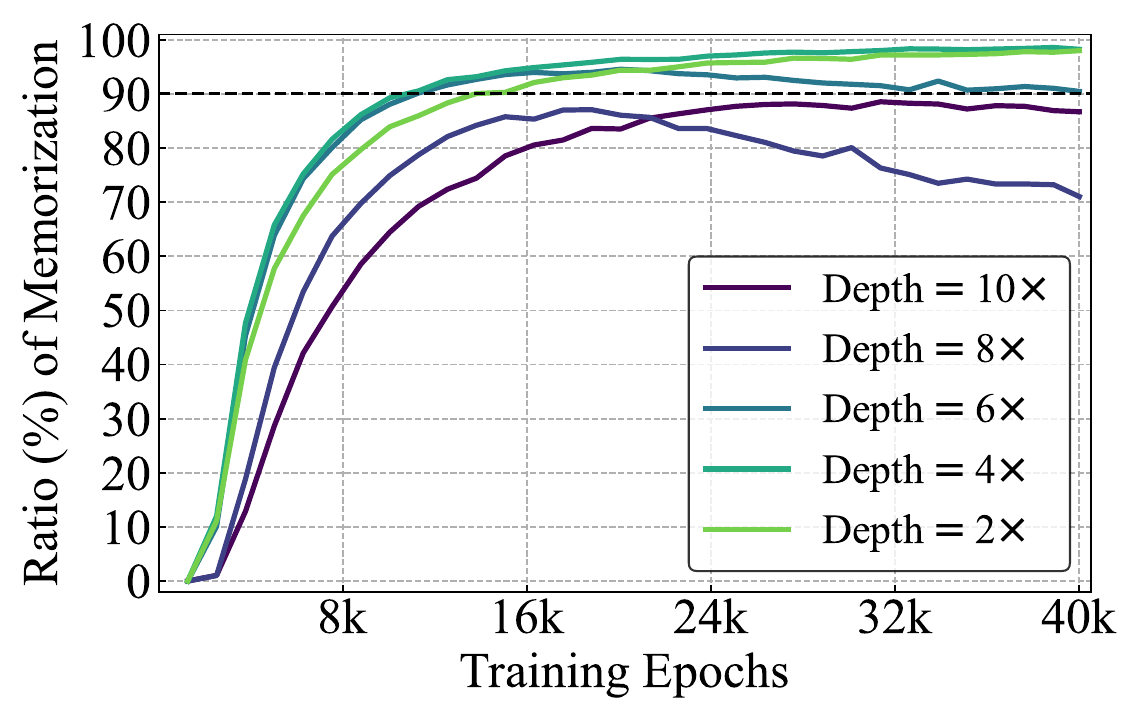}\label{fig_depth_process256}}
\caption{At the size of training data $|\mathcal{D}|=\,$2k, \textbf{memorization ratio (\%)} of diffusion models on CIFAR-10 with (a) varying model widths under 2 or 4 residual blocks per resolution; (b) varying model depths under the channel multiplier as 128 or 256; (c) varying model widths under 2 residual blocks per resolution during the training; (d) varying model depths under the channel multiplier as 128 during the training;  (e) varying model widths under 4 residual blocks per resolution during the training; (f) varying model depths under the channel multiplier as 256 during the training. }
\end{figure}

\textbf{Model width.} We investigate the influence of different channel multipliers, specifically exploring the values from \{64, 96, 128, 160, 192, 224, 256\}. We also consider two scenarios for the number of residual blocks per resolution: 2 and 4.  As illustrated in Figure~\ref{fig_model_width2}, when $|\mathcal{D}|=\,$2k, we observe a consistent and monotonic increase in the memorization ratio as the model width grows. This observation aligns with the conclusions drawn in Sec.~\ref{model_size}. Furthermore, we provide a dynamic view of the memorization ratios during the training process in Figure~\ref{fig_width_process2} and Figure~\ref{fig_width_process4}. It is worth noting that wider diffusion models consistently exhibit higher levels of memorization throughout the entire training process. \looseness=-1

\textbf{Model depth.} We delve into the impact of varying numbers of residual blocks per resolution, considering values in the range of \{2, 4, 6, 8, 10\}. We maintain two different channel multiplier values, specifically 128 and 256, and set the training data size $|\mathcal{D}|=\,$2k. The results, as presented in Figure~\ref{fig_model_depth2}, confirm the non-monotonic effect of model depth on memorization. . When the channel multiplier is set to 256, the curve depicting the relationship between memorization ratio and the number of residual blocks per resolution exhibits multiple peaks. To gain a better understanding of this non-monotonic effect, we visualize the training process in Figure~\ref{fig_depth_process128} and Figure~\ref{fig_depth_process256}. Occasionally, deeper diffusion models yield lower memorization ratios than shallower ones throughout the whole training process. It is noteworthy that when both model width and depth are set large values (e.g., the number of residual blocks per resolution is 8 and the channel multiplier is 256), the memorization of the diffusion model also demonstrates non-monotonic characteristics. 

\subsection{Skip Connections}\label{appendix_skip}

\begin{figure}[t]
\centering
\subfigure[]{\includegraphics[width=0.46\textwidth]{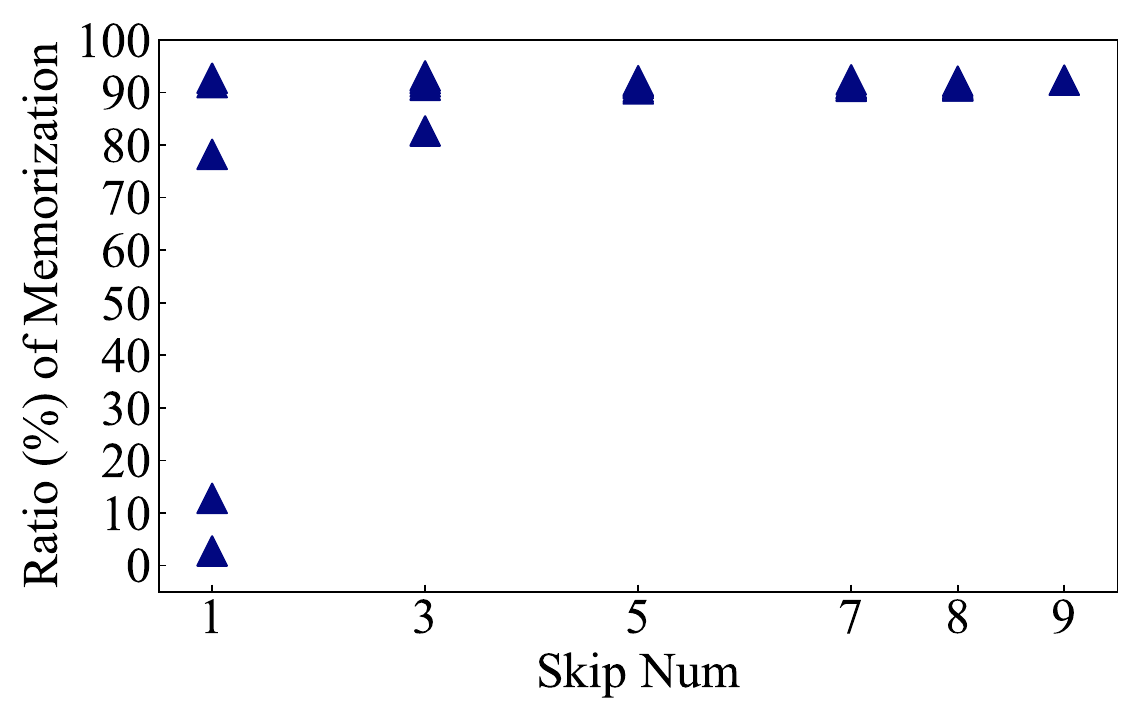}\label{fig_model_skip3_vp}}
\subfigure[]{\includegraphics[width=0.46\textwidth]{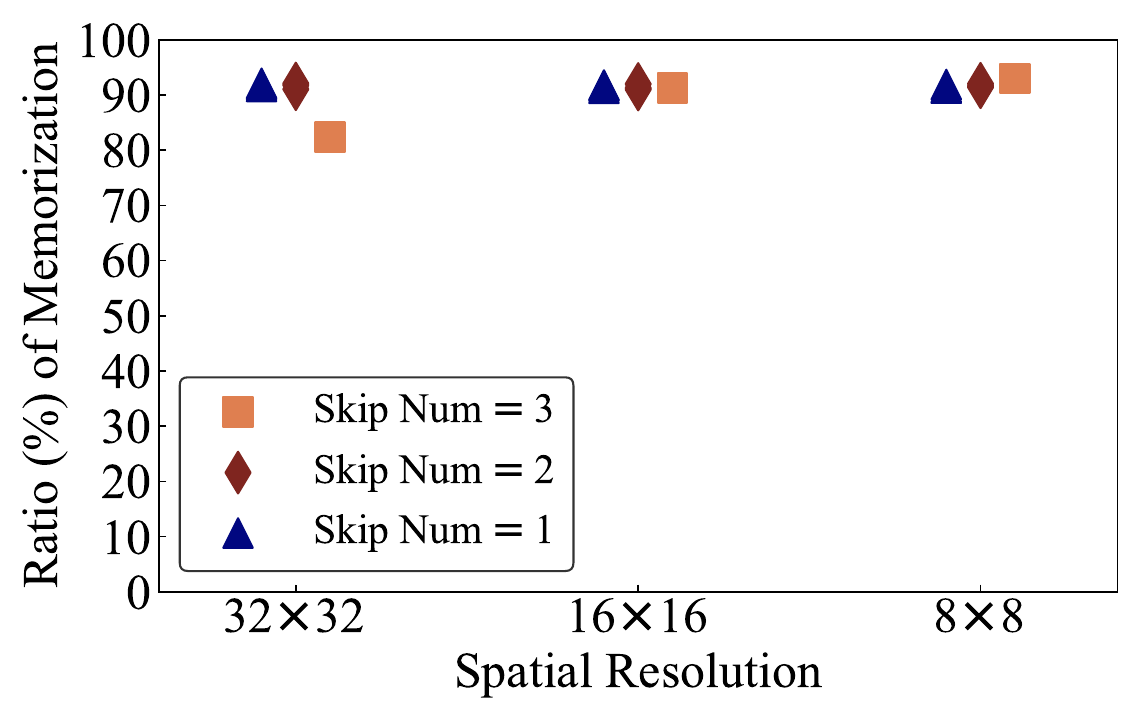}\label{fig_model_skip4_vp}}
\caption{\textbf{Memorization ratio (\%)} of diffusion models on CIFAR-10 when (a) retaining different numbers of skip connections; (b) deleting skip connections of certain spatial resolution.}
\vspace{-0.2cm}
\end{figure}

We employ DDPM++~\citep{song2021score,karras2022elucidating} to explore the influence of skip connection quantity on memorization. Specifically, we consider retaining $m=\,$1, 3, 5, 7, 8, 9 skip connections. To keep the costs tractable, we randomly select five distinct combinations of skip connections for each specific $m$. Our findings, depicted in Figure~\ref{fig_model_skip3_vp}, illustrate a consistent memorization ratio for larger values of $m$, whereas a considerable degree of variance is observed for smaller values of $m$. Notably, when $m=\,$1, the memorization ratio exhibits substantial variability, spanning a range from approximately 0\% to over 90\%.\looseness=-1 

In addition to remaining specific skip connections at different resolutions in the main paper, we also investigate the effect of selectively deleting certain skip connections at varying resolutions. As present in Figure~\ref{fig_model_skip4_vp}, when deleting one or two skip connections at any spatial resolution, the memorization ratios consistently remain above 90\%. It is noteworthy that the memorization ratio only falls below 90\% when three skip connections are deleted at the resolution of 32$\times$32. These findings further confirm the pivotal role played by skip connections at higher resolutions on memorization of diffusion models.\looseness=-1

\subsection{Unconditional v.s.\ Conditional Generation}\label{appendix_cond}

\begin{figure}[htbp]
\centering
\subfigure[]{\includegraphics[width=0.46\textwidth]{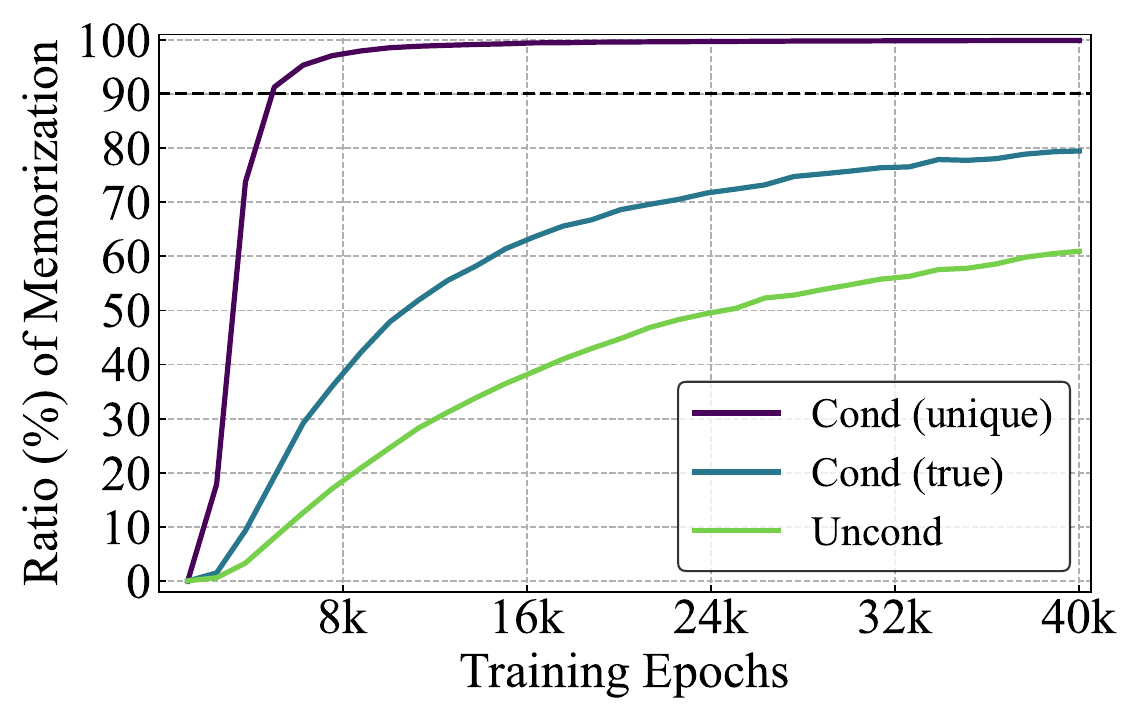}}
\subfigure[]{\includegraphics[width=0.46\textwidth]{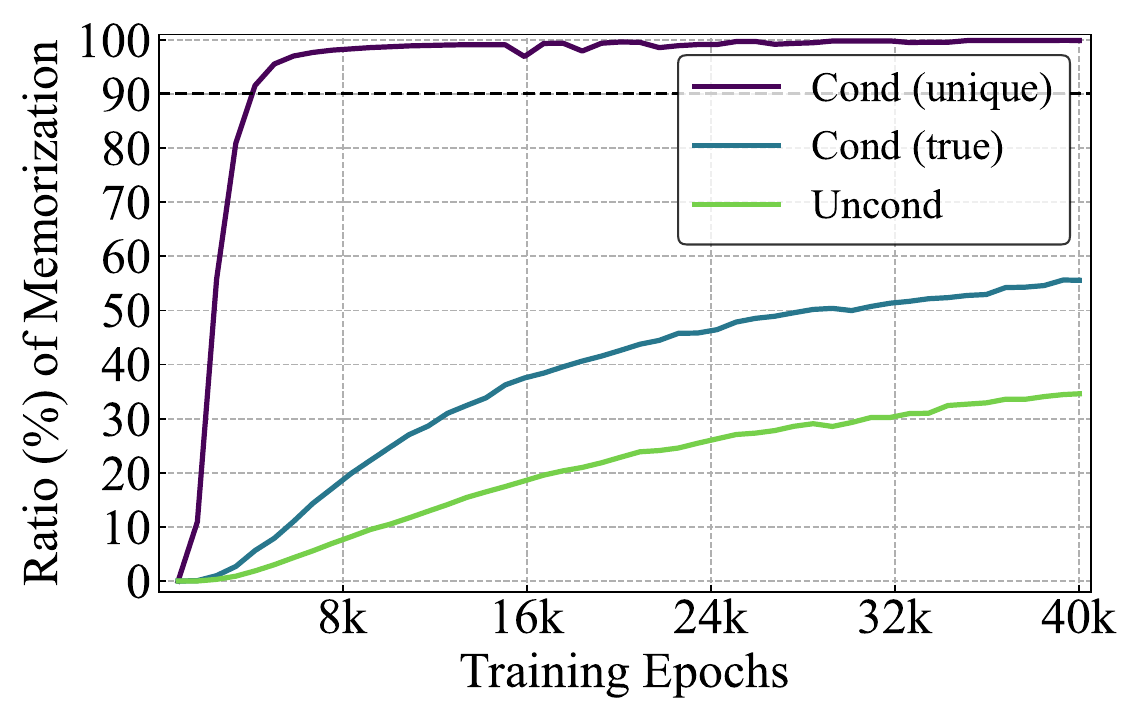}}
\subfigure[]{\includegraphics[width=0.46\textwidth]{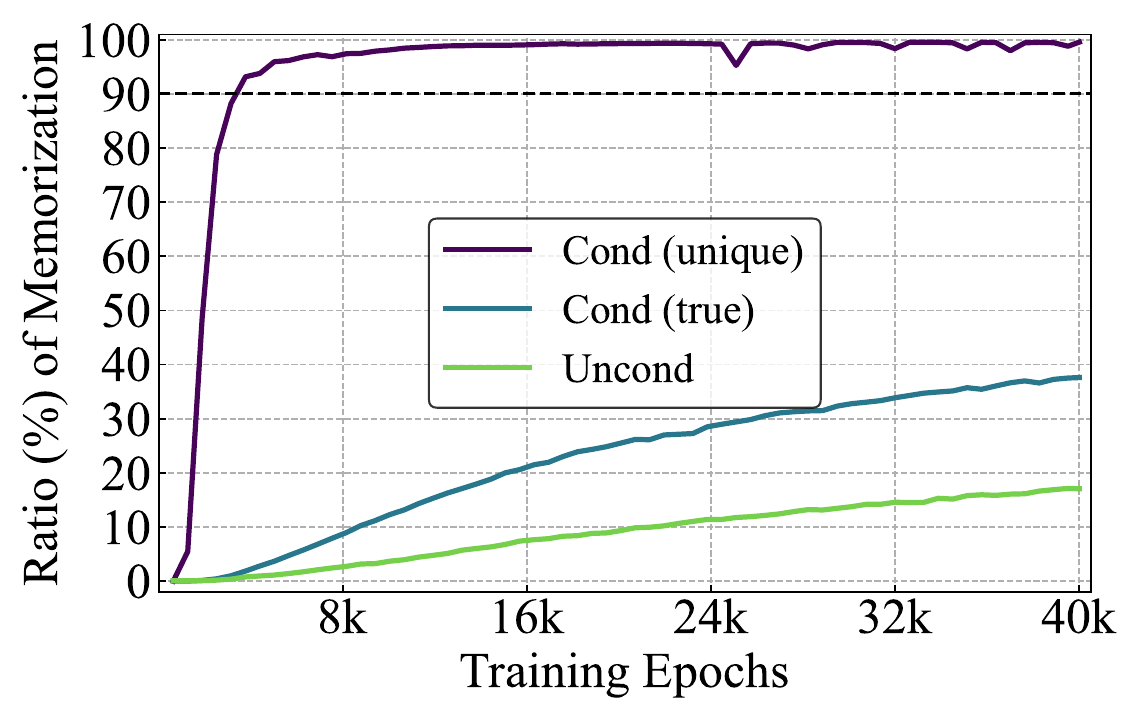}}
\subfigure[]{\includegraphics[width=0.46\textwidth]{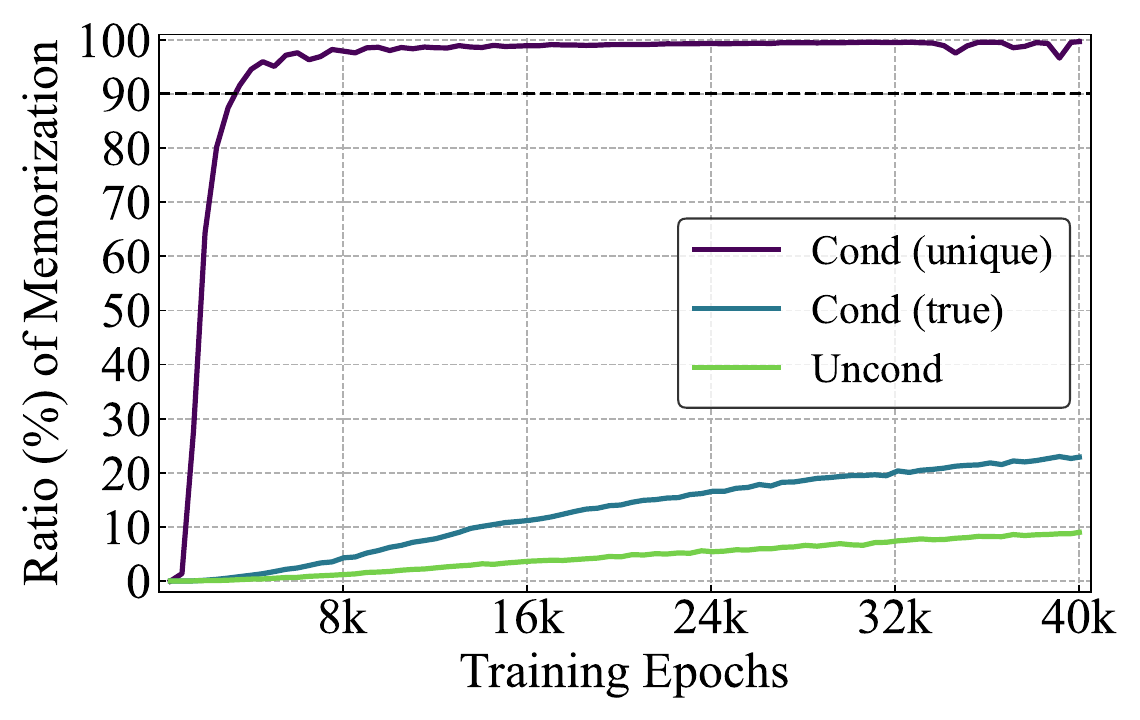}}
\caption{\textbf{Memorization ratios (\%)} of unconditional / conditional (with true labels) / conditional (with unique labels) diffusion models on CIFAR-10 during the training at the training data size (a) $|\mathcal{D}|=\,$2k; (b) $|\mathcal{D}|=\,$3k; (c) $|\mathcal{D}|=\,$4k; (d) $|\mathcal{D}|=\,$5k.\looseness=-1}
\label{fig_appendix_cond}
\end{figure}

We compare the memorization ratios of unconditional diffusion models, conditional diffusion models with true labels, and conditional diffusion models with unique labels. As summarized in Table~\ref{tbl_mem_unique}, it is noticed that with unique labels as class conditions, the trained diffusion models can only replicate training data as the memorization ratios achieve close to 100\% when $|\mathcal{D}|\leq\,$5k. Additionally, we visualize the memorization ratios of diffusion models during the training in Figure~\ref{fig_appendix_cond}, which reaffirms that the diffusion models with unique labels as input conditions memorize training data quickly. Typically, they can achieve more than 90\% memorization ratios within 8k training epochs. 

\begin{figure}[t]
\centering
\subfigure[]{\includegraphics[width=0.46\textwidth]{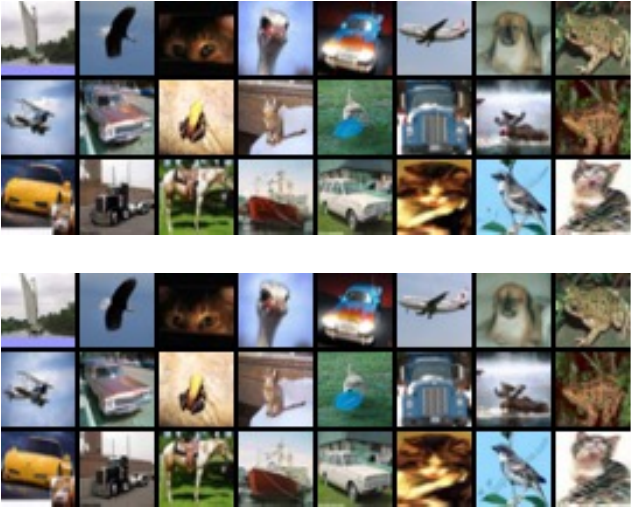}}
\subfigure[]{\includegraphics[width=0.46\textwidth]{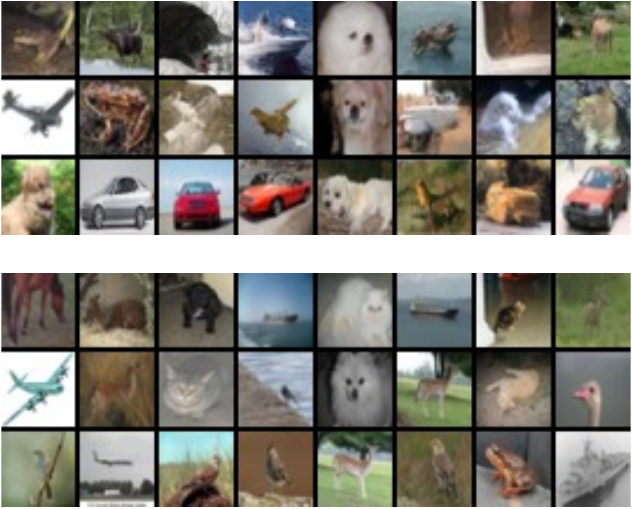}}
\caption{Generated images (top three rows) and their $\ell_2$-nearest training samples in $\mathcal{D}$ (bottom three rows) by (a) the conditional EDM with unique labels (b) the unconditional EDM.}
\label{fig_gen_knn}
\end{figure}

Finally, when trained on the entire CIFAR-10 dataset, i.e., $|\mathcal{D}|=\,$50k, the memorization ratio of conditional EDM with unique labels achieves more than 65\% within 12k training epochs, as present in Figure~\ref{fig_model_cond4}. However, its unconditional counterpart still maintains a zero value of memorization ratio. To further demonstrate this memorization gap, we visualize the generated images and their $\ell_2$-nearest training samples in $\mathcal{D}$ by the above two models in Figure~\ref{fig_gen_knn}. It is noticed that the conditional EDM with unique labels replicate a large proportion of training data while the unconditional EDM generates novel samples. \looseness=-1

% \begin{table}[t]
% \begin{center}
% \label{tbl_mem_unique}
% \caption{\textbf{Memorization ratios (\%)} of unconditional / conditional (true labels) / conditional (unique labels) diffusion models on CIFAR-10.}
% \begin{tabular}{l|ccccc}
% \toprule
%  & $|\mathcal{D}|=\,$1k & $|\mathcal{D}|=\,$2k & $|\mathcal{D}|=\,$3k & $|\mathcal{D}|=\,$4k & $|\mathcal{D}|=\,$5k \\
% \midrule
% % \multirow{2}{*}{$\mathcal{R}_{\text{mem}}$(\%)}  &  $|\mathcal{D}|=$1k & 89.52 & 90.87 & 91.52& 92.09& 91.40& 92.05 & \textbf{92.77}\\
% %  & $|\mathcal{D}|=$2k & 56.67 & 60.36& 60.31& 60.93& 62.61& \textbf{63.33} & 61.95\\

% Unconditional & \,\,92.09 & 60.93 & 34.60 & 17.12 & \,\,9.00\\
% Conditional (true labels) & \,\,96.59 & 79.46 & 55.62 & 37.59&  23.00\\
% Conditional (unique labels) & 100.00 & 99.88 & 99.89 & 99.57 & 99.66\\
% \bottomrule
% \end{tabular}
% \end{center}
% % \vspace{-.3cm}
% \end{table}

\begin{table}[t]
\begin{center}
\caption{\textbf{Memorization ratios (\%)} of unconditional / conditional (true labels) / conditional (unique labels) diffusion models on CIFAR-10.}
\begin{tabular}{l|ccccc}
\toprule
 & $|\mathcal{D}|=\,$1k & $|\mathcal{D}|=\,$2k & $|\mathcal{D}|=\,$3k & $|\mathcal{D}|=\,$4k & $|\mathcal{D}|=\,$5k \\
\midrule
Unconditional & \,\,92.09 & 60.93 & 34.60 & 17.12 & \,\,9.00\\
Conditional (true labels) & \,\,96.59 & 79.46 & 55.62 & 37.59&  23.00\\
Conditional (unique labels) & 100.00 & 99.88 & 99.89 & 99.57 & 99.66\\
\bottomrule
\end{tabular}
\label{tbl_mem_unique}
\end{center}
\end{table}

% \color{red}
% \subsection{Noise Scheduling}
% In terms of training procedure, we also explore the effects of noise scheduling on memorization in diffusion models. 

% \clearpage
\section{More Empirical Results on FFHQ}\label{appendix ffhq}

\subsection{Data Dimension}\label{ffhq dim}
We investigate the influence of data dimension on the memorization of diffusion models, particularly those trained using the FFHQ dataset. Specifically, we evaluate various resolutions: 64$\times$64, 32$\times$32, 16$\times$16, with the latter two resolutions achieved by downsampling. We keep the model configurations and training procedure the same. As shown in Figure~\ref{fig_model_dim_ffhq}, for the 64$\times$64 input resolution, we observe an EMM between 500 and 1k. While for the 32$\times$32 input resolution, the EMM is close to 4k. Furthermore, when input resolution is 16$\times$16, the memorization ratio is still above 90\% even for $|\mathcal{D}|=\,$8k. The EMM is estimated to be between 8k and 10k. These results further indicate the significance of data dimensionality on the memorization within diffusion models.

\subsection{Time Embedding}\label{ffhq embedding}
We conduct experiments to compare two distinct time embedding methods within the model structure of DDPM++~\citep{song2021score, karras2022elucidating}: positional embeeding~\citep{vaswani2017attention} and random fourier features~\citep{tancik2020fourier}. As illustrated in Figure~\ref{fig_model_embed_ffhq}, by varying training data size $|\mathcal{D}|$, there is notable decline in the memorization ratio when employing random fourier features for time embeddings in the DDPM++ model, which reconfirms our conclusions in the main paper.\looseness=-1

\subsection{Unconditional v.s. Conditional Generation}\label{ffhq cond}

\begin{figure}[t]
\centering
\subfigure[]{\includegraphics[width=0.46\textwidth]{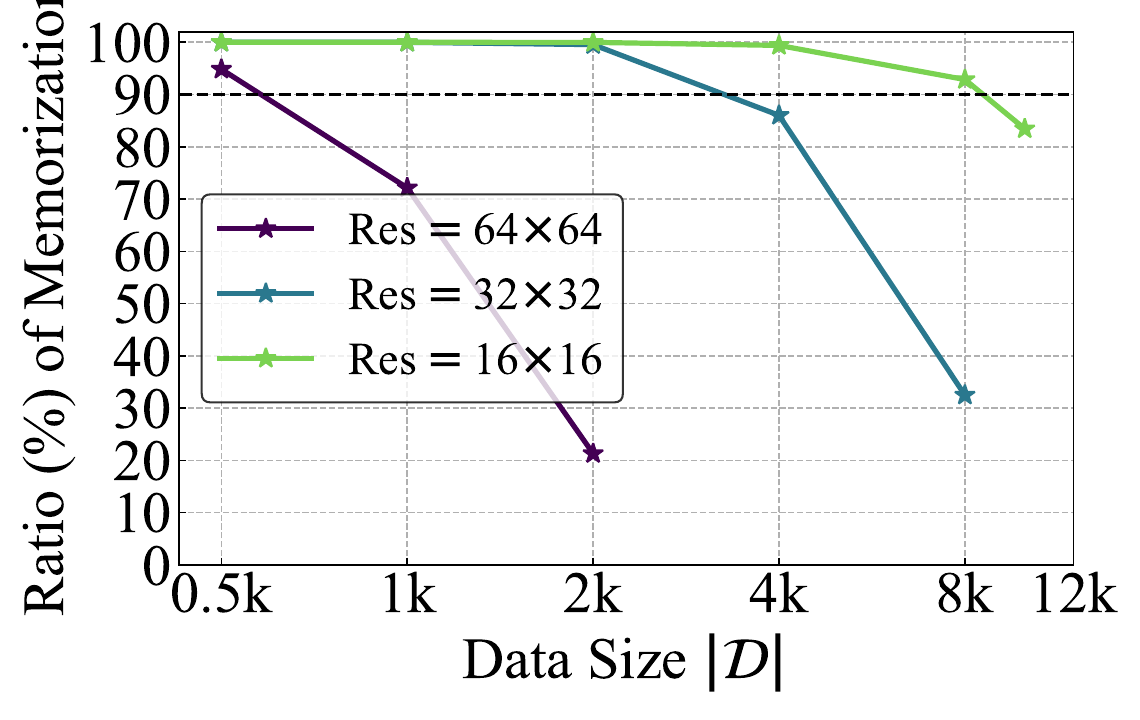}
         \label{fig_model_dim_ffhq}}
\subfigure[]{\includegraphics[width=0.46\textwidth]{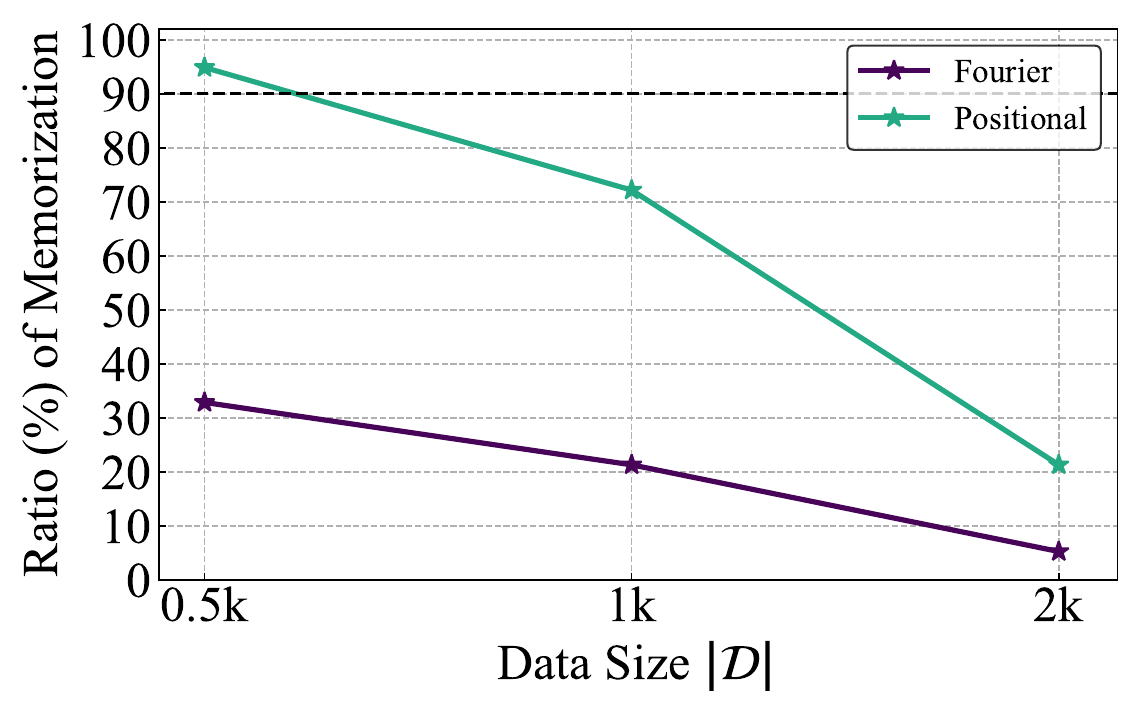}
         \label{fig_model_embed_ffhq}}
\subfigure[]{\includegraphics[width=0.46\textwidth]{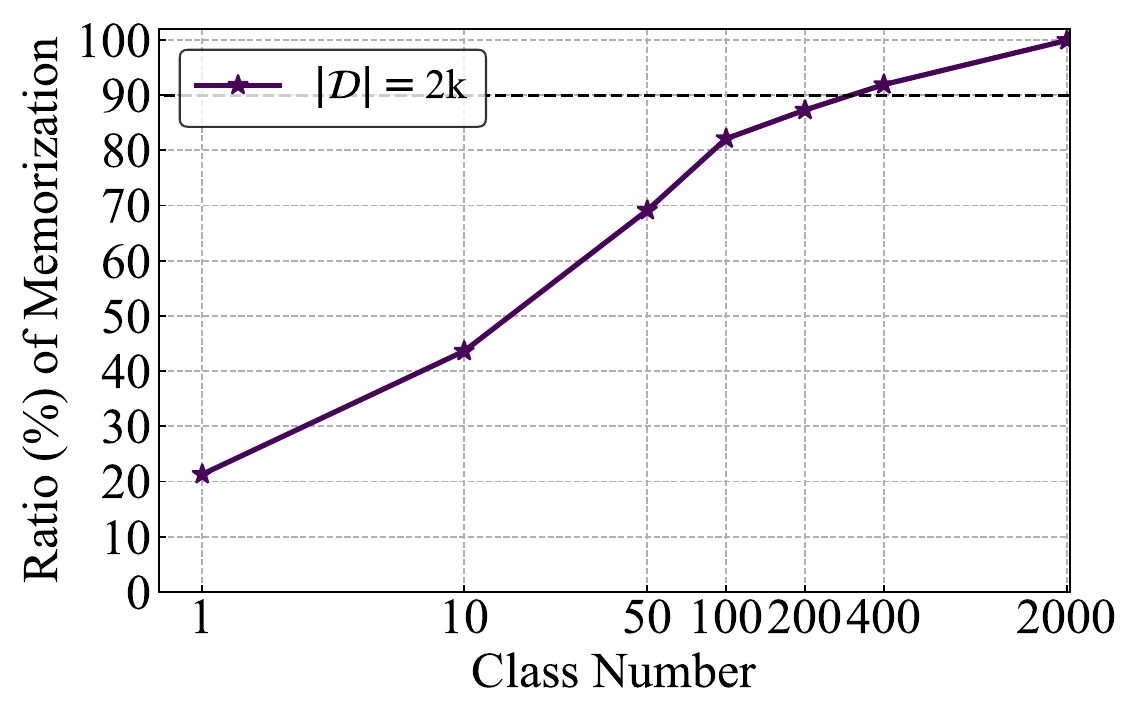}
         \label{fig_model_cond1_ffhq}}
\subfigure[]{\includegraphics[width=0.46\textwidth]{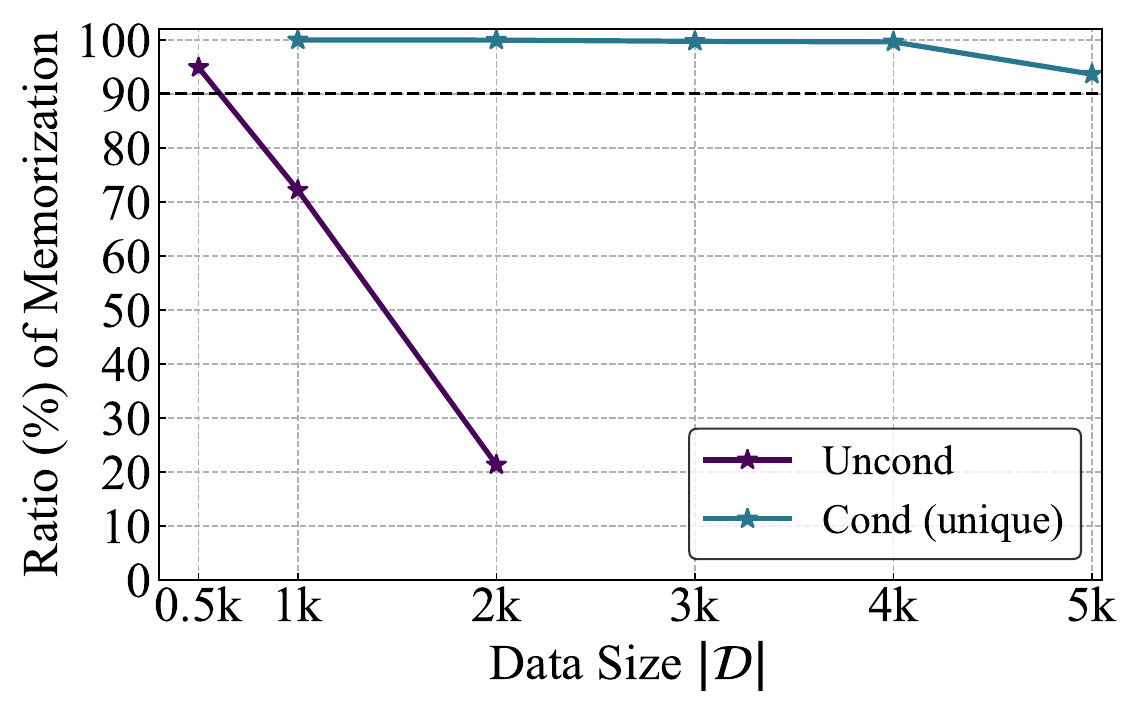}
         \label{fig_model_cond2_ffhq}}
\caption{\textbf{Memorization ratio (\%)} of diffusion models on FFHQ under (a) different data dimension; (b) different time embeddings; (c) different numbers of classes for random labels; (d) unconditional diffusion models and conditional diffusion models with unique conditions.\looseness=-1}
\label{fig_ffhq_cond}
\end{figure}

In the main paper, we demonstrated the significant contributions of random labels to memorization within diffusion models. To further substantiate these findings, we conduct additional experiments using the FFHQ dataset. We note that FFHQ has no ground truth class labels for each image, so we only consider random labels and unique labels in our experimental design. Firstly, we construct a training dataset with $|\mathcal{D}|=\,$2k. Subsequently, the number of classes $C$, for random labels, is varied within the set \{1, 10, 50, 100, 200, 400, 2000\}. Afterwards, we train conditional diffusion models on these training data of different random labels. As demonstrated in Figure~\ref{fig_model_cond1_ffhq}, the memorization ratio for $C=\,$1 (equivalent to the unconditional scenario) approximates to only 20\%. However, it achieves over 90\% with an increase in $C$ to 400. Additionally, we conduct another experiment to investigate the effects of unique labels. As visualized in Figure~\ref{fig_model_cond2_ffhq}, when unique labels are provided as conditions to diffusion models, the memorization ratio still exceeds 90\% at a training data size of $|\mathcal{D}|=\,$5k. In contrast, for unconditional diffusion models, the memorization ratio diminishes even below 90\% when $|\mathcal{D}|=\,$1k. To summarize, these findings show that the number of classes for random labels  plays a pivotal role in the memorization of diffusion models, thus aligning with our earlier results derived from experiments on the CIFAR-10 dataset.

\vspace{-0.1cm}
\section{Experimental Results of FIDs}\label{appendix fid}
\vspace{-0.1cm}

\begin{figure}[t]
\centering
\subfigure[]{\includegraphics[width=0.32\textwidth]{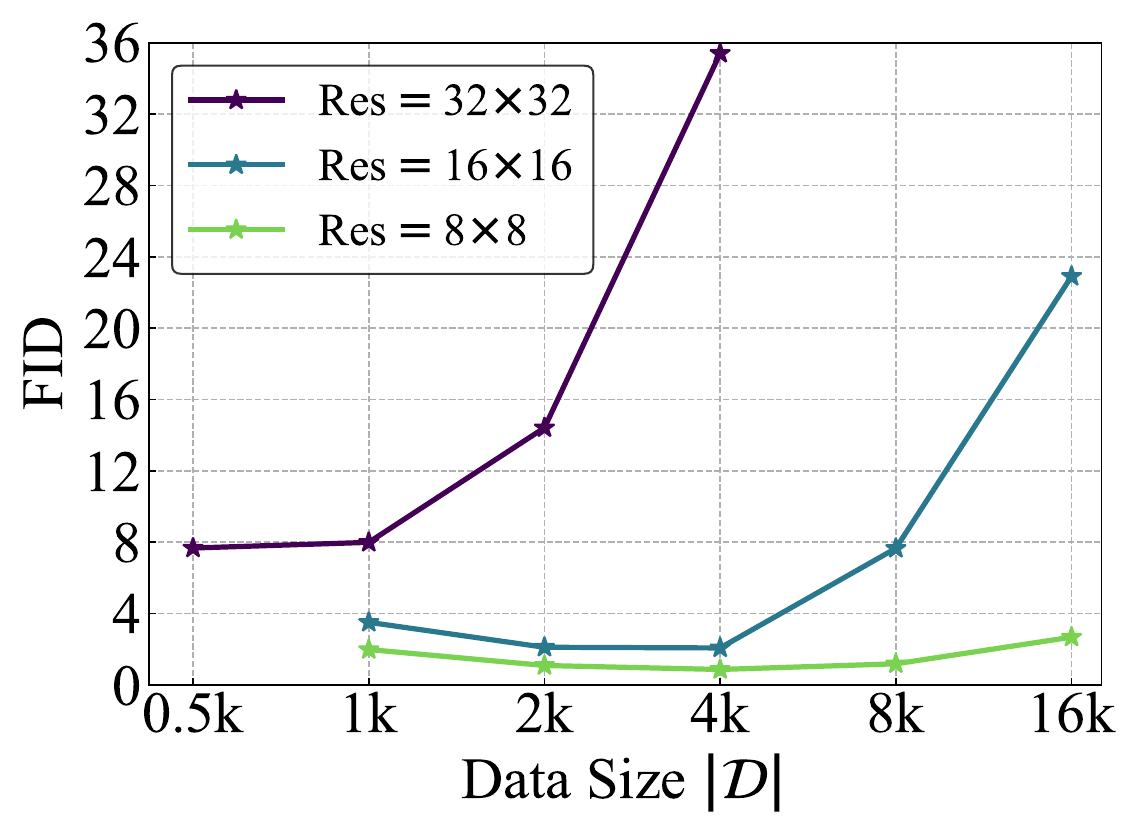}}
\subfigure[]{\includegraphics[width=0.32\textwidth]{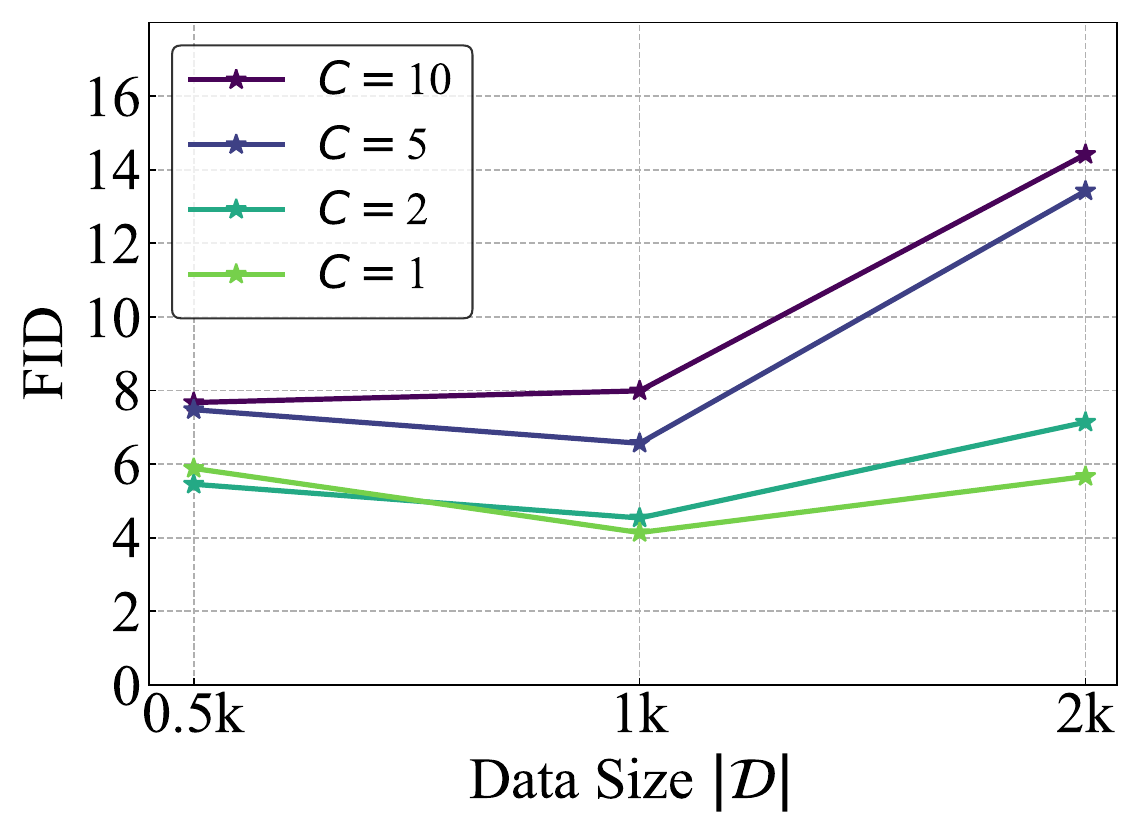}}
\subfigure[]{\includegraphics[width=0.32\textwidth]{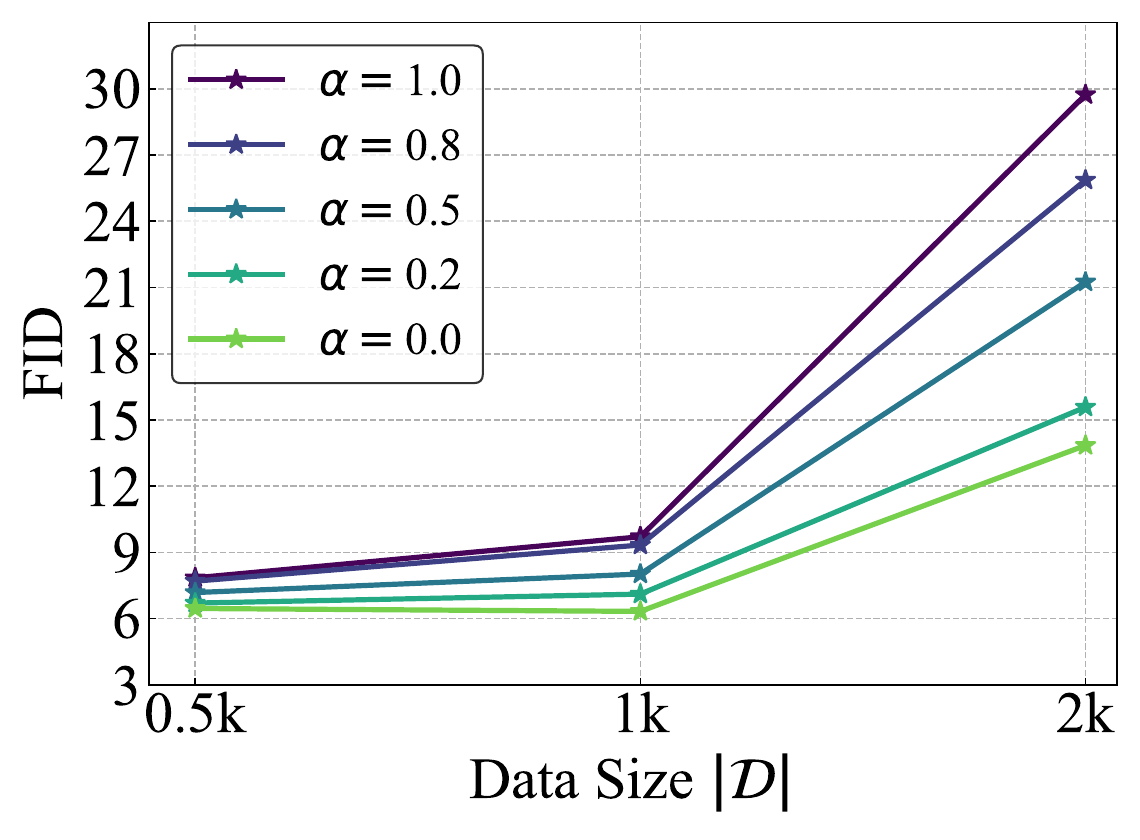}}
\caption{\textbf{FIDs} of diffusion models on CIFAR-10 under different (a) data dimensions; (b) inter-diversity; (c) intra-diversity.}
\label{fid data}
\end{figure}

\begin{figure}[t]
\centering
\subfigure[]{\includegraphics[width=0.32\textwidth]{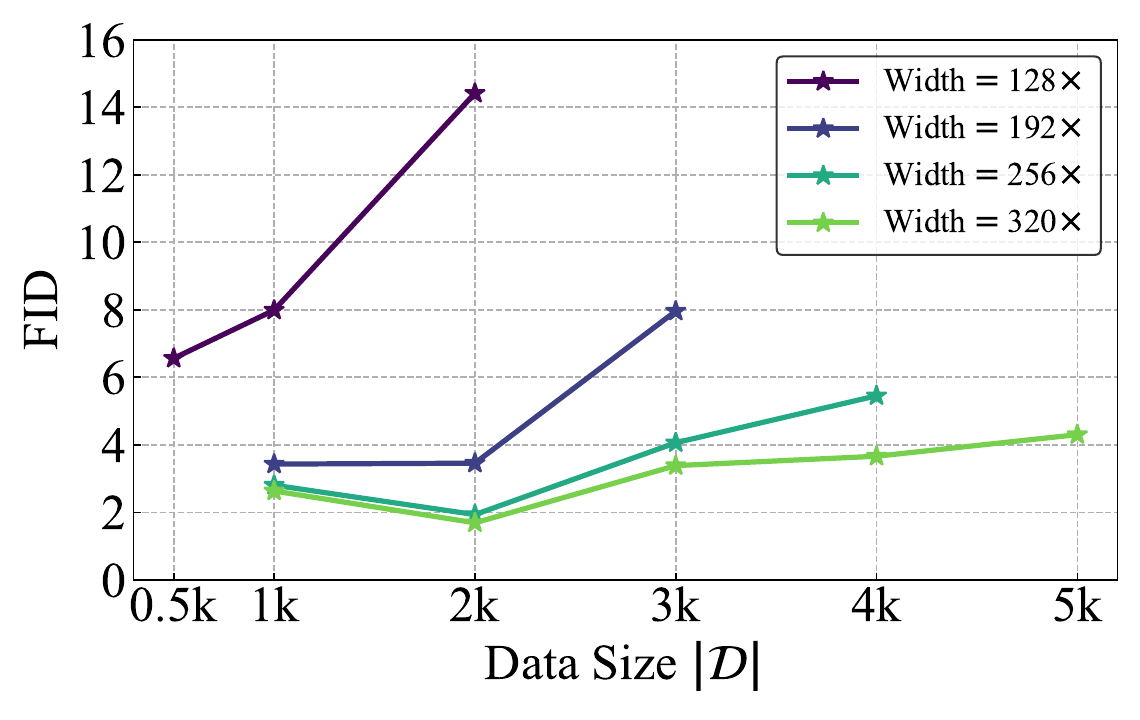}}
\subfigure[]{\includegraphics[width=0.32\textwidth]{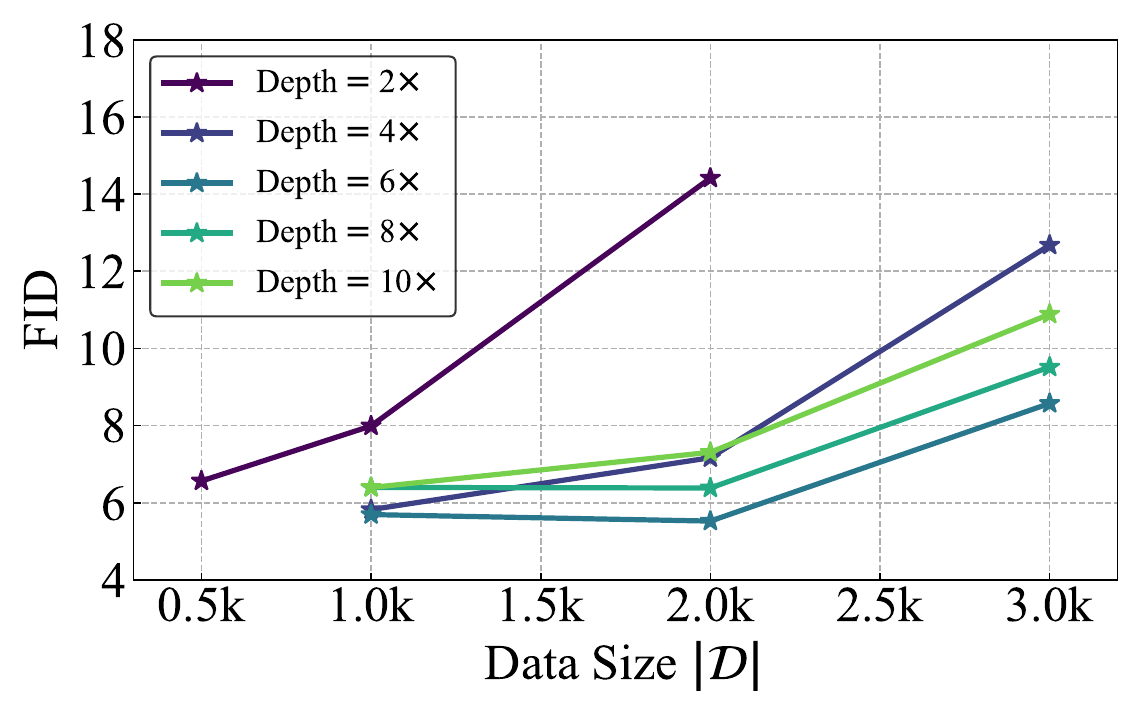}}
\subfigure[]{\includegraphics[width=0.32\textwidth]{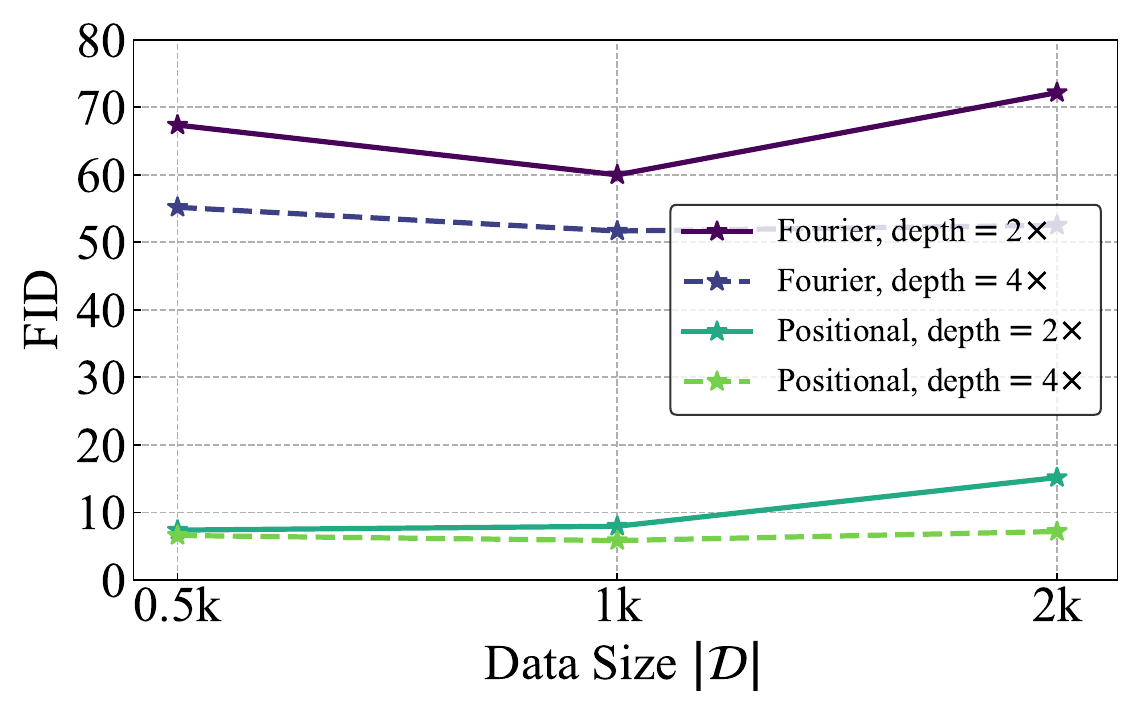}}
\label{fid model}
\caption{\textbf{FIDs} of diffusion models on CIFAR-10 under different (a) model widths; (b) model depths; (c) time embeddings.}
\end{figure}

\begin{figure}[t]
\centering
\subfigure[]{\includegraphics[width=0.46\textwidth]{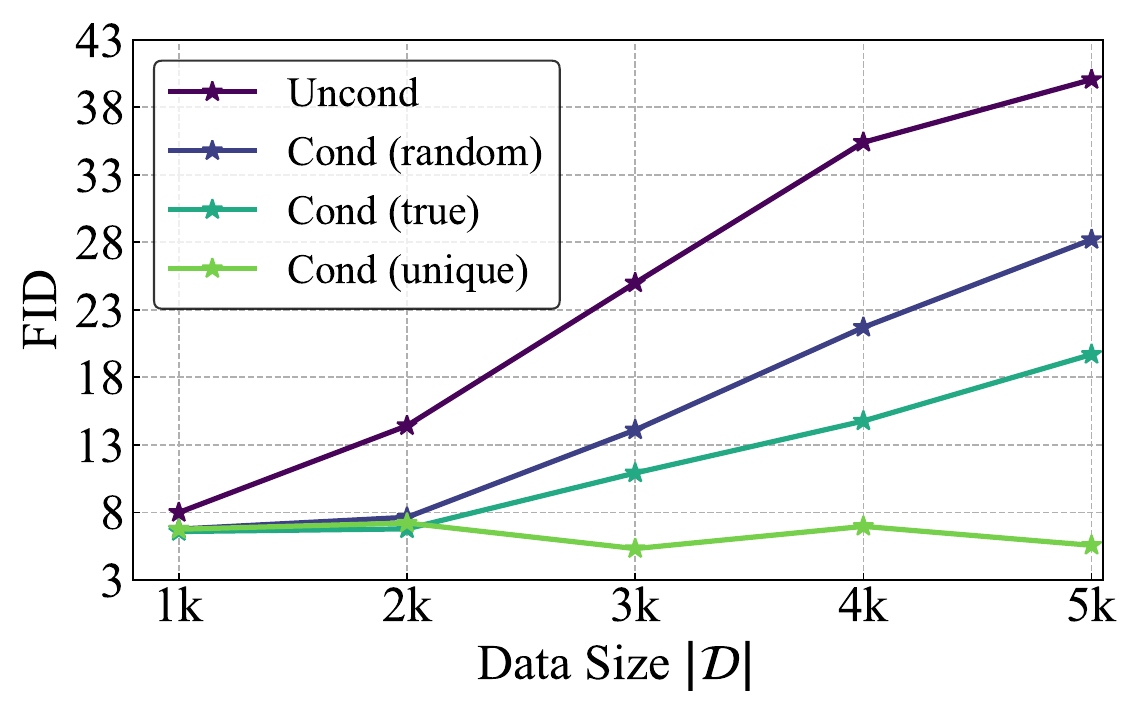}}
\subfigure[]{\includegraphics[width=0.46\textwidth]{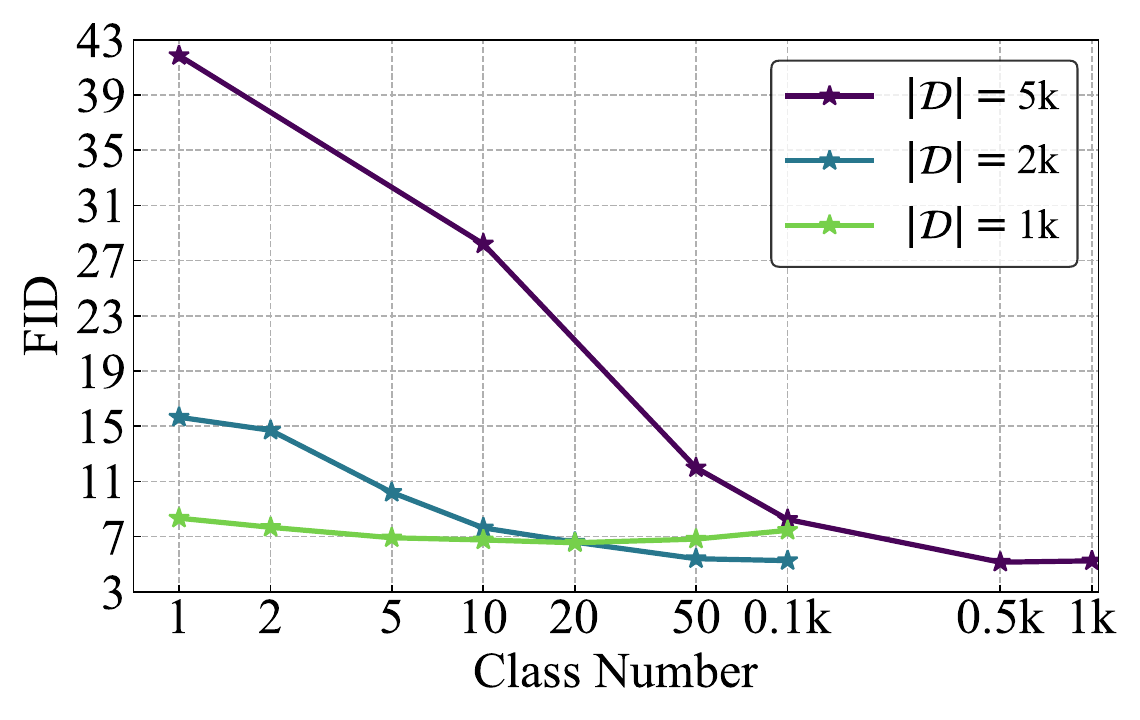}}
% \subfigure[]{\includegraphics[width=0.32\textwidth]{figures/FIDs/Table3.pdf}}
\caption{\textbf{FIDs} of (a) unconditional diffusion models and conditional diffusion models with true / random / unique labels on CIFAR-10; (b) conditional diffusion models with random labels of different $C$.\looseness=-1}
\label{fid cond}
\end{figure}

\begin{figure}[t]
\centering
\subfigure[]{\includegraphics[width=0.46\textwidth]{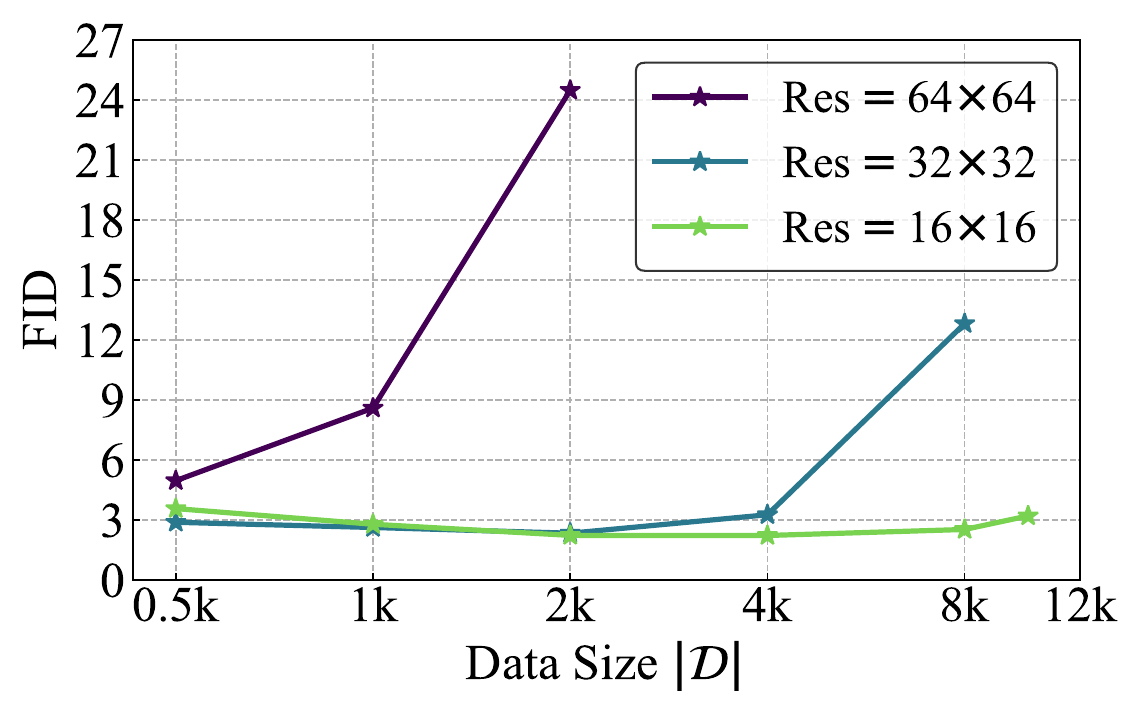}}
\subfigure[]{\includegraphics[width=0.46\textwidth]{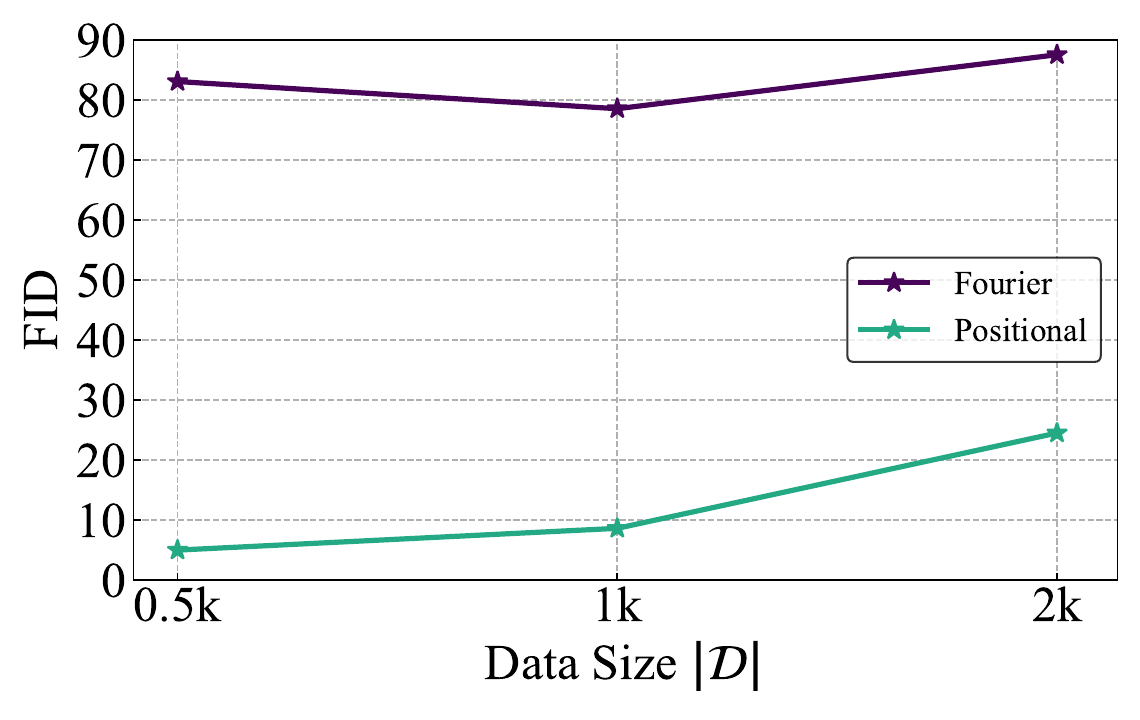}}
\subfigure[]{\includegraphics[width=0.46\textwidth]{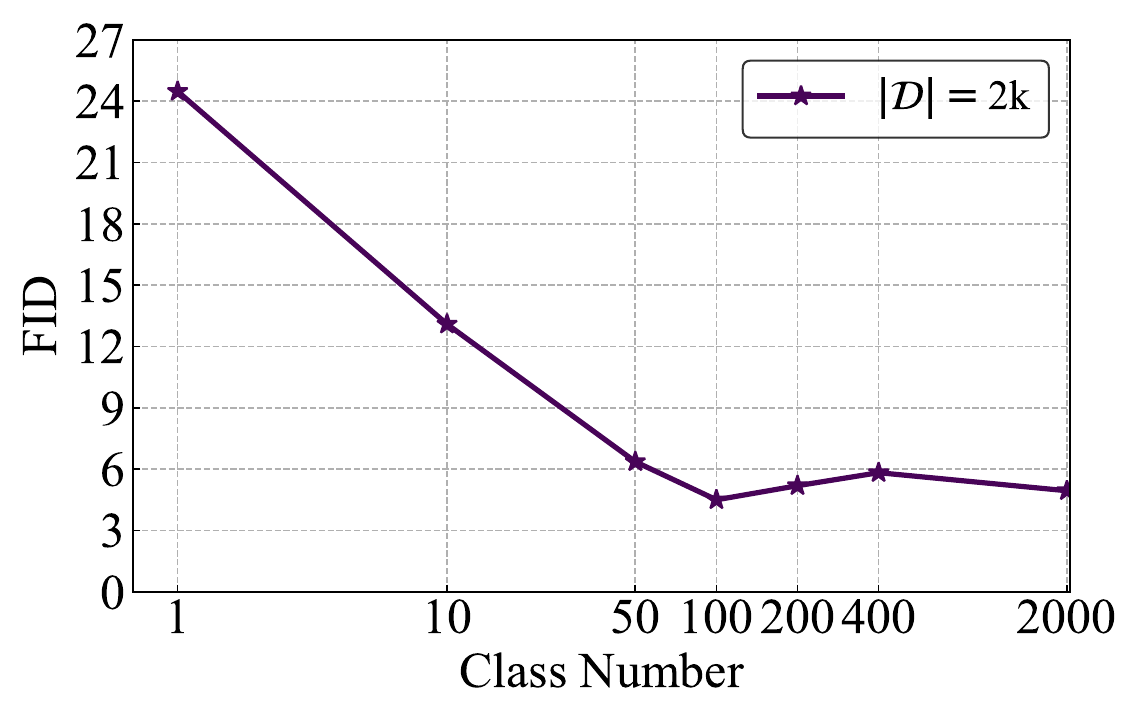}}
\subfigure[]{\includegraphics[width=0.46\textwidth]{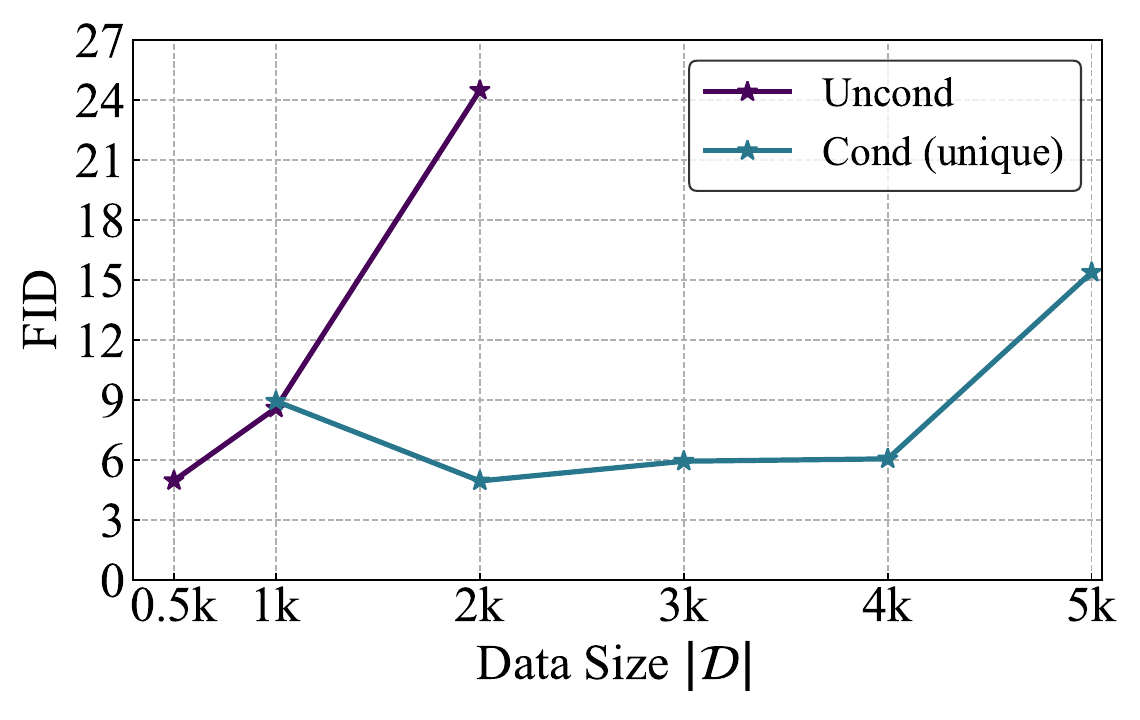}}
\caption{\textbf{FIDs} of diffusion models on FFHQ under (a) different data dimension; (b) different time embeddings; (c) different numbers of classes for random labels; (d) unconditional diffusion models and conditional diffusion models with unique conditions.\looseness=-1}
\label{fid ffhq}
\end{figure}

In this section, we provide the Fréchet Inception Distances (FIDs)~\citep{heusel2017gans} in parallel with the memorization ratios in the main paper. The results related to CIFAR-10 are shown in Figure~(\ref{fid data}-\ref{fid cond}) and Table~(\ref{fid batch}-\ref{fid decay_ema}). While the results on FFHQ are present in Figure~\ref{fid ffhq}. By analyzing the FID results, we observe that a configuration with higher memorization ratio tends to results in a lower FID. This observation is in line with the intuition: when diffusion models memorize a significant proportion of training data, FIDs are also close to optimal.

\begin{figure}[t]
\centering
\subfigure[]{\includegraphics[width=0.245\textwidth]{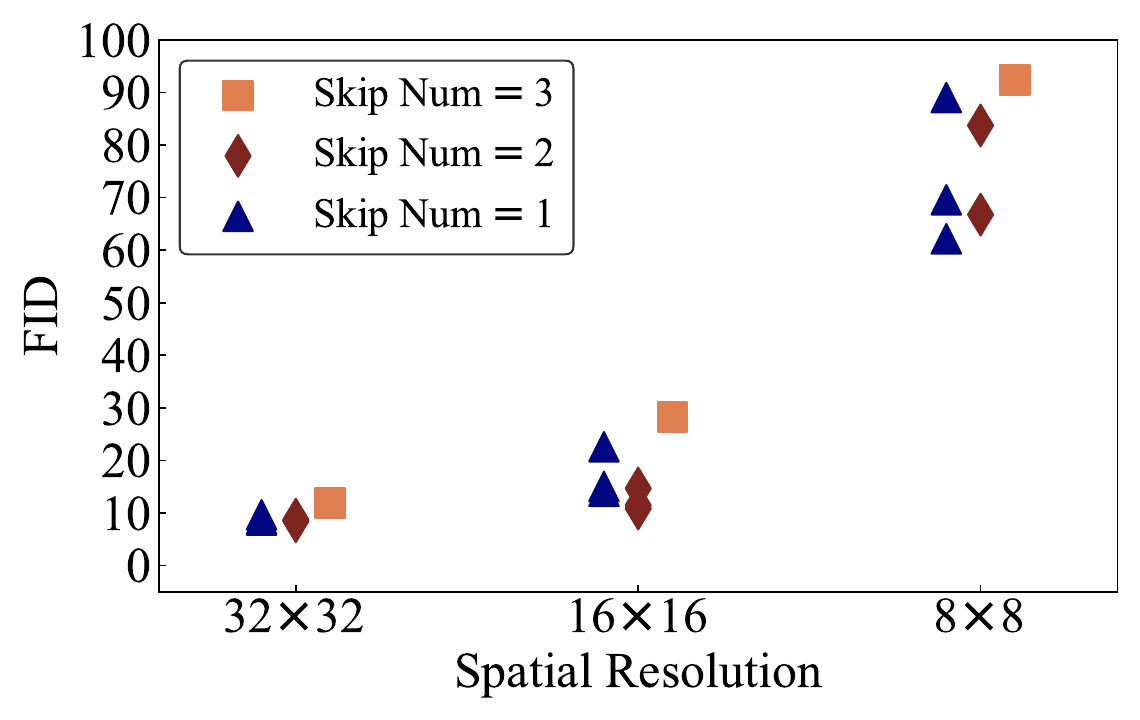}}
\subfigure[]{\includegraphics[width=0.245\textwidth]{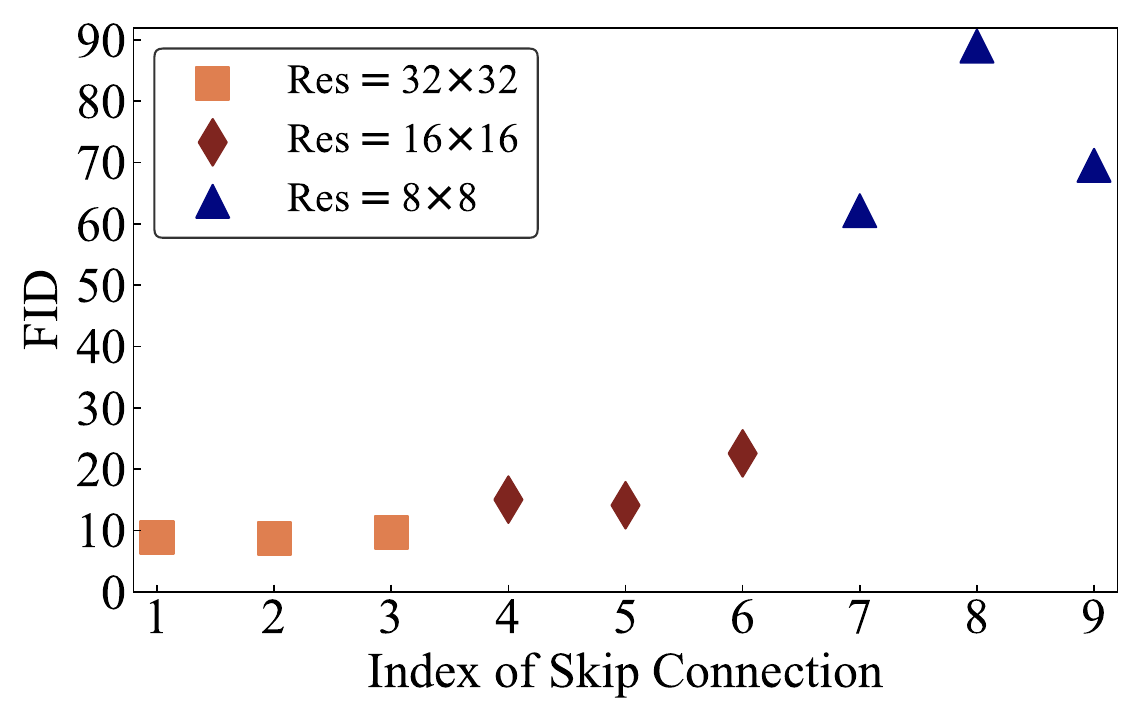}}
\subfigure[]{\includegraphics[width=0.245\textwidth]{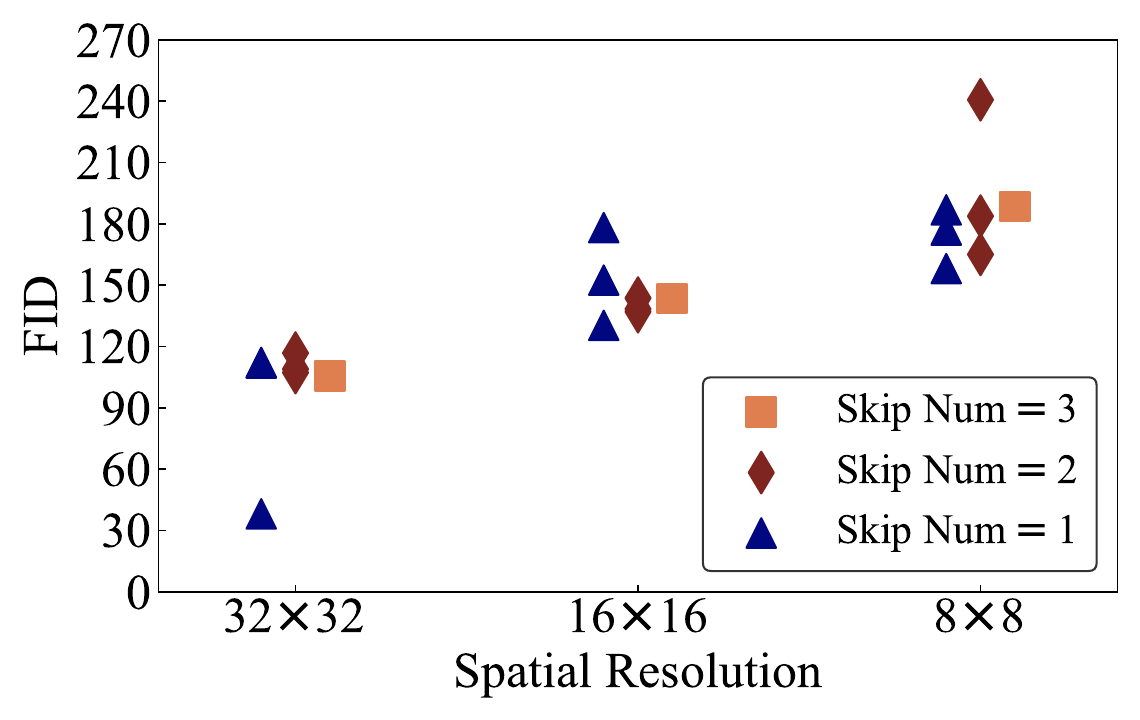}}
\subfigure[]{\includegraphics[width=0.245\textwidth]{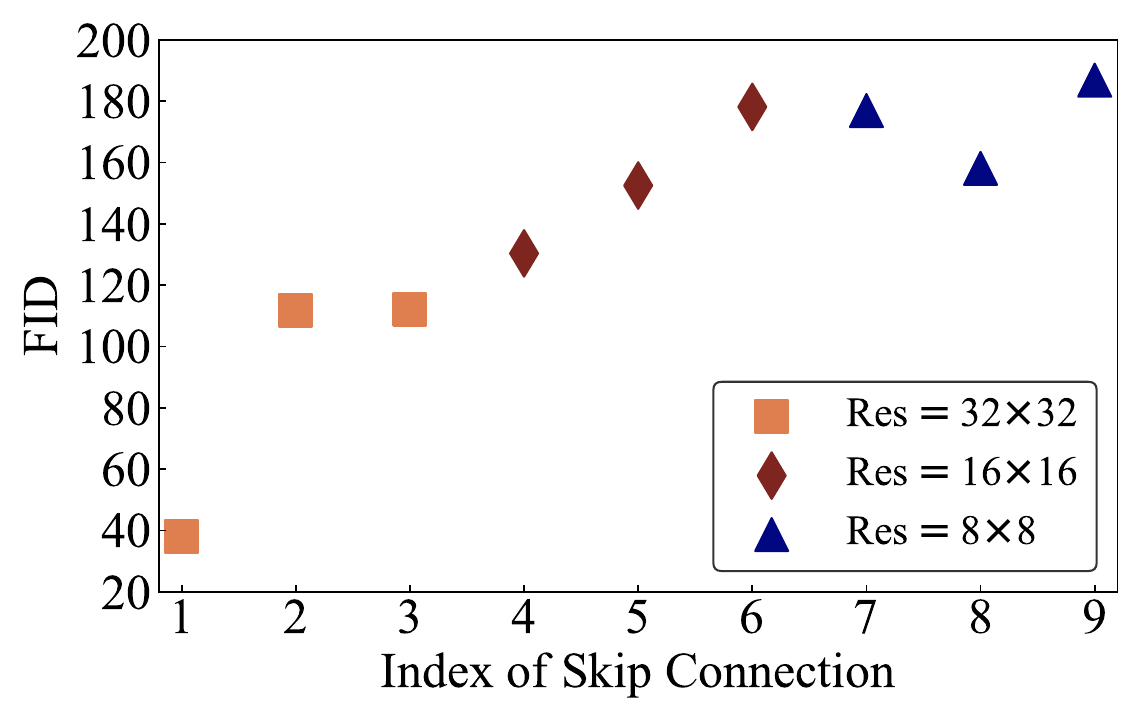}}
\caption{\textbf{FIDs} of diffusion models on CIFAR-10 when retaining (a) skip connections of certain spatial resolution for DDPM++; (b) single skip connection at different locations for DDPM++; (c) skip connections of certain spatial resolution for NCSN++; (d) single skip connection at different locations for NCSN++.}
\end{figure}

\begin{table*}[t]
\begin{center}
\caption{\textbf{FIDs} of diffusion models on CIFAR-10 under different batch sizes.}
\begin{tabular}{l|c|ccccccc}
\toprule
\multicolumn{2}{l|}{Batch Size} & 128 & 256 & 384 & 512 & 640 & 768 & 896 \\
\multicolumn{2}{l|}{Learning rate ($10^{-4}$)} & 0.5  & 1.0  & 1.5 & 2.0 & 2.5 & 3.0 & 3.5 \\
\midrule
\multirow{2}{*}{$\mathcal{R}_{\text{mem}}$(\%)}  &  $|\mathcal{D}|=$1k & 8.89 & 8.60 & 8.14 & 7.99 & 8.25 & 7.98 & \textbf{7.85}\\
 & $|\mathcal{D}|=$2k & 17.56 & 15.57 & 15.23 & \textbf{14.41} & 14.83 & 14.56 & 14.85 \\
\bottomrule
\end{tabular}
\label{fid batch}
\end{center}
\end{table*}
\begin{table}[t]
\caption{\textbf{FIDs} of diffusion models on CIFAR-10 under different (\emph{left}) weight decay values; (\emph{right}) EMA values.}
\begin{minipage}[t]{0.48\linewidth}
\begin{center}
\begin{tabular}{c|cc}
\toprule
\multirow{2}{*}{Weight decay} & \multicolumn{2}{c}{FID} \\
& $|\mathcal{D}|=$1k & $|\mathcal{D}|=$2k\\
\midrule
$0$  & \textbf{7.99} & 14.41\\
$1\times10^{-5}$ & 8.17 & 15.73 \\
$1\times10^{-4}$ & 8.08 & 14.36\\
$1\times10^{-3}$ & 9.37& \textbf{12.75}\\
$1\times10^{-2}$ & 13.47& 17.02\\
$5\times10^{-2}$ & 41.08& 51.70\\
$1\times10^{-1}$ & 63.78 & 48.21\\
\bottomrule
\end{tabular}
\end{center}
% \vspace{-.3cm}
% \caption{Memorization ratios (\%) of different values of weight decay.}
% \vspace{-.3cm}
\end{minipage}
\hfill
\begin{minipage}[t]{0.48\linewidth}
\begin{center}
\begin{tabular}{l|cc}
\toprule
\multirow{2}{*}{EMA} & \multicolumn{2}{c}{FID} \\
& $|\mathcal{D}|=$1k & $|\mathcal{D}|=$2k\\
\midrule
0.99929  & \textbf{7.99} & \textbf{14.41}\\
0.999 & 8.11 & 15.75\\
0.99 & 8.55 & 17.11\\
0.9 & 9.83 & 18.04\\
0.5 & 10.32 & 21.28\\
0.1 & 12.38 & 23.12\\
0.0 & 11.77 & 22.39\\
\bottomrule
\end{tabular}
\end{center}
\end{minipage}
% \vspace{-.2cm}
% \vspace{-.2cm}
\label{fid decay_ema}
\end{table}

\clearpage

% \vspace{-0.2cm}
\section{Experimental Results of Alternative Memorization Criteria}\label{appendix criteria}
% \vspace{-0.2cm}
% \subsection{Validation of the memorization ratio criteria}
% In this section, we validate our choice of memorization ratio criteria. 

% \subsection{Modification of memorization criteria}
In this section, we aim to corroborate our findings regarding memorization of diffusion models by incorporating alternative memorization metrics. The memorization ratio, as defined in the main paper, can be formally articulated as follows

\begin{equation}
    \mathcal{R}_{\textrm{Mem}}=\frac{1}{N'}\sum_{n=1}^{N'}\mathbb{I}(\frac{\left\|x_n'\textrm{,}\,\,\textrm{NN}_1(x_n', \mathcal{D})\right\|_2}{\left\|x_n'\textrm{,}\,\,\textrm{NN}_2(x_n', \mathcal{D})\right\|_2}<\frac{1}{3})\mathrm{,}
\end{equation}
where $\mathbb{I}$ is an indicator function, $\textrm{NN}_j(x_n', \mathcal{D})$ is the $j$-th nearest training sample of $x_n'$ in $\mathcal{D}$, and $N'$ is the number of generations sampled by the diffusion model.

We also consider the KNN distance used in \citet{carlini2023extracting} as a surrogate memorization metric:

\begin{equation}
    \mathcal{R}'_{\textrm{Mem}}=\frac{1}{N'}\sum_{n=1}^{N'}\left\|x_n'\textrm{,}\,\,\textrm{NN}_1(x_n', \mathcal{D})\right\|_2\mathrm{.}
\end{equation}

\begin{figure}[t]
\centering
\subfigure[]{\includegraphics[width=0.32\textwidth]{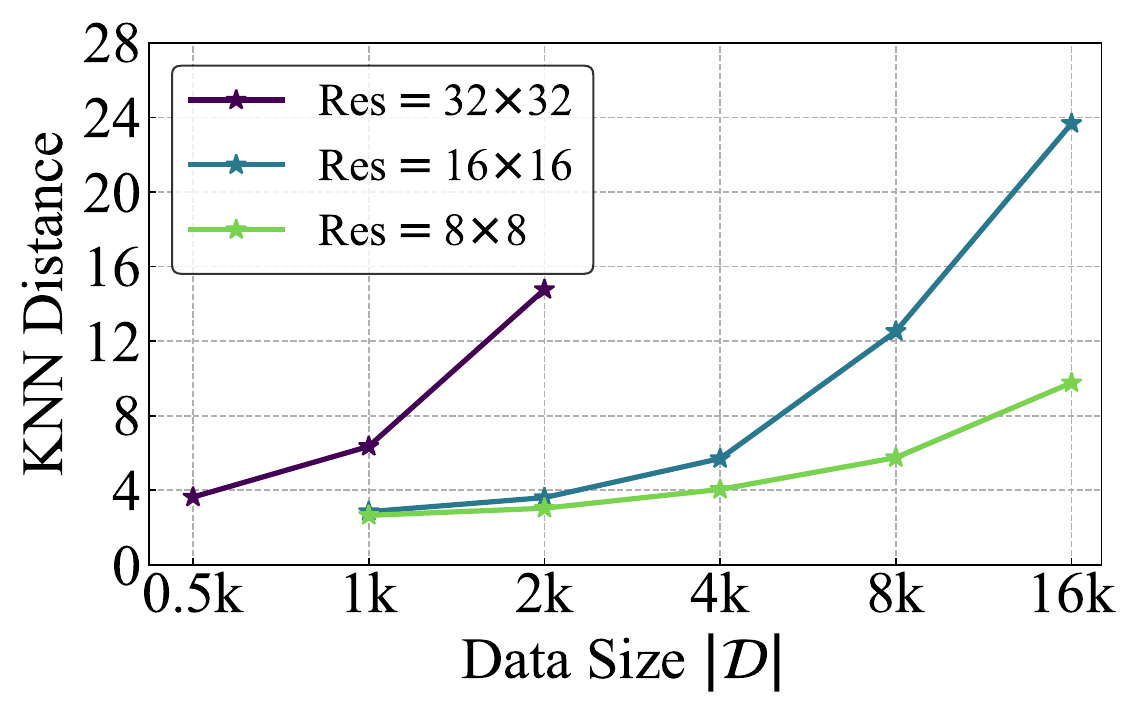}}
\subfigure[]{\includegraphics[width=0.32\textwidth]{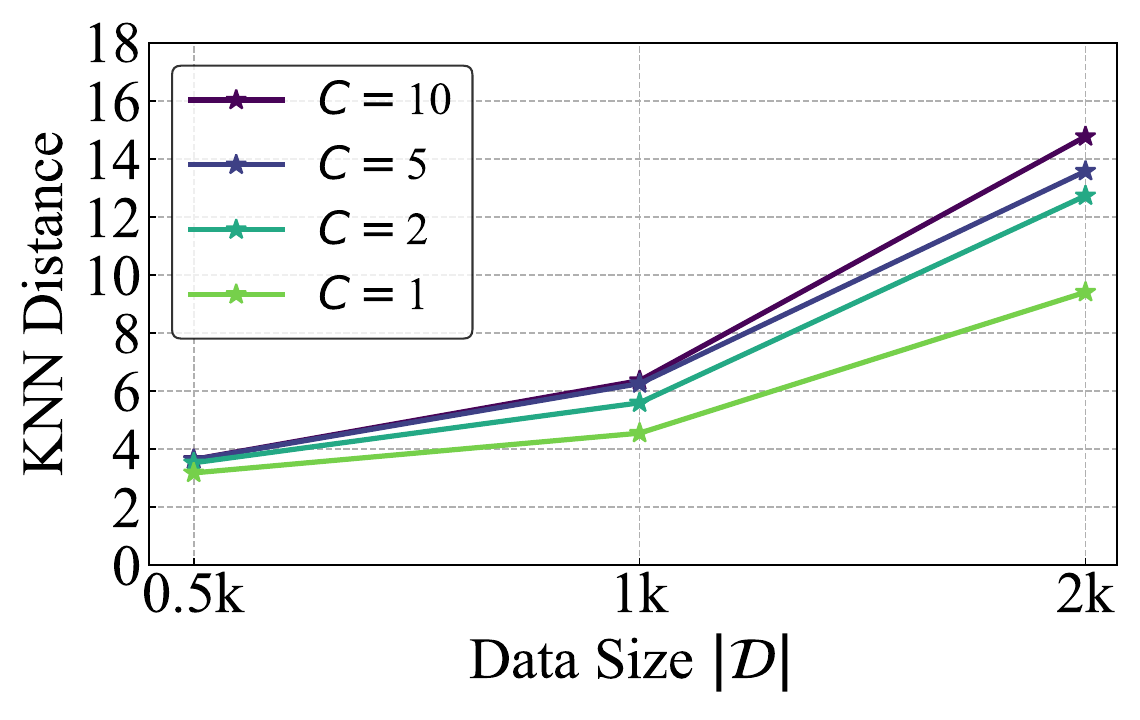}}
\subfigure[]{\includegraphics[width=0.32\textwidth]{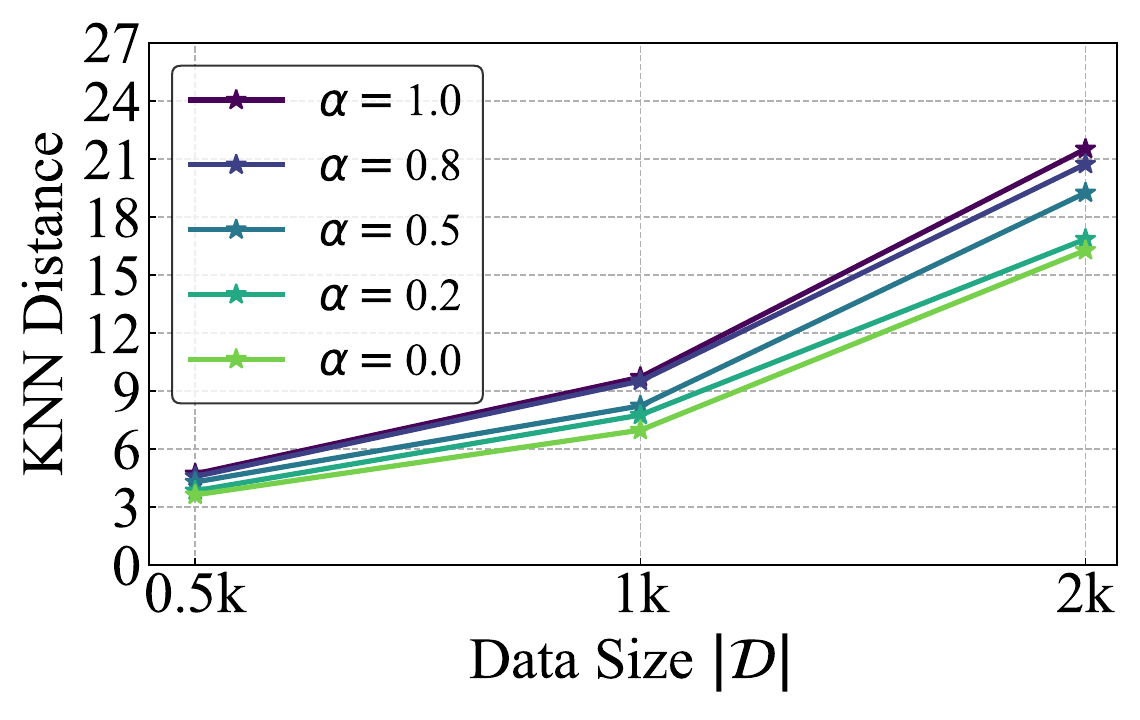}}
\caption{\textbf{KNN distances} of diffusion models on CIFAR-10 under different (a) data dimensions; (b) inter-diversity; (c) intra-diversity.}
\label{knn data}
\end{figure}

\begin{figure}[t]
\centering
\subfigure[]{\includegraphics[width=0.32\textwidth]{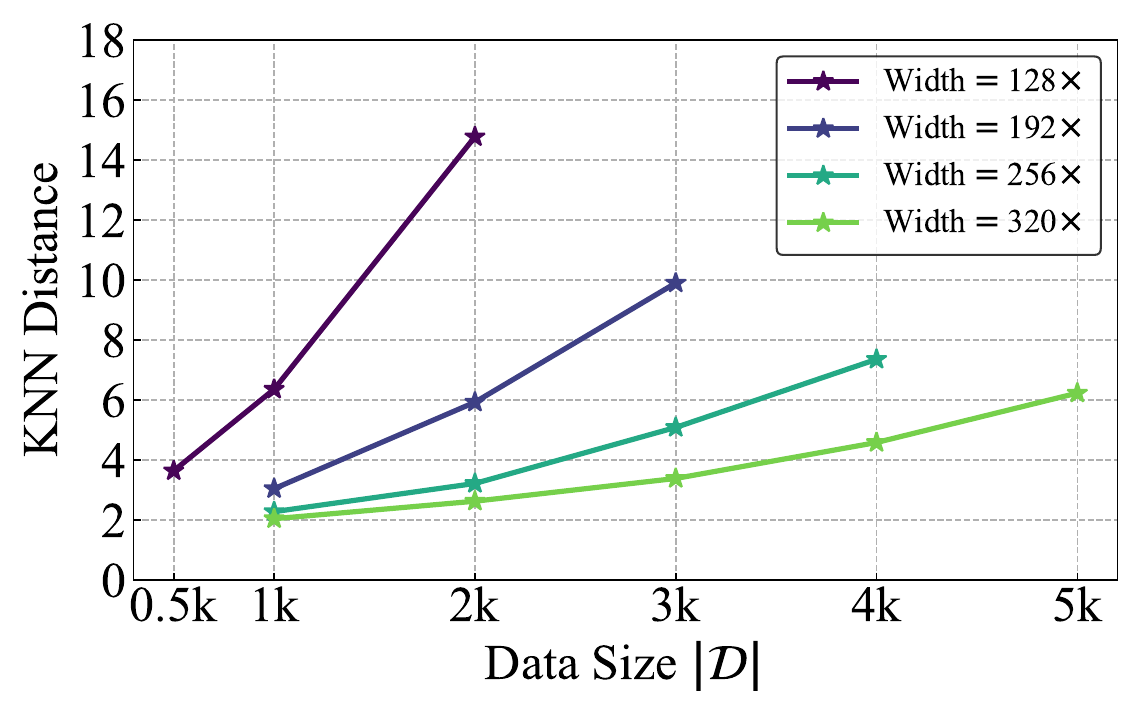}}
\subfigure[]{\includegraphics[width=0.32\textwidth]{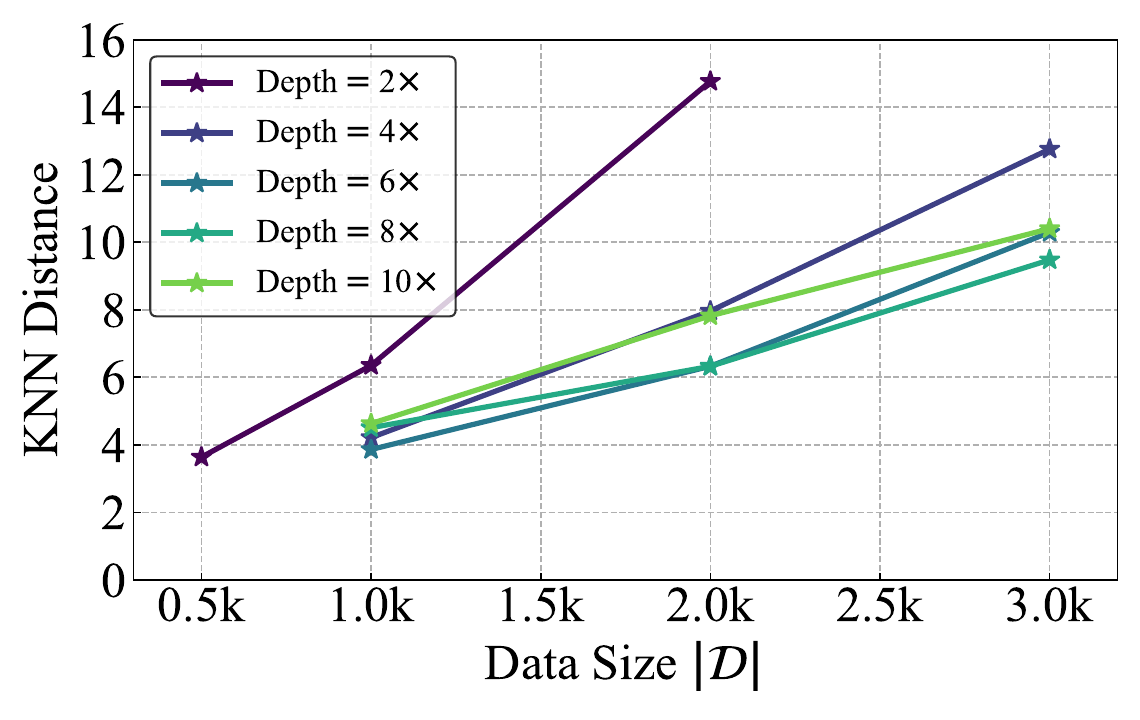}}
\subfigure[]{\includegraphics[width=0.32\textwidth]{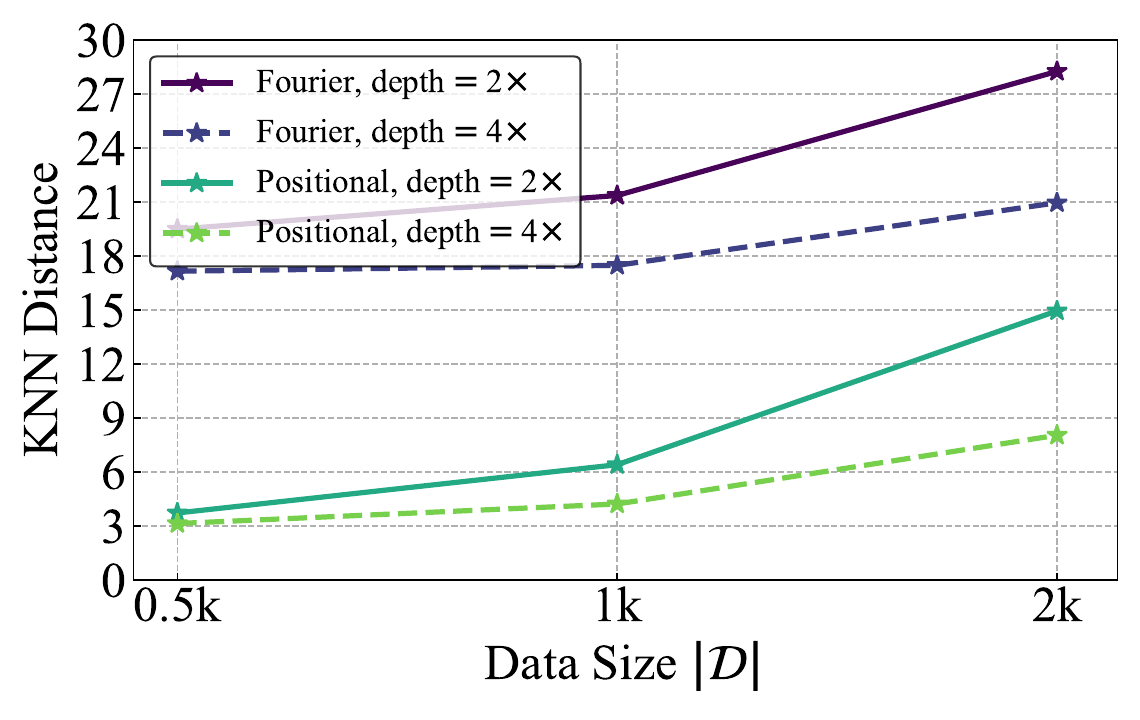}}
\label{fig_model_knn}
\caption{\textbf{KNN distances} of diffusion models on CIFAR-10 under different (a) model widths; (b) model depths; (c) time embeddings.}
\label{knn model}
\end{figure}

\begin{figure}[t]
\subfigure[]{\includegraphics[width=0.32\textwidth]{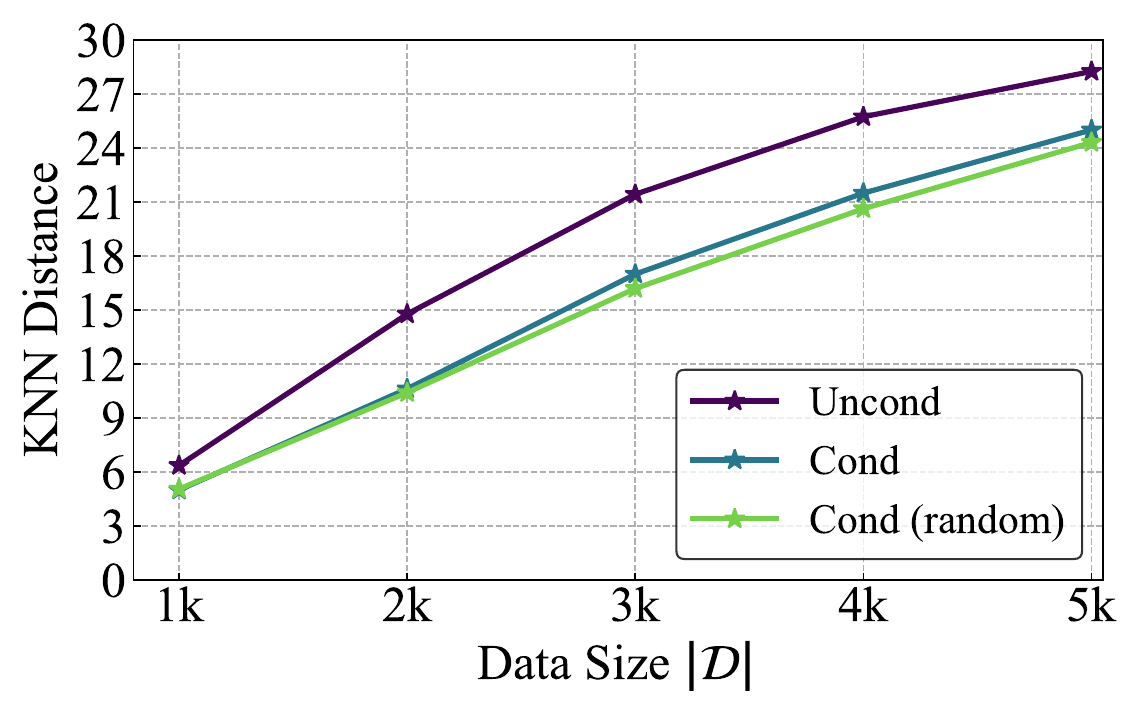}}
\subfigure[]{\includegraphics[width=0.32\textwidth]{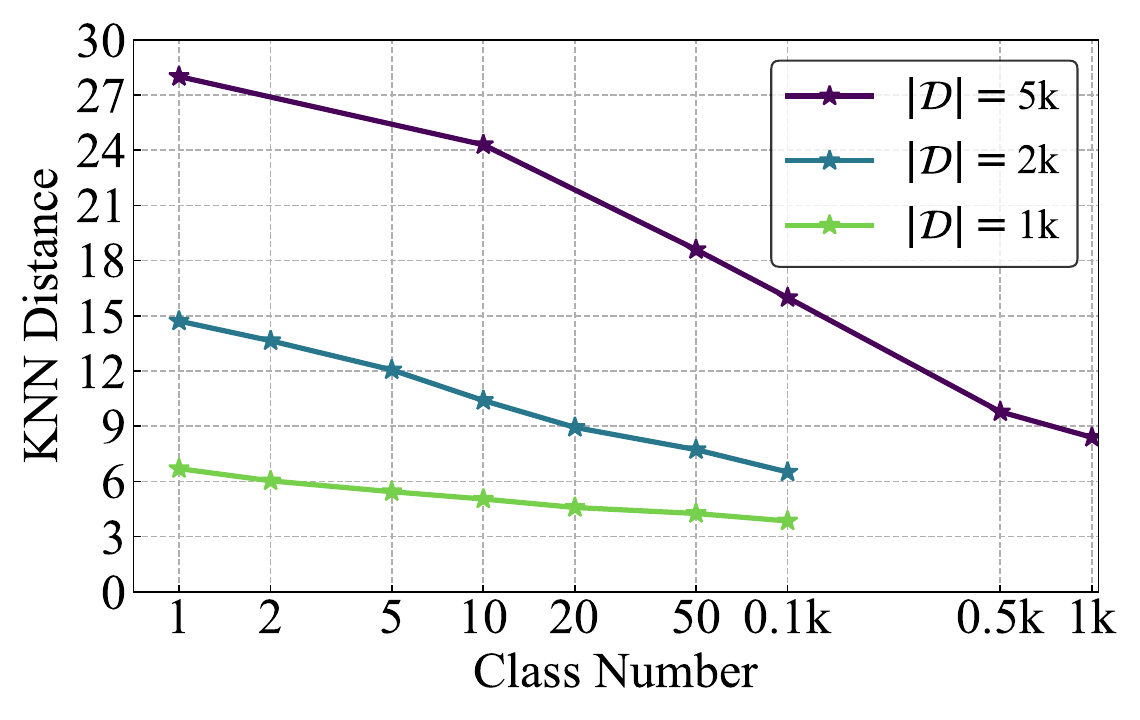}}
\subfigure[]{\includegraphics[width=0.32\textwidth]{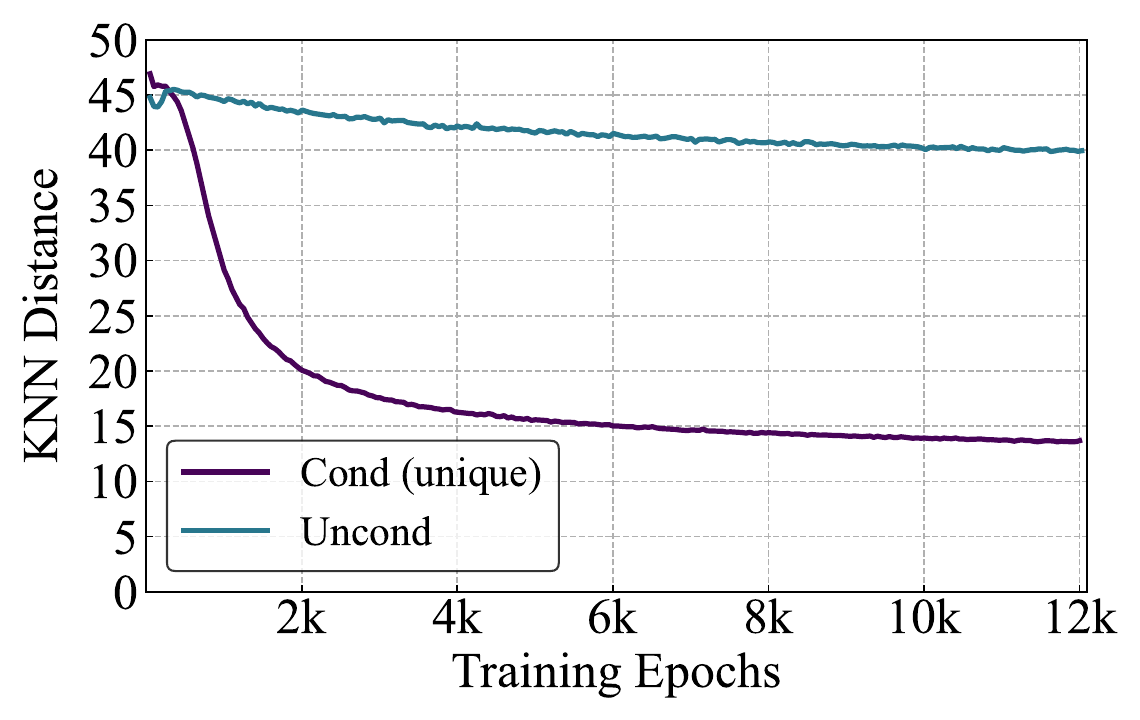}}
\caption{\textbf{KNN distances} of (a) unconditional diffusion models and conditional diffusion models with true labels or random labels on CIFAR-10; (b) conditional diffusion models with random labels of different $C$; (c) conditional EDM with unique labels and unconditional EDM at $|\mathcal{D}|=\,$50k during the training.}
\label{knn cond}
\end{figure}

\begin{figure}[t]
\centering
\subfigure[]{\includegraphics[width=0.44\textwidth]{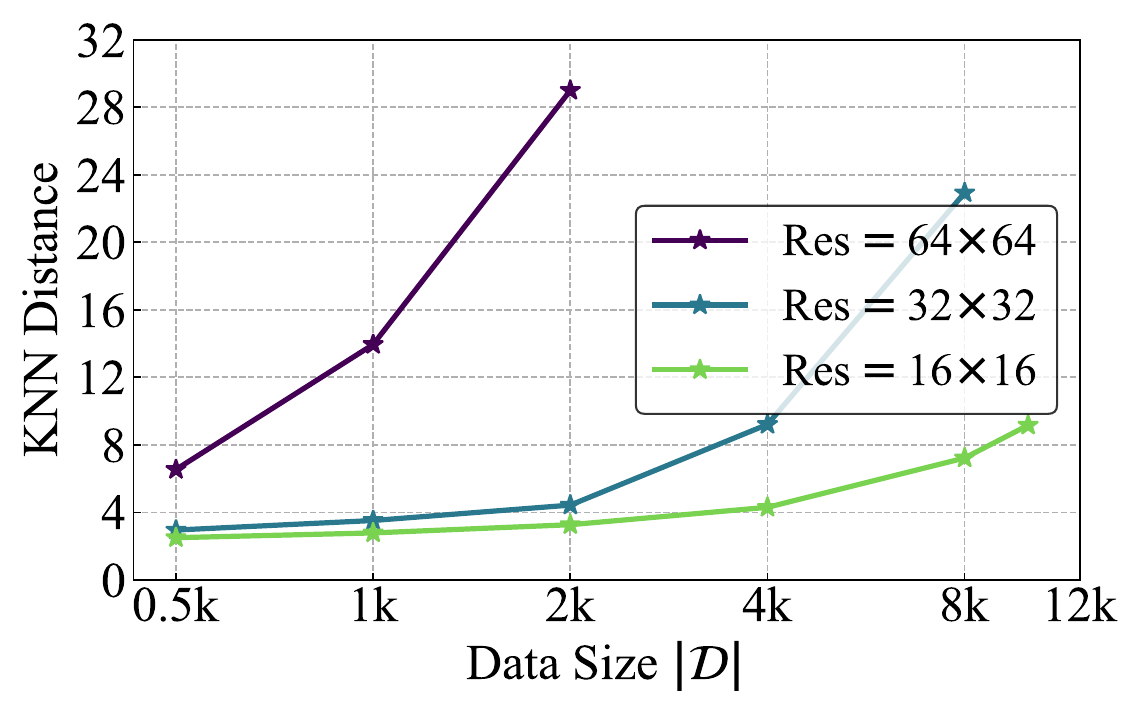}}
\subfigure[]{\includegraphics[width=0.44\textwidth]{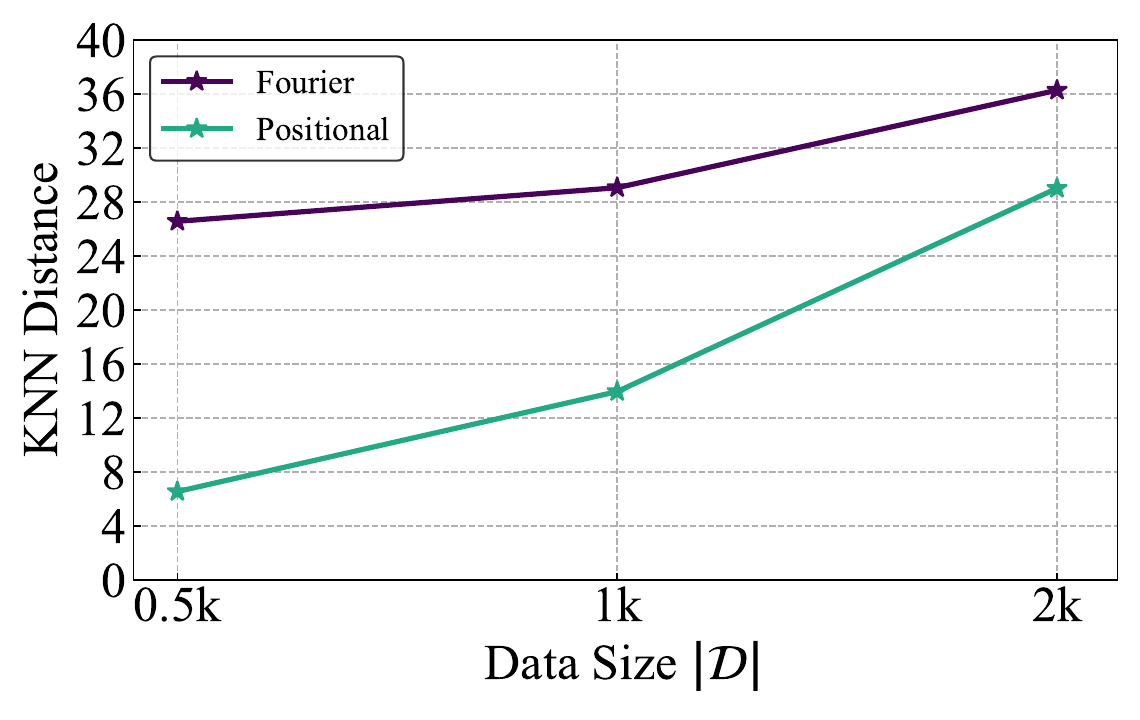}}
\subfigure[]{\includegraphics[width=0.44\textwidth]{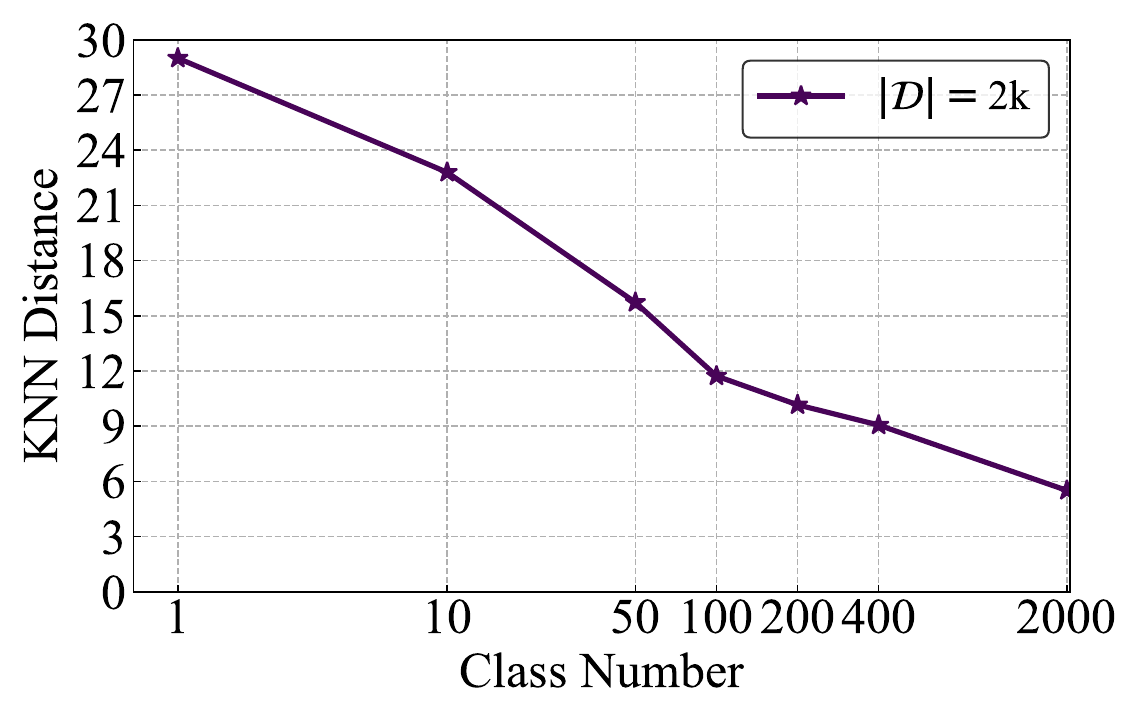}}
\subfigure[]{\includegraphics[width=0.44\textwidth]{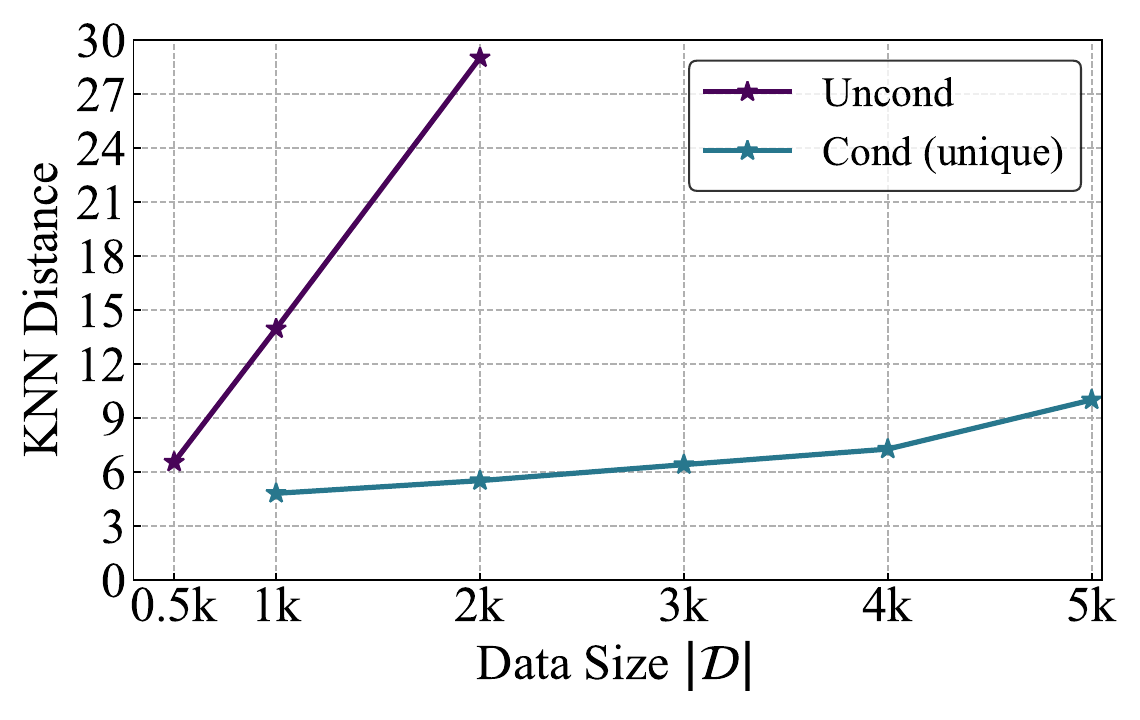}}
\vspace{-0.2cm}
\caption{\textbf{KNN distances} of diffusion models on FFHQ under (a) different data dimension; (b) different time embeddings; (c) different numbers of classes for random labels; (d) unconditional diffusion models and conditional diffusion models with unique conditions.\looseness=-1}
\label{knn ffhq}
\vspace{-0.2cm}
\end{figure}

It is noticed that when memorization ratio is high or KNN distance is low, the diffusion models feature memorizing more on the training data. Afterwards, we re-evaluate our results in the main paper by using the KNN distance as memorization metric. The results on CIFAR-10 are shown in Figure~(\ref{knn data}-\ref{knn cond}) and Table~(\ref{tbl_mem_batch_knn}-\ref{knn train}) while the results related to FFHQ are present in Figure~\ref{knn ffhq}. We notice that these new results are in alignment with our original conclusions using the memorization ratio metric.\looseness=-1

% We also consider the FID score~\citep{heusel2017gans} in our experiments.

\begin{figure}[t]
\centering
\subfigure[]{\includegraphics[width=0.245\textwidth]{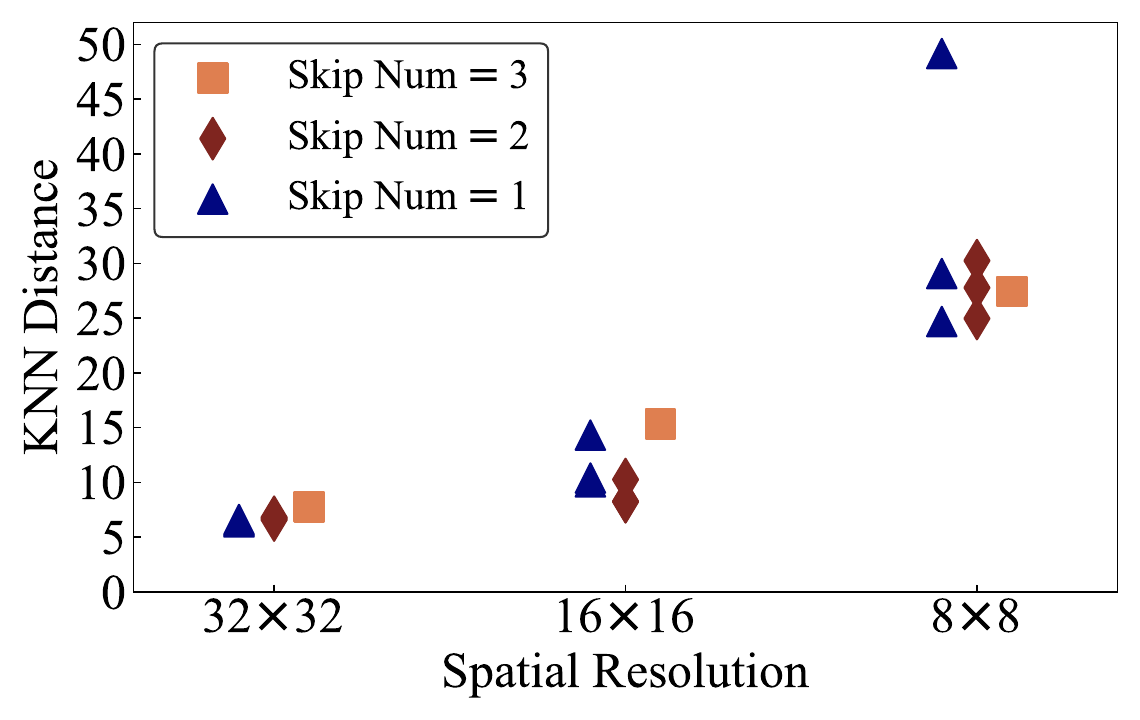}}
\subfigure[]{\includegraphics[width=0.245\textwidth]{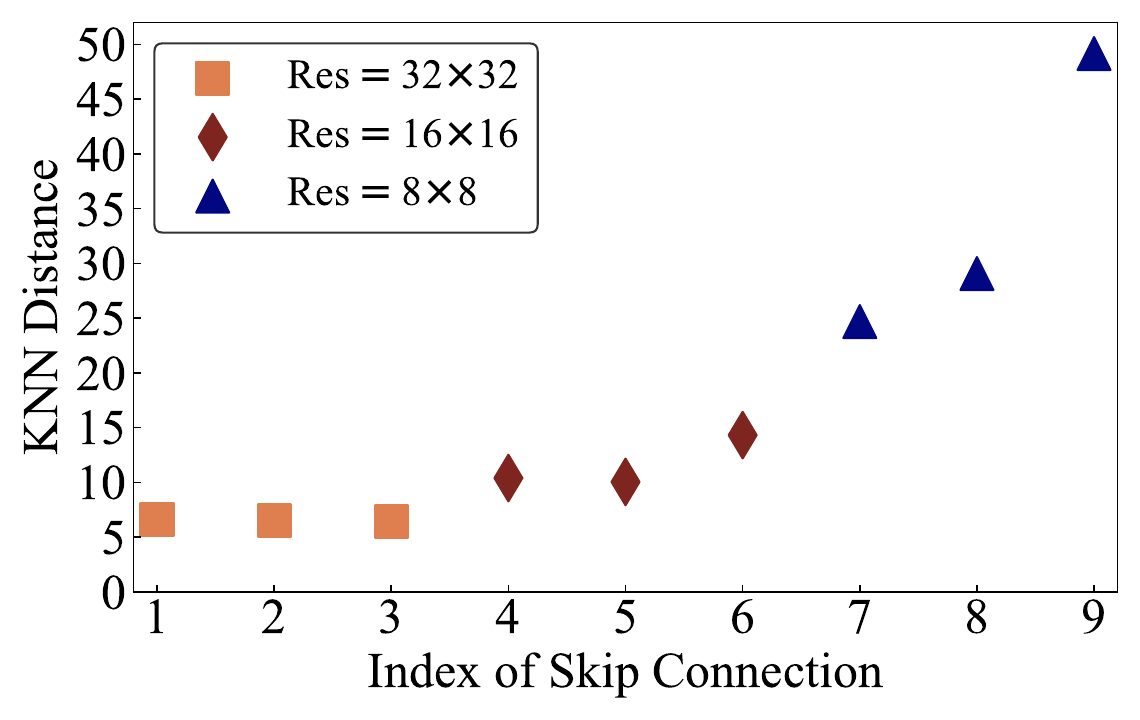}}
\subfigure[]{\includegraphics[width=0.245\textwidth]{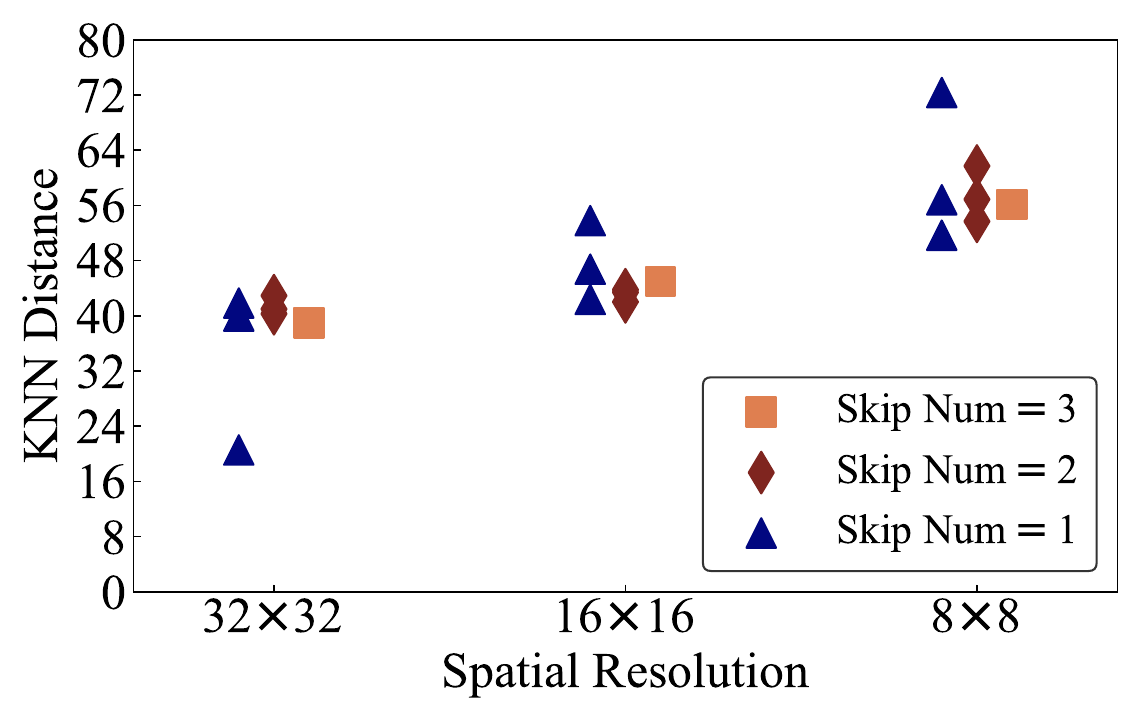}}
\subfigure[]{\includegraphics[width=0.245\textwidth]{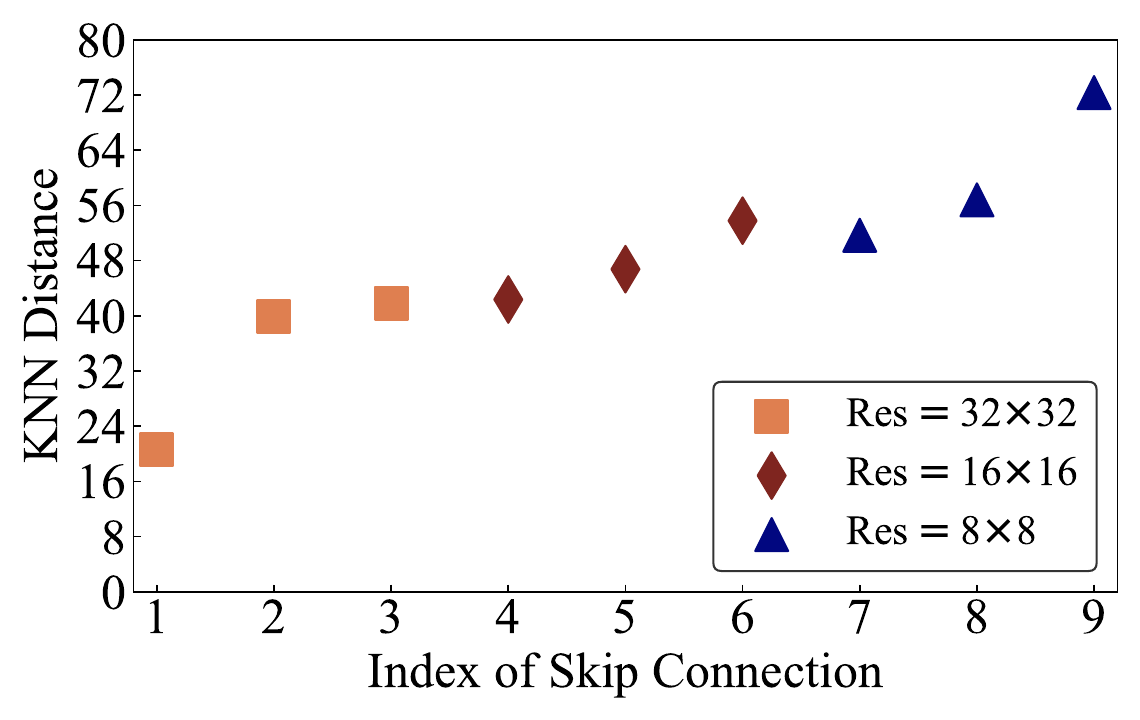}}
\caption{\textbf{KNN distances} of diffusion models on CIFAR-10 when retaining (a) skip connections of certain spatial resolution for DDPM++; (b) single skip connection at different locations for DDPM++; (c) skip connections of certain spatial resolution for NCSN++; (d) single skip connection at different locations for NCSN++.}
\end{figure}

% \begin{table}[t]
% \begin{minipage}[t]{0.46\linewidth}
% \begin{center}
% \begin{tabular}{c|cc}
% \toprule
% Learning rate     & $|\mathcal{D}|=1$k & $|\mathcal{D}|=2$k \\
% \midrule
% 1e-4 & 88.46& 54.82\\
% 2e-4 & 92.09& 60.93\\
% 5e-4 & \textbf{94.05}& 66.32\\
% 1e-3 & 93.65& \textbf{66.47}\\
% 2e-3 & 93.98& 65.72\\
% \bottomrule
% \end{tabular}
% \end{center}
% \caption{Effects of learning rate on ratio (\%) of memorization.}
% \label{mem_lr}
% \end{minipage}
% \hfill
% \begin{minipage}[t]{0.46\linewidth}
% \begin{center}
% \begin{tabular}{c|cc}
% \toprule
% Batch size     & $|\mathcal{D}|=1$k & $|\mathcal{D}|=2$k \\
% \midrule
% 128 & \textbf{92.85} & \textbf{64.91}\\
% 256 & 92.58& 63.55\\
% 512 & 92.09& 60.93\\
% 768 & 90.13& 59.18\\
% 1000 & 89.30& 57.53\\
% \bottomrule
% \end{tabular}
% \end{center}
% \caption{Effects of batch size on ratio (\%) of memorization.}
% \label{mem_bs}
% \end{minipage}
% \end{table}

\begin{table*}[t]
\begin{center}
\caption{\textbf{KNN distances} of diffusion models on CIFAR-10 under different batch sizes.}
\begin{tabular}{l|c|ccccccc}
\toprule
\multicolumn{2}{l|}{Batch Size} & 128 & 256 & 384 & 512 & 640 & 768 & 896 \\
\multicolumn{2}{l|}{Learning rate ($10^{-4}$)} & 0.5  & 1.0  & 1.5 & 2.0 & 2.5 & 3.0 & 3.5 \\
\midrule
\multirow{2}{*}{KNN distance}  &  $|\mathcal{D}|=$1k & 7.35 & 6.73 & 6.51 & 6.35 & 6.51 & 6.40 & \textbf{6.30} \\
 & $|\mathcal{D}|=$2k & 15.43 & 14.92 & 14.88 & 14.77 & 14.53 & \textbf{14.12} & 14.57\\
\bottomrule
\end{tabular}
\label{tbl_mem_batch_knn}
\end{center}
\end{table*}

\begin{table}[t]
\caption{\textbf{KNN distances} of diffusion models on CIFAR-10 under different (\emph{left}) weight decay values; (\emph{right}) EMA values.}
\begin{minipage}[t]{0.48\linewidth}
\begin{center}
\begin{tabular}{c|cc}
\toprule
\multirow{2}{*}{Weight decay} & \multicolumn{2}{c}{KNN distance} \\
& $|\mathcal{D}|=$1k & $|\mathcal{D}|=$2k\\
\midrule
$0$  & \textbf{6.35} & 14.77\\
$1\times10^{-5}$ & 6.45 & \textbf{14.63} \\
$1\times10^{-4}$ & 6.42 & 15.29\\
$1\times10^{-3}$ & 7.32& 16.99\\
$1\times10^{-2}$ & 9.99& 23.40\\
% $2\times10^{-2}$ & 14.17& 33.11\\
$5\times10^{-2}$ & 26.69& 38.73\\
% $8\times10^{-2}$ & 33.41& 39.99\\
$1\times10^{-1}$ & 34.44& 39.51\\
\bottomrule
\end{tabular}
\end{center}
% \vspace{-.3cm}
% \caption{Memorization ratios (\%) of different values of weight decay.}
% \vspace{-.3cm}
\end{minipage}
\hfill
\begin{minipage}[t]{0.48\linewidth}
\begin{center}
\begin{tabular}{l|cc}
\toprule
\multirow{2}{*}{EMA} & \multicolumn{2}{c}{KNN distance} \\
& $|\mathcal{D}|=$1k & $|\mathcal{D}|=$2k\\
\midrule
0.99929  & \textbf{6.35} & 14.77\\
0.999 & 6.44 & \textbf{14.58}\\
0.99 & 6.51 & 15.21\\
0.9 & 7.02 & 15.39\\
% 0.8 & 7.03 & 15.54\\
0.5 & 7.11 & 15.04\\
% 0.2 & 7.03 & 14.83\\
0.1 & 6.90 & 15.08\\
0.0 & 7.08 & 14.59\\
\bottomrule
\end{tabular}
\end{center}
\end{minipage}
% \vspace{-.2cm}
% \vspace{-.2cm}
\label{knn train}
\end{table}

% \input{tables/decay_and_ema_fid}

% \begin{figure}
%      \begin{subfigure}[t]{0.32\textwidth}
%          \centering
%          \includegraphics[width=\textwidth]{rebuttal_figures/memorize_model_width_knn.pdf}
%          \caption{}
%          % \label{fig_data_dim}
%      \end{subfigure}
%      \hfill
%      \begin{subfigure}[t]{0.32\textwidth}
%          \centering
%          \includegraphics[width=\textwidth]{rebuttal_figures/memorize_model_depth_knn.pdf}
%          \caption{}
%          % \label{fig_data_inter}
%      \end{subfigure}
%      \hfill
%      \begin{subfigure}[t]{0.32\textwidth}
%          \centering
%          \includegraphics[width=\textwidth]{rebuttal_figures/memorize_model_embed_knn.pdf}
%          \caption{}
%          % \label{fig_data_intra}
%      \end{subfigure}
%      \label{fig_skip_knn}
%         \caption{KNN distances of different (a) model widths; (b) model depths; (c) time embeddings.}
% \end{figure}
\clearpage

\section{Implementation Details of Finetuning on Stable Diffusion} \label{appendix sd}

\textbf{Data distribution.} Imagenette consists of $C=10$ classes from the ImageNet dataset~\citep{deng2009imagenet}. The classes include ``chain saw'', ``garbage truck'', ``tench'', ``French horn'', ``gas pump'', ``English springer'', ``parachute'', ``church'', ``cassette player'', ``golf ball''. In our experiments, the image resolution is specified as 256$\times$256. Similar to our experiments on CIFAR-10 and FFHQ, we disable all data augmentation.\looseness=-1

% The Artbench dataset is an artwork image dataset with class-balance, high quality as well as clean annotation~\citep{liao2022artbench}. It comprises 70k images with 10 categories, each of which represents the artistic style of image. The styles include ``art nouveau'', ``baroque'', ``expressionism'', ``impressionism'', ``post impressionism'', ``realism'', ``renaissance'', ``romanticism'', ``surrealism'', ``ukiyo e''. In our experiments, the image resolution is specified as 256$\times$256. Similar to our experiments on CIFAR-10 and FFHQ, we disable all data augmentation.

\textbf{Model configuration.} Stable diffusion~\citep{rombach2022high} includes an image encoder, an image decoder, a unet, and a text encoder. The image encoder is used to encode images into latent representations while the decoder reconstructs images from the latent. The unet plays a role as the latent diffusion model conditioning on text embeddings. Before fine-tuning the model on the Artbench dataset, we load the model weights from pre-trained stable diffusion\footnote{\url{https://huggingface.co/lambdalabs/miniSD-diffusers}}.\looseness=-1

\textbf{Training procedure.} During the fine-tuning, we only train the unet in stable diffusion and freeze the model weights of image encoder/decoder and text encoder. We adopt a learning rate of $1\times10^{-4}$ and $3\times10^{-4}$ for full fine-tuning and LoRA fine-tuning~\citep{hu2021lora} with cosine learning rate schedule and weight decay of $1\times10^{-6}$. The batch size is set as 16 and the gradient accumulation step is set as 4. For each training dataset $\mathcal{D}$, we enable the EMA and fine-tune the stable diffusion until 400k images have been observed.  

\textbf{Text-to-image generation.} We follow the pipeline of DDIM~\citep{song2021denoising} with 50 steps to sample 10k images from fine-tuned stable diffusion. During the generation, we disable the safety checker to prevent generating black images. For \emph{plain conditioning}, we use the text prompt ``a picture''. While for \emph{class conditioning}, we sample 1k images using the prompt ``a picture of <CLASS\_NAME>'' for each class, totaling 10k generated samples.\looseness=-1

\section{Sensitivity of Effective Model Memorization on \texorpdfstring{$\mathcal{\zeta}$}{TEXT}}\label{sensitivity}

\begin{figure}[t]
\centering
\subfigure[]{\includegraphics[width=0.32\textwidth]{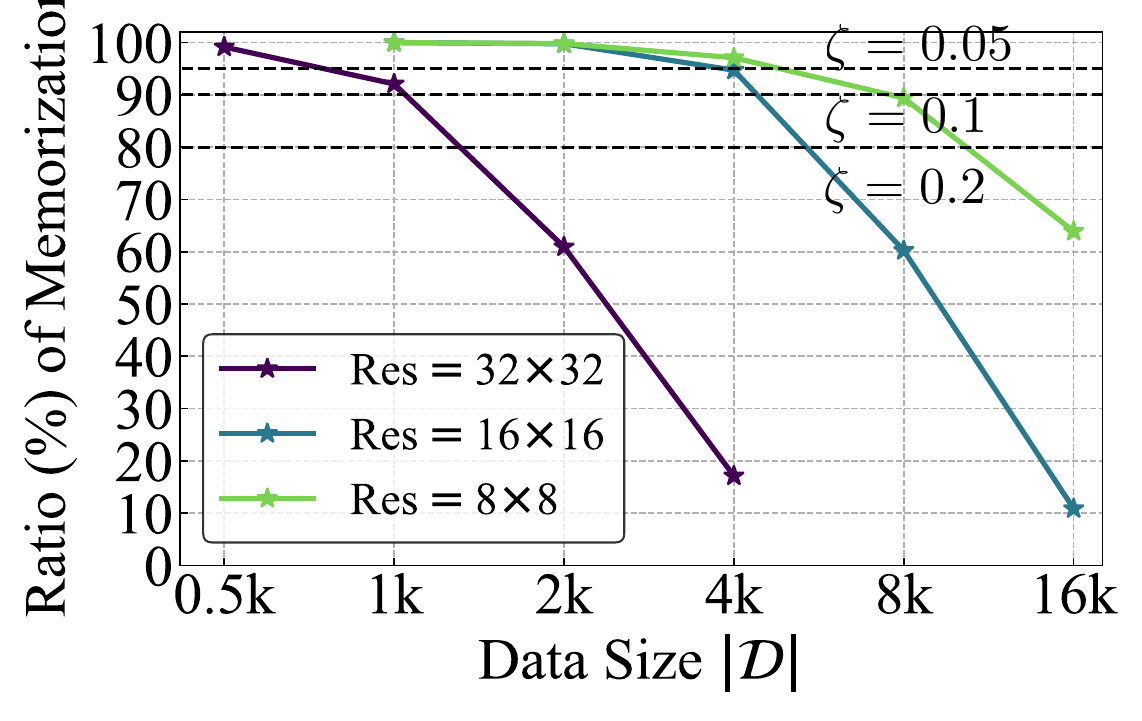}}
\subfigure[]{\includegraphics[width=0.32\textwidth]{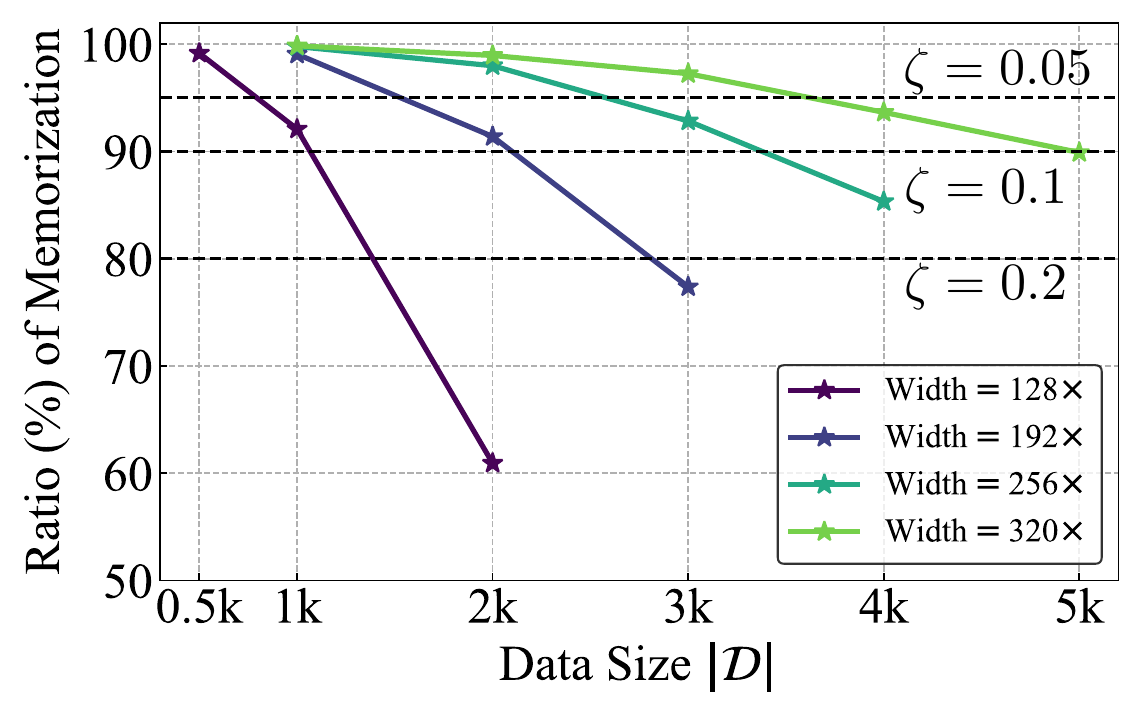}}
\subfigure[]{\includegraphics[width=0.32\textwidth]{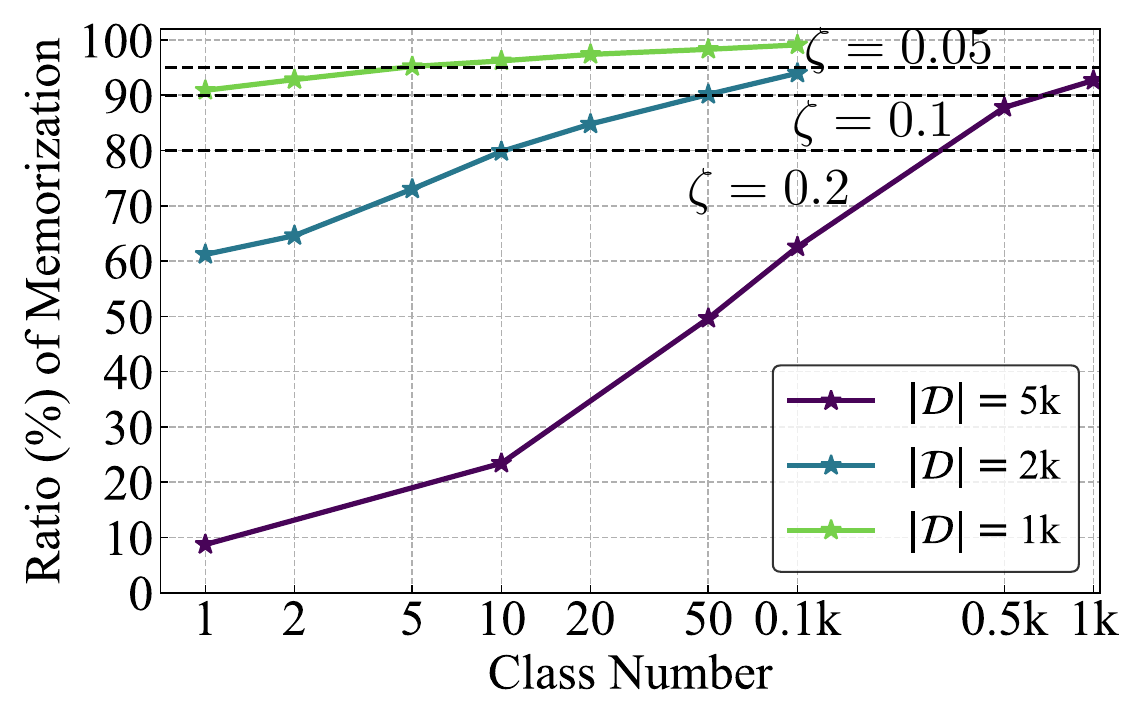}}
\vspace{-0.2cm}
\caption{EMMs of diffusion models with various $\zeta$ under configurations of (a) different data dimensions; (b) different model widths; (c) different numbers of classes for random labels.\looseness=-1}
\label{zeta sensitive}
\vspace{-0.2cm}
\end{figure}

The EMM metric defined in \ref{def1} is used to quantify the conditions which empirically trained diffusion models exhibit memorization behavior similar to the theoretical optimum.  The selection of $\zeta$ determines the closeness between the empirical model and theoretical one. Generally, a smaller $\zeta$ requires a higher memorization ratio for the empirical diffusion model to approximate the theoretical optimum. In Figure~\ref{zeta sensitive}, we select various $\zeta$ in different configurations. We observe that a smaller $\zeta$ typically results in a smaller EMM. Meanwhile, the conclusions in our main paper remain consistent. For instance, smaller data resolutions, larger model widths, larger numbers of classes for conditioning lead to more memorization.

% We include more experimental results on stable diffusion models using both ArtBench-10 and Imagenette datasets.

% \clearpage

% \subsection{Rebuttal}

% \begin{figure}[t]
% \subfigure[]{\includegraphics[width=0.32\textwidth]{rebuttal_figures/neurips2024_rebuttal/cat_mem.pdf}}
% \subfigure[]{\includegraphics[width=0.32\textwidth]{rebuttal_figures/neurips2024_rebuttal/cat_knn.pdf}}
% \subfigure[]{\includegraphics[width=0.32\textwidth]{rebuttal_figures/neurips2024_rebuttal/cat_fid.pdf}}
% % \caption{\textbf{KNN distances} of (a) unconditional diffusion models and conditional diffusion models with true labels or random labels on CIFAR-10; (b) conditional diffusion models with random labels of different $C$; (c) conditional EDM with unique labels and unconditional EDM at $|\mathcal{D}|=\,$50k during the training.}
% \label{rebuttal knn}
% \end{figure}

\end{document}